\documentclass{article}

\usepackage[preprint]{neurips_2026}

\usepackage{wrapfig}
\usepackage[utf8]{inputenc} % allow utf-8 input
\usepackage[T1]{fontenc}    % use 8-bit T1 fonts
\usepackage{hyperref}       % hyperlinks
\usepackage{url}            % simple URL typesetting
\usepackage{booktabs}       % professional-quality tables
\usepackage{amsfonts}       % blackboard math symbols
\usepackage{nicefrac}       % compact symbols for 1/2, etc.
\usepackage{microtype}      % microtypography
\usepackage{xcolor}         % colors
\usepackage{graphicx}
\usepackage{booktabs}
\usepackage{makecell}
\usepackage{float}
\usepackage{listings}
\usepackage{amsmath}
\usepackage{amsmath}
\usepackage[breakable,skins]{tcolorbox}
\usepackage{titletoc}
\usepackage{capt-of}
\usepackage{wrapfig}
\usepackage{booktabs}        % \toprule, \midrule, \bottomrule
\usepackage{graphicx}        % \resizebox
\usepackage[table]{xcolor}   % \rowcolor
\usepackage{caption}
\usepackage{makecell}   % for \makecell
\usepackage{pifont}     % for \ding
\usepackage{tabularx}
\usepackage{adjustbox}
\usepackage{caption}
\newcommand{\cmark}{\ding{51}}
\newcommand{\xmark}{\ding{55}}

\title{DocScope: Benchmarking Verifiable Reasoning for Trustworthy Long-Document Understanding}

\author{
\makebox[\textwidth][c]{%
Xiang Feng\textsuperscript{1} \hspace{1.5em}
Jiawei Zhou\textsuperscript{1} \hspace{1.5em}
Zhangfeng Huang\textsuperscript{2} \hspace{1.5em}
Kewei Wang\textsuperscript{3}}\\
\makebox[\textwidth][c]{%
\textbf{Shanshan Ye\textsuperscript{4}} \hspace{1.5em}
\textbf{Jinxin Hu\textsuperscript{2}} \hspace{1.5em}
\textbf{Zulong Chen\textsuperscript{2}}  \thanks{Corresponding Author} \hspace{1.5em}
\textbf{Yong Luo\textsuperscript{1}} \footnotemark[1] \hspace{1.5em}
\textbf{Jing Zhang\textsuperscript{1}}\footnotemark[1] \thanks{Project Leader}} \\[0.5em]
\makebox[\textwidth][c]{%
\textsuperscript{1} School of Computer Science, National Engineering Research Center for Multimedia Software}\\
\makebox[\textwidth][c]{%
and Hubei Key Laboratory of Multimedia and Network Communication Engineering,}\\
\makebox[\textwidth][c]{%
Wuhan University, China}\\
\makebox[\textwidth][c]{%
\textsuperscript{2} Alibaba Group, Hangzhou, China}\\
\makebox[\textwidth][c]{%
\textsuperscript{3} Independent Researcher}\\
\makebox[\textwidth][c]{%
\textsuperscript{4} Department of Machine Learning,} \\
\makebox[\textwidth][c]{
 Mohamed bin Zayed University of Artificial Intelligence, United Arab Emirates 
}
\\[0.5em]
\makebox[\textwidth][c]{%
\texttt{fengxiang\_cs@whu.edu.cn, 2021302111478@whu.edu.cn}%
}\\
\makebox[\textwidth][c]{%
\texttt{huangzhangfeng.hzf@alibaba-inc.com, kyrakeweiwang@mail.ustc.edu.cn}%
}\\
\makebox[\textwidth][c]{%
\texttt{cassie.ye133@hotmail.com, jinxin.hjx@alibaba-inc.com}%
}\\
\makebox[\textwidth][c]{%
\texttt{zulong.czl@alibaba-inc.com, luoyong@whu.edu.cn, jingzhang.cv@gmail.com}%
}}

\begin{document}

\maketitle

\begin{abstract}
Evaluating whether Multimodal Large Language Models can produce trustworthy, verifiable reasoning over long, visually rich documents requires evaluation beyond end-to-end answer accuracy. We introduce DocScope, a benchmark that formulates long-document QA as a structured reasoning trajectory prediction problem: given a complete PDF document and a question, the model outputs evidence pages, supporting evidence regions, relevant factual statements, and a final answer. We design a four-stage evaluation protocol---Page Localization, Region Grounding, Fact Extraction, and Answer Verification---that audits each level of the trajectory independently through inter-stage decoupling, with all judges selected and calibrated via human alignment studies. DocScope comprises 1,124 questions derived from 273 documents, with all hierarchical evidence annotations completed by human annotators. We benchmark 6 proprietary models, 12 open-weight models, and several domain-specific systems. Our experiments reveal that answer accuracy cannot substitute for trajectory-level evaluation: even among correct answers, the highest observed rate of complete evidence chains is only 29\%. Across all models, region grounding remains the weakest trajectory stage. Furthermore, the primary difficulty stems from aggregating evidence dispersed across long distances and multiple document clusters, while an oracle study identifies faithful perception and fact extraction as the dominant capability bottleneck. Cross-architecture comparisons further suggest that activated parameter count matters more than total scale. The benchmark and code will be publicly released at \href{https://github.com/MiliLab/DocScope}{https://github.com/MiliLab/DocScope}.
\end{abstract}

\begin{figure}[h!]
    \centering
    \includegraphics[width=\linewidth]{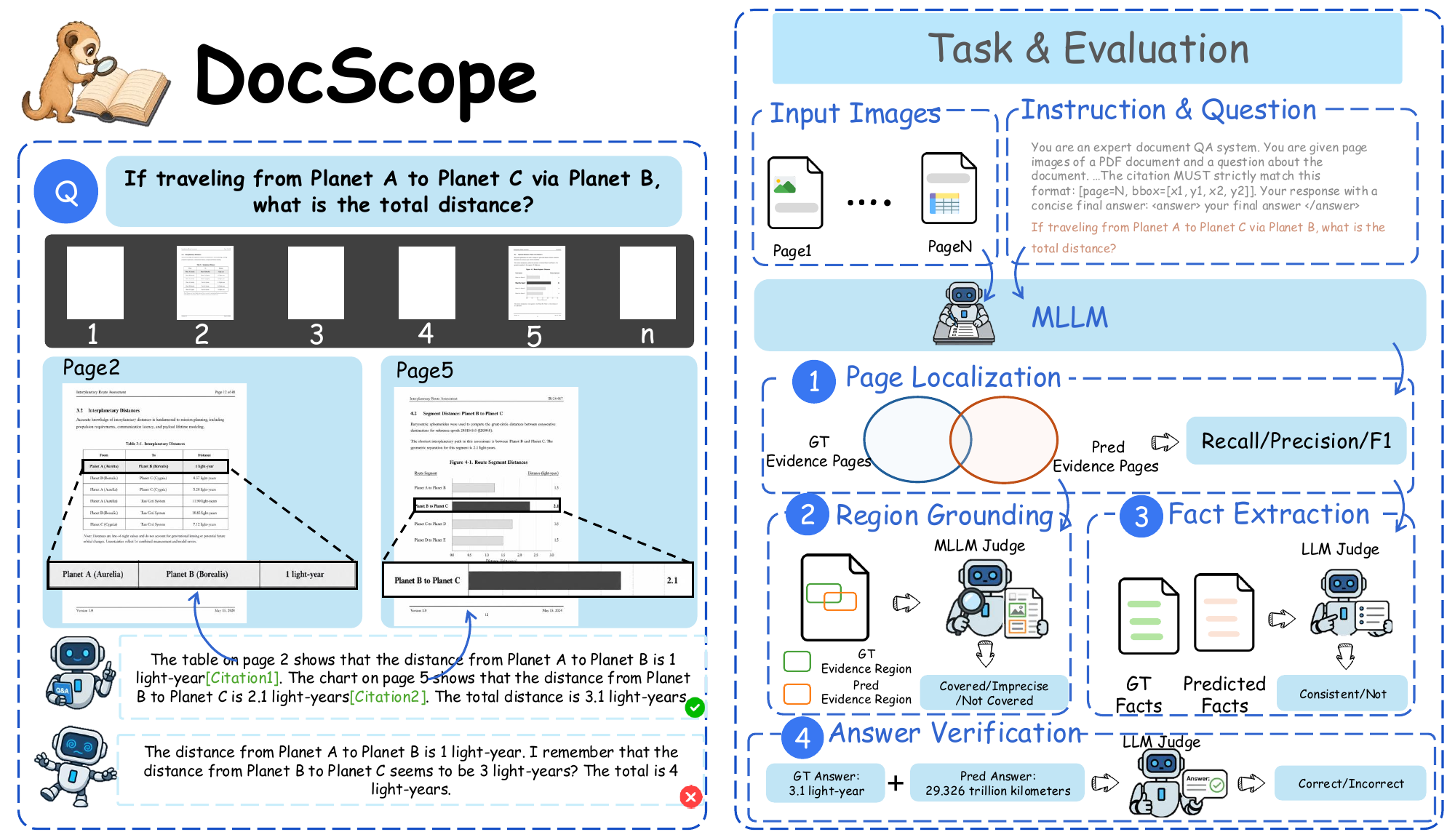}
    \caption{Overview of DocScope. Left: given a long document and a question (the example shown is illustrative and fictional), a trustworthy response should ground each claim in specific pages and regions with inline citations, whereas an unverifiable response may produce a plausible but hallucinated answer. Right: our four-stage evaluation protocol audits the model's structured reasoning trajectory independently at each level---Page Localization, Region Grounding, Fact Extraction, and Answer Verification.}
    \label{fig:dataset_pipeline}
\end{figure}

\section{Introduction}
\label{sec:introduction}

In recent years, Multimodal Large Language Models (MLLMs) have achieved significant advances in document understanding tasks, driven by continuous improvements in their perception, reasoning, and context-handling capabilities~\citep{jaech2024openai,qwen35,huang2026visionr}. As these models are increasingly deployed in realistic application scenarios, merely generating answers is no longer sufficient to meet user requirements. The trustworthiness of a question-answering system becomes paramount, requiring model outputs to be not only accurate but also grounded in the source document, so that responses are verifiable and auditable. A critical and underexplored question therefore naturally arises: Can MLLMs serve as question-answering systems that produce trustworthy, end-to-end reasoning traces grounded in long documents?

Although recent benchmarks have begun to explore long-document understanding~\citep{ma2024mmlongbenchdoc,deng2025longdocurl,chia2025mlongdoc} and verifiable multimodal generation~\citep{hu-etal-2025-mcitebench,song2026mavis}, a notable gap persists between current benchmark designs and the practical requirements of trustworthy document question answering. As summarized in Tab.~\ref{tab:intro_benchmark_comparison}, prior benchmarks typically address only a subset of the evidence-verification problem, leaving several critical dimensions underexplored. \textbf{(1) Abstracted input context.} Several citation-oriented benchmarks~\citep{hu-etal-2025-mcitebench,song2026mavis} evaluate models over a fixed or pre-retrieved evidence pool, rather than requiring the model to process the full document as it naturally appears in realistic use. While this setting is valuable for measuring citation fidelity, it bypasses the challenging step of navigating lengthy, visually rich documents where relevant evidence may be dispersed across numerous pages. \textbf{(2) Coarse verification granularity.} The granularity of evidence supervision is often limited to document sources, pages, passages, or pre-defined evidence items. Such granularity proves insufficient for visually complex documents, where a single page may encompass multiple tables, figures, captions, paragraphs, and layout regions. In practice, users need to know not only \emph{which page} supports an answer, but also \emph{which specific region} and \emph{which factual statement} provide the supporting evidence. \textbf{(3) Implicit evidence trajectories.} Existing long-document benchmarks~\citep{ma2024mmlongbenchdoc,deng2025longdocurl} frequently leverage evidence metadata for dataset construction or post-hoc analysis, yet do not require models to explicitly produce a complete evidence trajectory as an integral part of the task output. Consequently, they cannot jointly assess whether a model is capable of locating relevant evidence, grounding it spatially, extracting faithful facts, and arriving at a correct final answer within a unified evaluation protocol.

To address these limitations, we introduce \textbf{DocScope} (Fig.~\ref{fig:dataset_pipeline}), a hierarchical evaluation benchmark for verifiable long-document question answering. DocScope is constructed from complete, visually rich PDF documents rather than isolated evidence pools or pre-retrieved pages. To support fine-grained evaluation, we recruited 13 annotators from two research institutions to perform hierarchical evidence annotation. For each question, we provide human annotations at three evidence levels: evidence pages, evidence regions, and factual statements distilled from the corresponding regions; for answerable questions, a gold answer is provided. In total, DocScope comprises 1,124 questions synthesized by MLLMs, while all answers and evidence annotations are completed by human annotators.

At evaluation time, models are required to output a structured reasoning trajectory in a unified format, enabling us to assess not only the correctness of the final answer but also whether it is supported by verifiable evidence at the page, region, and fact levels. Building upon this framework, we conduct extensive experiments benchmarking 6 proprietary models, 12 open-weight models, and several domain-specific systems on DocScope, accompanied by analyses that yield key insights. First, answer accuracy on long-document questions remains limited, and, more importantly, it cannot substitute for trajectory-level evaluation: even among correctly answered samples, the highest observed rate of complete evidence chains is only 29\%, revealing a pronounced decoupling between answer correctness and reasoning trustworthiness. Second, the difficulty of long-document QA is driven not only by the amount of required evidence, but more critically by whether the evidence is dispersed across long distances and multiple document clusters, a real-world challenge that fixed-evidence-pool settings largely bypass. Third, an oracle evidence access study shows that faithful fact extraction constitutes a dominant capability bottleneck, indicating that reliable document QA requires models not merely to retrieve relevant evidence, but also to perceive and transform it into accurate fact.

In summary, we make three contributions: (1)~we formulate long-document QA as a structured reasoning trajectory prediction problem and design a well-calibrated, four-stage evaluation protocol that diagnoses each level of the trajectory independently; (2)~we construct DocScope, a high-quality benchmark of 1{,}124 questions with hierarchical human-annotated evidence at the page, region, and fact levels; and (3)~through extensive experiments and analyses, we provide actionable insights into the gap between answer accuracy and reasoning trustworthiness in current MLLMs.

\begin{table}[t]
\centering
\small
\setlength{\tabcolsep}{3.2pt}
\caption{Comparison between \textsc{DocScope} and previous representative benchmarks.
Green indicates that the feature is supported, while red indicates that it is not.
``Gen.'' is short for ``Generated automatically''.}
\resizebox{\linewidth}{!}{%
\begin{tabular}{@{}l l l ccc cc c r@{}}
\toprule
& & &
\multicolumn{3}{c}{\textbf{Annotation Granularity}} &
& & & \\
\cmidrule(lr){4-6}
\textbf{Benchmark} &
\textbf{Venue} &
\textbf{Input} &
\makecell{\textbf{Page/}\\\textbf{Item}} &
\textbf{Bbox} &
\textbf{Fact} &
\makecell{\textbf{Answer}\\\textbf{Source}} &
\makecell{\textbf{Annotation}\\\textbf{Source}} &
\makecell{\textbf{Verifiable}\\\textbf{Evaluation}} &
\textbf{\#Q} \\
\midrule

MCiteBench &
\scriptsize\color{gray}Findings EMNLP 2025 &
Fixed evidence pool &
\cellcolor{green!8}\cmark &
\cellcolor{red!6}\xmark &
\cellcolor{red!6}\xmark &
Web + Gen. &
Human &
Item &
3{,}000 \\

MAVIS &
\scriptsize\color{gray}AAAI 2026 &
Retrieved text/images &
\cellcolor{green!8}\cmark &
\cellcolor{red!6}\xmark &
\cellcolor{green!8}\cmark &
Web &
Gen. &
Item + Fact &
1{,}000 \\

MMLongBench-Doc &
\scriptsize\color{gray}NeurIPS 2024 DB &
Full document &
\cellcolor{green!8}\cmark &
\cellcolor{red!6}\xmark &
\cellcolor{red!6}\xmark &
Human &
Human &
-- &
1{,}062 \\

LongDocURL &
\scriptsize\color{gray}ACL 2025 &
30-page window &
\cellcolor{green!8}\cmark &
\cellcolor{green!8}\cmark &
\cellcolor{red!6}\xmark &
Gen. &
Gen. &
-- &
2{,}325 \\

M-LongDoc &
\scriptsize\color{gray}EMNLP 2025 &
Retrieved pages &
\cellcolor{green!8}\cmark &
\cellcolor{red!6}\xmark &
\cellcolor{red!6}\xmark &
Gen. &
Gen. &
-- &
851 \\

MMDocRAG &
\scriptsize\color{gray}NeurIPS 2025 DB &
Fixed evidence pool &
\cellcolor{green!8}\cmark &
\cellcolor{red!6}\xmark &
\cellcolor{red!6}\xmark &
Gen. + Human &
Gen. + Human &
Item &
4{,}055 \\

\midrule

\textbf{DocScope} &
\textbf{--} &
\textbf{Full document} &
\cellcolor{green!15}\cmark &
\cellcolor{green!15}\cmark &
\cellcolor{green!15}\cmark &
\textbf{Human} &
\textbf{Human} &
\makecell[c]{\textbf{Page + Bbox}\\\textbf{+ Fact}} &
\textbf{1{,}124} \\

\bottomrule
\end{tabular}%
}
\label{tab:intro_benchmark_comparison}
\vspace{-1.0em}
\end{table}

\section{Dataset}
\label{sec:dataset}

This section presents the design of DocScope. We first formalize the structured reasoning task for trustworthy long-document question answering (Section~\ref{sec:dataset:taskdefinition}), then describe how we construct the benchmark and corresponding ground-truth annotations (Section~\ref{sec:dataset:construction}), present dataset statistics (Section~\ref{sec:dataset:overview}), and detail the evaluation protocol (Section~\ref{subsec:evaluation_protocol}).

\subsection{Task Definition}
\label{sec:dataset:taskdefinition}

When a QA system is used over long, visually rich documents, users need more than a correct answer---they need to verify \emph{why} the answer is correct by tracing it back to specific evidence in the source. This requires the system to output not merely a final answer, but a structured reasoning trajectory that can be independently checked. We therefore formulate long-document question answering as a structured prediction problem in which the model must produce an explicit, multi-level evidence trajectory alongside its answer.

Specifically, given a long document $\mathcal{D}=\{p_1,\dots,p_N\}$ and a question $q$, where $p_i$ denotes the $i$-th page, the model is required to output a structured reasoning trajectory:
\begin{equation}
\label{eq:1}
y
=
\big(
\mathcal{P},
\mathcal{R},
\mathcal{F},
a
\big).
\end{equation}
Each level of the trajectory addresses a progressively finer audit question (the inference prompt is given in Appendix~\ref{appendix:inference_prompt}). $\mathcal{P}\subseteq\{1,\dots,N\}$ identifies \emph{where} the evidence resides (evidence pages); $\mathcal{R} = \{(i, [x_1, y_1, x_2, y_2]) \mid i \in \hat{\mathcal{P}},\; 0 \leq x_1, y_1, x_2, y_2 \leq 1\}$ specifies \emph{what} to look at on those pages (grounded evidence regions), where each four-tuple $[x_1, y_1, x_2, y_2]$ denotes a bounding box with $(x_1, y_1)$ as the top-left corner and $(x_2, y_2)$ as the bottom-right corner in normalized page coordinates; $\mathcal{F}$ makes explicit \emph{what the model understood} from those regions (factual statements); and $a$ is the final answer derived from these facts. Any break in this chain---a missing page, an imprecise region, or a hallucinated or missing fact---renders the output unverifiable, regardless of whether the final answer happens to be correct.

As shown in Fig.~\ref{fig:dataset_pipeline}, we evaluate each level of this trajectory independently against human-annotated ground truth, enabling fine-grained diagnosis of where and how the reasoning trace fails. \emph{DocScope} is the benchmark we construct to support this evaluation: it provides hierarchical evidence annotations at the page, region, and fact levels, making it possible to assess not only answer correctness but also the trustworthiness of the entire reasoning process that leads to it.

\subsection{Dataset Construction}
\label{sec:dataset:construction}

\begin{wrapfigure}[11]{l}{0.5\linewidth}
    \vspace{-1.0em}
    \centering
    \includegraphics[width=\linewidth]{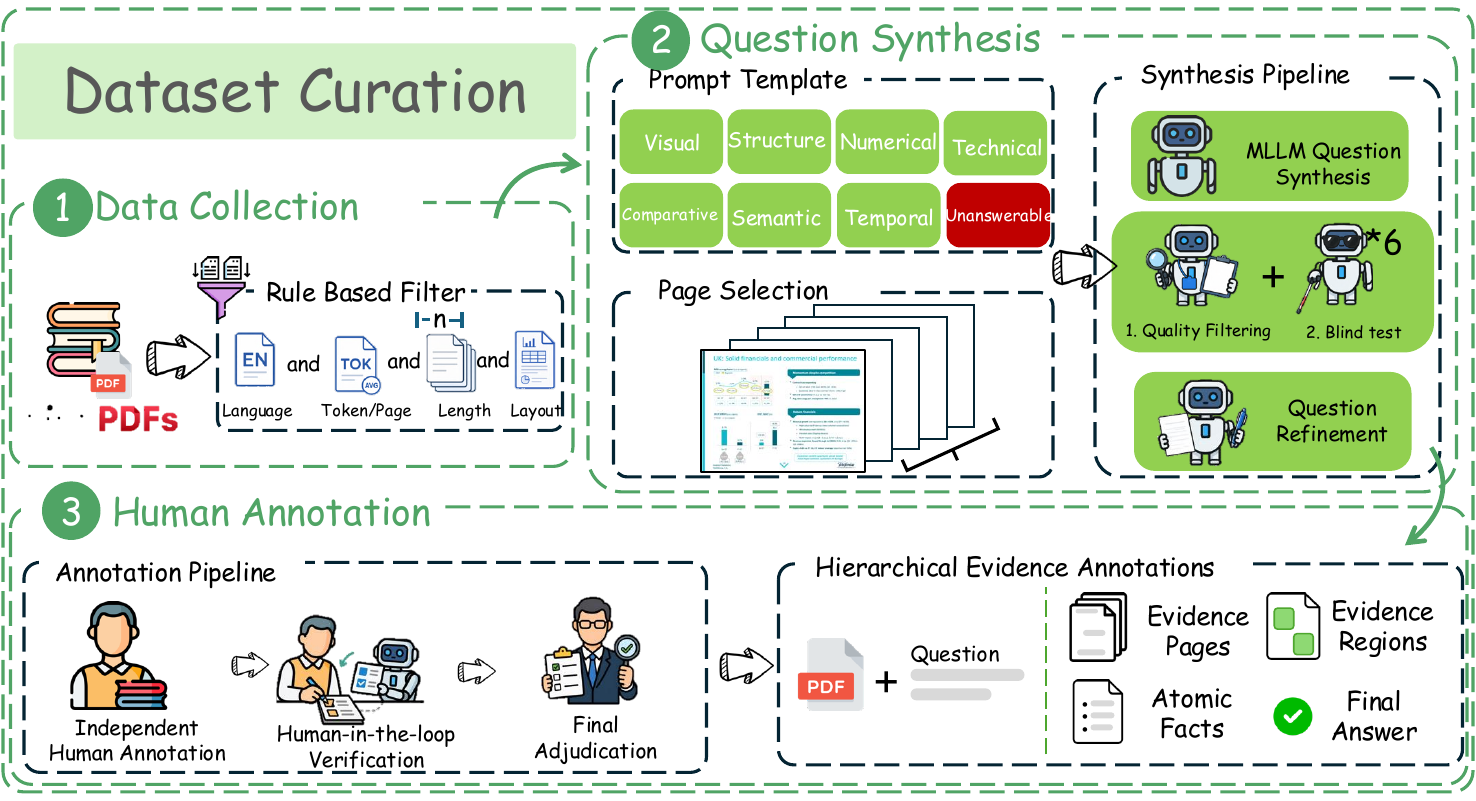}
    \caption{Dataset curation pipeline.  }
    \label{fig:dataset_dataset}
    \vspace{-1.0em}
\end{wrapfigure}
\paragraph{Data Collection.} We source documents from the publicly available FinePDF\footnote{https://huggingface.co/datasets/HuggingFaceFW/finepdfs} corpus, applying metadata-based and layout-based filters to retain long, visually rich documents with interleaved text, figures, and tables, followed by manual inspection to remove low-quality or overly specialized material (detailed filtering criteria in Appendix~\ref{appendix:construction_details}), producing a pool of high-quality, visually rich documents.
\paragraph{Question Synthesis.}
We design an automated pipeline that clusters page embeddings to identify information-dense segments, then prompts Claude-Opus-4.6~\citep{claude-opus46} to synthesize diverse questions across eight categories (Appendix~\ref{appendix:question_synthesis_prompts}). After quality filtering and deduplication, we conduct multi-model blind testing with six frontier models to discard questions answerable without document access. The full synthesis pipeline is described in Appendix~\ref{appendix:construction_details}.
\paragraph{Human Annotation and Quality Control.}
13 annotators from two research institutions construct the gold reasoning trajectory for each question following Eq.~\ref{eq:1}: evidence pages~($\mathcal{P}^{*}$), bounding-box regions~($\mathcal{R}^{*}$), factual statements~($\mathcal{F}^{*}$), and a final answer~($a^{*}$). Annotations are refined through a human-in-the-loop verification stage with model-assisted checking~\citep{gemini31-flash-lite}, followed by adjudication from two senior members outside the annotation team. Further details are provided in Appendix~\ref{appendix:annotation_details}.

\subsection{Overview and Statistics}
\label{sec:dataset:overview}

\begin{figure}[t]
\centering

\begin{adjustbox}{valign=b}
\begin{minipage}{0.5\linewidth}
\centering
\small
\setlength{\tabcolsep}{2pt}
\captionof{table}{Basic statistics of DocScope.}
\begin{tabular}{@{}lr@{}}
\toprule
\textbf{Statistic} & \textbf{Number} \\
\midrule
\multicolumn{2}{@{}l}{\textbf{Documents}} \\
\quad - Documents with questions & 273 \\
\quad - Avg./Max. pages & 51.3 / 100 \\
\quad - Avg./Max. text tokens & 24,561 / 143,868 \\
\quad - Avg./Max. questions per document & 4.12 / 17 \\
\midrule
\multicolumn{2}{@{}l}{\textbf{Question \& Answer}} \\
\quad - Total questions & 1,124 \\
\quad - Avg./Max. question tokens & 29.4 / 70 \\
\quad - Avg./Max. answer tokens & 5.3 / 71 \\
\midrule
\multicolumn{2}{@{}l}{\textbf{Dataset Split}} \\
\quad - Test set & 730 (64.9\%) \\
\quad - Validation set & 394 (35.1\%) \\
\midrule
\multicolumn{2}{@{}l}{\textbf{Number of Evidence Pages}} \\
\quad - Single-page questions & 397 (35.3\%) \\
\quad - Multi-page questions & 649 (57.7\%) \\
\quad - Unanswerable questions & 78 (6.9\%) \\
\midrule
\multicolumn{2}{@{}l}{\textbf{Evidence Region}} \\
\quad - Avg./Max. evidence per question & 3.99 / 64 \\
\quad - Avg./Max. evidence relative area & 9.79\% / 83.09\% \\
\midrule
\multicolumn{2}{@{}l}{\textbf{Facts}} \\
\quad - Avg./Max. facts per question & 4.99 / 64 \\
\quad - Avg./Max. facts per evidence & 1.62 / 14.00 \\
\quad - Avg./Max. fact description tokens & 19.9 / 112 \\
\bottomrule
\end{tabular}
\label{tab:dataset_basic_statistics}
\end{minipage}
\end{adjustbox}
\hfill
\begin{adjustbox}{valign=b}
\begin{minipage}{0.42\linewidth}
\centering
\includegraphics[width=0.9\linewidth]{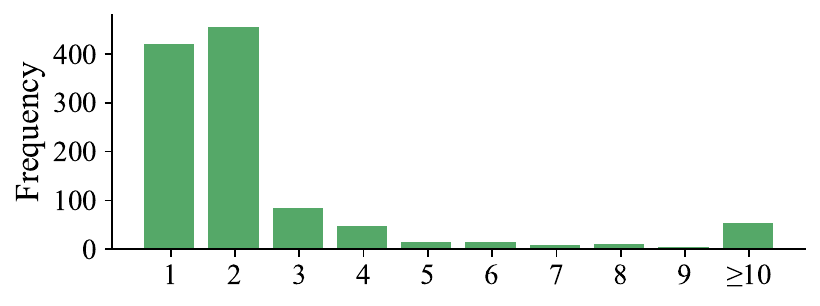}\\[-0.3em]
\scriptsize\textbf{(a)} Evidence pages per question\\[0.25em]

\includegraphics[width=0.9\linewidth]{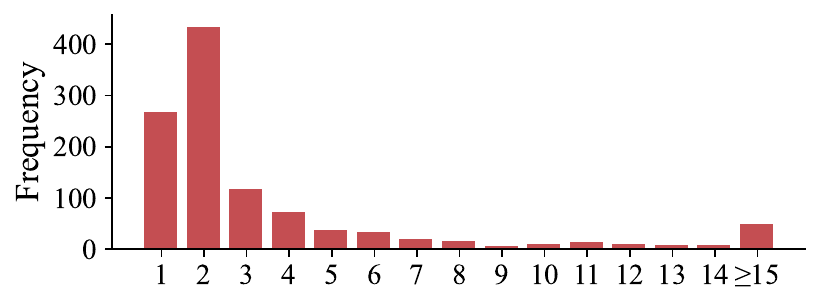}\\[-0.3em]
\scriptsize\textbf{(b)} Evidence regions per question\\[0.25em]

\includegraphics[width=0.9\linewidth]{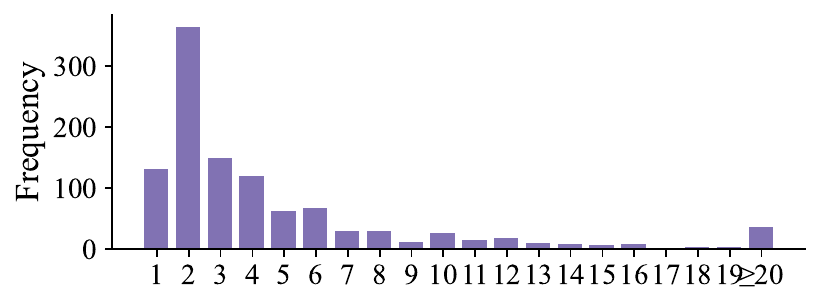}\\[-0.3em]
\scriptsize\textbf{(c)} Facts per question\\[0.25em]

\includegraphics[width=0.7\linewidth]{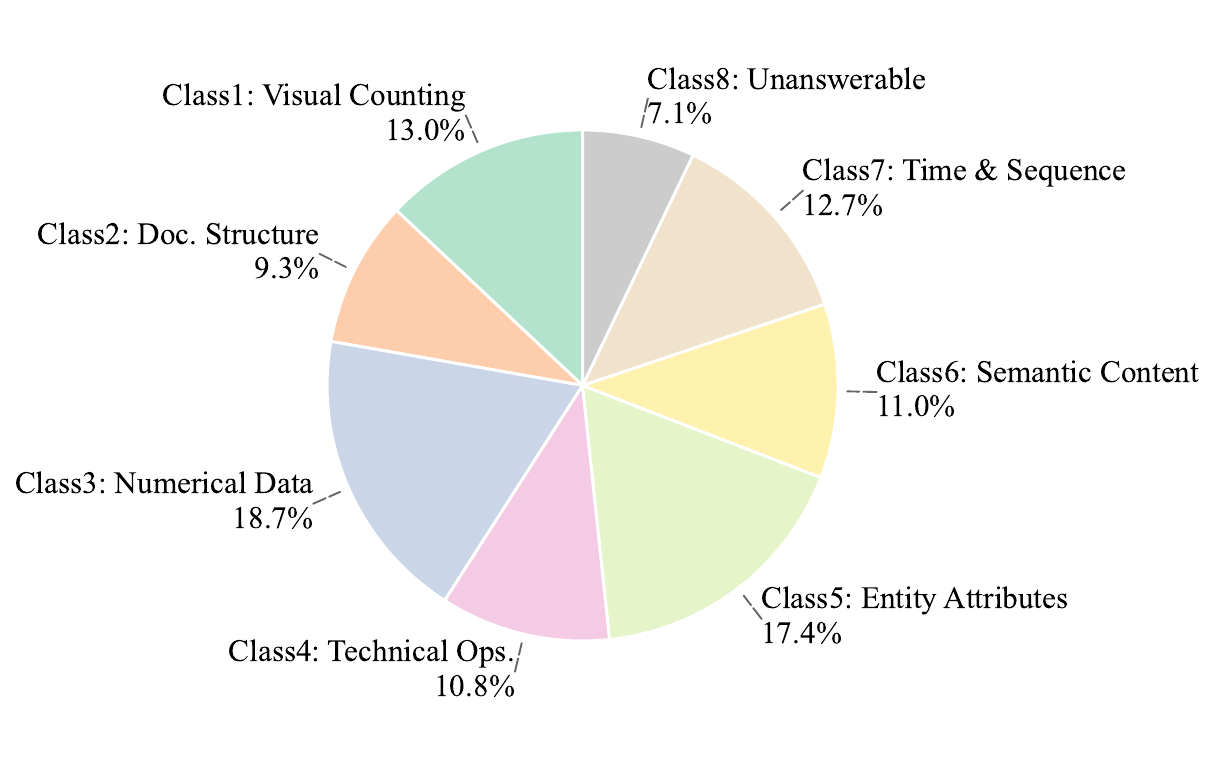}\\[-0.3em]
\scriptsize\textbf{(d)} Question category distribution in DocScope

\caption{Evidence and fact distributions in DocScope.}
\label{fig:dataset_evidence_fact_distributions}
\end{minipage}
\end{adjustbox}

\vspace{-0.3em}
\end{figure}

Tab.~\ref{tab:dataset_basic_statistics} summarizes the overall statistics of DocScope. The benchmark contains 1,124 questions derived from 273 documents, with an average of 4.12 questions per document. The source documents are long and information-rich, averaging 51.3 pages and 24,561 text tokens. The dataset is split into 730 test and 394 validation questions. Of the 1,124 questions, 649 require multi-page evidence, 397 can be answered from a single page, and 78 are unanswerable, together exercising localized reasoning, cross-page reasoning, and missing-information detection. As shown in Fig.~\ref{fig:dataset_evidence_fact_distributions}, the question category distribution in DocScope also remains balanced. Additional document-level distributions (page counts and text-token counts) are provided in Appendix~\ref{appendix:additional_stats}.

\subsection{Evaluation Protocol and Metrics}
\label{subsec:evaluation_protocol}

We evaluate each level of the structured output $y=(\mathcal{P},\mathcal{R},\mathcal{F},a)$ independently. A key design principle is \emph{inter-stage decoupling}: downstream stages are computed only on correctly retrieved pages $\hat{\mathcal{P}}=\mathcal{P}\cap\mathcal{P}^{*}$, so that page-localization errors do not cascade into region or fact metrics. The first three stages apply only to answerable questions; answer verification covers all questions including unanswerable ones. Formal metric definitions are provided in Appendix~\ref{appendix:metric_definitions}.

\paragraph{Page Localization.}
We report micro-averaged precision, recall, and F1 using exact page matching between predicted pages $\mathcal{P}$ and gold pages $\mathcal{P}^{*}$.

\paragraph{Region Grounding.}
A multimodal judge (GPT-5.5~\citep{gpt-5.5-0423-system-card}), selected via a human alignment study (Appendix~\ref{sec:analysis:bbox_alignment}; full judge prompts in Appendix~\ref{appendix:bbox_v4_prompt}), labels each gold region on correctly retrieved pages as \texttt{covered}, \texttt{imprecise}, or \texttt{not\_covered}. We report \emph{strict} F1 (counting only \texttt{covered}) and \emph{lenient} F1 (counting both \texttt{covered} and \texttt{imprecise}). All three LLM judges (region grounding, fact extraction, answer verification) are validated against human annotations in Appendix~\ref{appendix:judge_validation}.

\paragraph{Fact Extraction.}
A text-only judge (Qwen3.6-Plus), calibrated via a human alignment study (Appendix~\ref{sec:analysis:judge_alignment}), labels each extracted fact as \texttt{consistent} or not against the gold evidence on $\hat{\mathcal{P}}_q$. We report the micro-averaged consistency rate.

\paragraph{Answer Verification.}
A text-only judge (Qwen3.6-Plus), calibrated via a human alignment study (Appendix~\ref{app:analysis:judge_alignment_answer}), determines whether the predicted answer $a$ and the gold answer $a^{*}$ are semantically equivalent, tolerating minor surface variations while penalizing missing or incorrect key information.

\paragraph{Comparison with alternative scoring methods.}
For region grounding, the LLM judge achieves significantly higher human alignment than rule-based geometric metrics (Appendix~\ref{appendix:bbox_judge_vs_geom}). For answer verification, our judge improves AUROC by 0.135--0.150 over the MMLongBench-Doc pipeline~\citep{ma2024mmlongbenchdoc} with all $p<2.2\times10^{-5}$ (Appendix~\ref{app:compare_with_mmlongbench}).

\section{Evaluation}
\label{sec:evaluation}

\subsection{Experimental Setup}
\label{sec:evaluation:setup}

We evaluate a broad set of systems spanning four categories. \textbf{Proprietary models}: Gemini 3.1 Pro, Gemini 3.1 Flash Lite, Claude Opus 4.7, Claude Sonnet 4.6, GPT-5.4, and Qwen3.6 Plus. \textbf{Open-weight models}: Intern-S1-Pro and four widely adopted model families---Qwen3.5, Qwen3 VL, Gemma4, and Ministral3---with sizes ranging from 8B to 1T parameters (12 models). \textbf{Agentic RAG frameworks}: SimpleDoc and VidoRAG. \textbf{End-to-end document understanding models}: URaG and Docopilot. All domain-specific frameworks and models use the default inference settings reported in their original papers. Detailed configurations are provided in Appendix~\ref{appendix:experiment_environment}.

\subsection{Main Results}
\label{sec:evaluation:results}
\begin{table*}[t]
\centering
\fontsize{4.9pt}{5.6pt}\selectfont
\setlength{\tabcolsep}{1.25pt}
\renewcommand{\arraystretch}{0.88}
\caption{Main results on DocScope. \textbf{Params}: Total = total parameter count, Act.\ = activated parameters. \textbf{Page Localization}: P = precision, R = recall. \textbf{Fact}: Cons.\ = consistency rate. \textbf{ACC}: Ans.\ = answerable only, Unans.\ = unanswerable only, All = accuracy on all questions. Best and second-best scores across all visible rows are shown in \textbf{bold} and \underline{underlined}.}
\label{tab:main_results}
\resizebox{\linewidth}{!}{%
\begin{tabular}{lcc ccc cc c ccc}
\toprule
\textbf{Model} & \multicolumn{2}{c}{\textbf{Params}} & \multicolumn{3}{c}{\textbf{Page Localization}} & \multicolumn{2}{c}{\textbf{Region Grounding}} & \multicolumn{1}{c}{\textbf{Fact}} & \multicolumn{3}{c}{\textbf{ACC}} \\
\cmidrule(lr){2-3} \cmidrule(lr){4-6} \cmidrule(lr){7-8} \cmidrule(lr){9-9} \cmidrule(lr){10-12}
 & \textbf{Total} & \textbf{Act.} & \textbf{P} & \textbf{R} & \textbf{F1} & \textbf{Strict F1} & \textbf{Lenient F1} & \textbf{Cons.} & \textbf{Ans.} & \textbf{Unans.} & \textbf{All} \\
\midrule
\rowcolor{gray!10}
\multicolumn{12}{c}{\textit{Proprietary Models}} \\
Gemini 3.1 Pro~\citep{gemini31-pro} & -- & -- & \textbf{93.0} & 82.7 & \textbf{87.6} & 39.7 & 58.9 & 68.7 & \textbf{79.1} & 75.5 & \textbf{78.9} \\
Gemini 3.1 Flash Lite~\citep{gemini31-flash-lite} & -- & -- & 83.3 & 84.2 & 83.8 & 36.2 & 54.3 & 70.7 & 68.3 & 65.3 & 68.1 \\
Claude Opus 4.7~\citep{claude-opus-47} & -- & -- & 83.9 & \underline{88.3} & \underline{86.0} & 42.9 & \underline{63.0} & \textbf{77.4} & \underline{76.5} & 69.4 & \underline{76.0} \\
Claude Sonnet 4.6~\citep{claude-sonnet-46} & -- & -- & 78.8 & \textbf{90.1} & 84.0 & \underline{44.5} & 55.3 & 72.1 & 70.2 & 75.5 & 70.5 \\
GPT-5.4~\citep{gpt-5.4-0305-global} & -- & -- & \underline{85.6} & 72.4 & 78.4 & \textbf{57.4} & \textbf{71.9} & 59.4 & 64.0 & 44.9 & 62.7 \\
Qwen3.6 Plus~\citep{qwen36} & -- & -- & 73.3 & 85.4 & 78.9 & 29.6 & 45.5 & 66.4 & 67.0 & 65.3 & 66.8 \\
\midrule
\rowcolor{gray!10}
\multicolumn{12}{c}{\textit{Open-weight Models}} \\
Intern-S1-Pro~\citep{zou2026intern} & 1T & 22B & 25.2 & 8.8 & 13.0 & 16.0 & 29.7 & 38.8 & 34.4 & 42.9 & 34.9 \\
Qwen3.5-397B-A17B~\citep{qwen35} & 397B & 17B & 78.3 & 70.4 & 74.1 & 24.3 & 48.8 & 58.7 & 63.0 & 75.5 & 63.8 \\
Qwen3.5-122B-A10B~\citep{qwen35} & 122B & 10B & 79.1 & 42.2 & 55.0 & 24.2 & 42.9 & 52.6 & 62.6 & 77.5 & 63.6 \\
Qwen3.5-27B~\citep{qwen35} & 27B & 27B & 78.4 & 86.1 & 82.1 & 28.8 & 45.2 & 61.3 & 65.5 & 79.6 & 66.4 \\
Qwen3-VL-235B-A22B~\citep{bai2025qwen3} & 235B & 22B & 67.8 & 77.0 & 72.1 & 27.2 & 47.7 & 58.6 & 48.9 & \textbf{85.7} & 51.4 \\
Qwen3-VL-32B~\citep{bai2025qwen3} & 32B & 32B & 69.3 & 68.0 & 68.7 & 22.7 & 38.7 & 48.9 & 51.0 & 63.3 & 51.8 \\
Qwen3-VL-30B-A3B~\citep{bai2025qwen3} & 30B & 3B & 50.6 & 46.5 & 48.5 & 24.6 & 44.3 & 28.5 & 31.6 & 71.4 & 34.2 \\
Qwen3-VL-8B~\citep{bai2025qwen3} & 8B & 8B & 69.8 & 32.0 & 43.9 & 21.6 & 36.4 & 52.0 & 31.3 & 73.5 & 34.1 \\
Gemma-4-31B~\citep{gemma4} & 31B & 31B & 71.8 & 82.8 & 76.9 & 27.3 & 38.9 & \underline{73.6} & 59.0 & 65.3 & 59.4 \\
Gemma-4-26B-A4B~\citep{gemma4} & 26B & 4B & 50.8 & 63.4 & 56.4 & 7.5 & 9.2 & 53.1 & 30.9 & \underline{83.7} & 34.4 \\
Ministral3-14B~\citep{liu2026ministral} & 14B & 14B & 78.1 & 46.6 & 58.4 & 10.5 & 35.0 & 41.6 & 41.4 & 42.9 & 41.5 \\
Ministral3-8B~\citep{liu2026ministral} & 8B & 8B & 70.4 & 42.6 & 53.0 & 14.6 & 34.1 & 44.4 & 36.3 & 34.7 & 36.2 \\
\midrule
\rowcolor{gray!10}
\multicolumn{12}{c}{\textit{Document Understanding Frameworks (Agentic RAG)}} \\
SimpleDoc~\citep{jain2025simpledoc} & -- & -- & -- & -- & -- & -- & -- & -- & 51.2 & 40.8 & 50.5 \\
VidoRAG~\citep{wang2025vidorag} & -- & -- & -- & -- & -- & -- & -- & -- & 22.2 & 5.9 & 21.2 \\
\midrule
\rowcolor{gray!10}
\multicolumn{12}{c}{\textit{Document Understanding Specific Models (E2E)}} \\
URaG~\citep{shi2026urag} & 3B & 3B & -- & -- & -- & -- & -- & -- & 15.7 & 24.5 & 16.3 \\
Docopilot 8B~\citep{duan2025docopilot} & 8B & 8B & -- & -- & -- & -- & -- & -- & 6.5 & 20.4 & 7.4 \\
Docopilot 2B~\citep{duan2025docopilot} & 2B & 2B & -- & -- & -- & -- & -- & -- & 4.4 & 40.8 & 6.9 \\
\bottomrule
\end{tabular}%
}
\end{table*}

As shown in Tab.~\ref{tab:main_results}, producing correct answers to long-document questions remains challenging---only Gemini 3.1 Pro, Claude Opus 4.7, and Claude Sonnet 4.6 surpass 70\% accuracy. Yet the more striking finding is the gap between answer correctness and trajectory quality: even the highest-scoring model (Gemini 3.1 Pro, 78.9\% ACC) achieves only 39.7\% Strict Region F1 and 68.7\% Fact Consistency, indicating that a correct answer rarely comes with a fully verifiable evidence chain. Across all models, Region Grounding is the weakest stage of the trajectory, with Page F1 $\to$ Strict Region F1 drops exceeding 40 percentage points for most systems. Proprietary models substantially outperform both open-weight and domain-specialized frameworks or systems. Among the latter, Agentic RAG frameworks outperform end-to-end document-understanding models, suggesting that retrieval and iterative reasoning can partially compensate for deficiencies in long-context modeling. However, under our evaluation protocol, the tested domain-specific systems either struggle to produce effective structured reasoning trajectories or suffer substantial performance degradation when adapted to trajectory-oriented prompting, highlighting the need for future specialized systems to better support verifiable and traceable evidence chains.

Examining the trajectory stage by stage reveals that different models exhibit distinct capability profiles. GPT-5.4 achieves the strongest region grounding (Strict F1 57.4, Lenient F1 71.9) but the lowest Fact Consistency among proprietary models (59.4\%), suggesting it can locate evidence yet struggles to extract faithful factual representations. Conversely, Claude Opus 4.7 attains the highest Fact Consistency (77.4\%) with balanced page and region scores. Gemini 3.1 Pro leads in page localization (F1 87.6) and answer accuracy (78.9\%) but falls behind in region grounding (Strict F1 39.7), producing correct answers whose spatial evidence trail remains coarse. No model dominates all trajectory stages, indicating that each stage may be driven by different underlying capabilities.

Fine-grained analysis reveals a clear \emph{bottleneck effect}. Page localization is a necessary but insufficient condition for strong overall performance. For instance, Gemini 3.1 Flash Lite and Claude Opus 4.7 differ only slightly in Page Localization F1 (83.8 vs.\ 86.0), yet the former scores substantially lower in Region F1 and Fact Consistency, resulting in an 8-point accuracy gap. Likewise, Qwen3.5-122B-A10B and Gemma-4-26B-A4B achieve near-identical Page Localization and Fact Consistency scores but diverge markedly in final accuracy, highlighting that precise within-page evidence grounding is equally critical. Once page localization reaches a sufficient level, downstream grounding and fact extraction capabilities continue to determine the performance ceiling. Regarding model architecture, within the same family (e.g., Qwen3.5, Qwen3 VL, Gemma4), dense models consistently outperform their MoE counterparts despite the latter having several times more total parameters, suggesting that the number of activated parameters matters more than total scale for long-document trajectory construction. Finally, refusal ability on unanswerable questions is only weakly correlated with model size---open-weight models such as Qwen3.5, Qwen3 VL, and Gemma4 exhibit refusal capabilities comparable to, or exceeding, those of proprietary models---confirming that it should be treated as an independent evaluation dimension.

\section{Analysis and Discussion}

\label{sec:analysis}

The main results show that accuracy remains relatively low and cannot substitute for trajectory-level evaluation. We therefore analyze difficulty factors (Section~\ref{sec:analysis:evidence_distribution}), evidence-chain completeness (Section~\ref{sec:evidence_chain_completeness}), capability bottlenecks (Section~\ref{sec:analysis:oracle}), and stage-specific failure modes (Section~\ref{sec:error_analysis}).

\subsection{Evidence Page Distribution and Question Difficulty}
\label{sec:analysis:evidence_distribution}

\begin{figure*}[h]
    \centering
    \includegraphics[width=\textwidth]{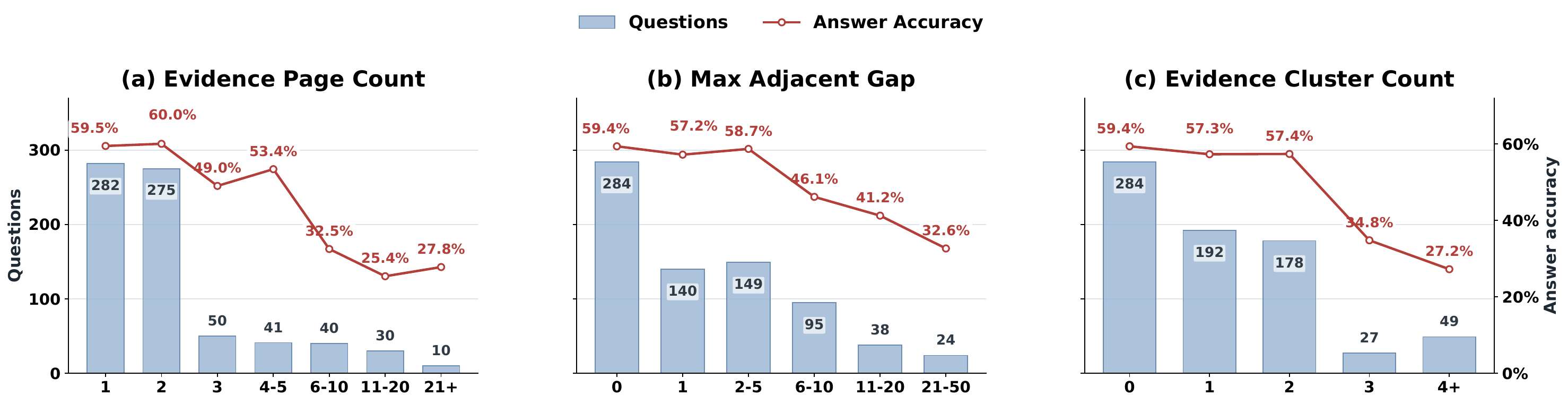}
    \caption{
    Relationship between evidence page distribution and answer accuracy.
    Bars denote the number of questions in each bin, while red lines denote answer accuracy.
    We analyze three factors: 
    (a) the number of ground-truth evidence pages,
    (b) the maximum adjacent gap between evidence pages, and
    (c) the number of separated evidence clusters.
    }
    \label{fig:evidence_factors_accuracy}
\end{figure*}

To understand what makes constructing a complete reasoning trajectory difficult, we analyze evidence-layout statistics from the ground-truth evidence pages and relate them to final answer accuracy.
As shown in Fig.~\ref{fig:evidence_factors_accuracy}(a), accuracy generally declines as the number of evidence pages increases, although the trend is not strictly monotonic, suggesting that page count is only a coarse indicator of difficulty.
In contrast, the spatial dispersion of evidence shows a clearer effect: accuracy drops substantially when adjacent evidence pages are separated by larger gaps (Fig.~\ref{fig:evidence_factors_accuracy}(b)) and when evidence is split into more disconnected clusters (Fig.~\ref{fig:evidence_factors_accuracy}(c)).
These results suggest that current models are challenged not only by the need to handle more evidence pages, but more importantly by the need to retrieve and integrate dispersed evidence across long contexts. More details are provided in Appendix~\ref{appendix:evidence_distribution_difficulty}.

\subsection{Evidence-Chain Completeness among Correct Answers}
\label{sec:evidence_chain_completeness}

\begin{wraptable}[10]{r}{0.4\linewidth}
    \centering
    \vspace{-1.0em}
    \captionsetup{width=\linewidth}
    \caption{Distribution of evidence-chain completeness among answer-correct samples, using GPT-5.4 as an example.}
    \label{tab:evidence_chain_correct_samples_gpt54}
    \small
    \setlength{\tabcolsep}{3pt}
    \renewcommand{\arraystretch}{0.98}
    \begin{tabular*}{\linewidth}{@{\extracolsep{\fill}}lr}
        \toprule
        Evidence-chain status & Ratio \\
        \midrule
        \textbf{Complete} & \textbf{29.06\%} \\
        Completely unreliable & 19.02\% \\
        Partially reliable & 51.92\% \\
        \bottomrule
    \end{tabular*}
    \vspace{-1.0em}
\end{wraptable}

The main-results analysis (Section~\ref{sec:evaluation:results}) and the difficulty analysis above both measure performance in aggregate. A complementary question is whether models that answer correctly also produce trustworthy evidence chains. As shown in Tab.~\ref{tab:evidence_chain_correct_samples_gpt54}, GPT-5.4---which exhibits the strongest region grounding capability among all evaluated models (Section~\ref{sec:evaluation:results})---attains the highest proportion of strictly complete evidence chains among answer-correct samples under this conditional distribution; even so, this proportion reaches only 29.06\%. The remaining answer-correct samples are predominantly only partially reliable, accounting for 51.92\%, while 19.02\% are completely unreliable. This result shows that even the frontier model under the evidence-chain completeness criterion often reaches correct final answers without fully reliable supporting evidence chains. The prevalence of ``answer-correct but process-unreliable'' samples demonstrates a clear decoupling between final-answer accuracy and evidence-chain reliability, confirming that relying solely on answer accuracy can substantially overestimate model reliability in verifiable long-document QA. Detailed per-model results are provided in Appendix~\ref{app:evidence_chain_completeness}.

\subsection{Oracle Evidence Access Study}
\label{sec:analysis:oracle}

The preceding analysis shows that correct answers frequently lack complete evidence chains. A follow-up question is: which stage of the trajectory constitutes the dominant bottleneck? We conduct an Oracle Evidence Access Study that supplies four models (Claude Sonnet 4.6, GPT-5.4, Qwen3-VL-235B-A22B, and Ministral3-8B) with gold evidence pages, gold regions, and gold atomic facts, incrementally removing the demands of each trajectory stage (setup details in Appendix~\ref{appendix:oracle_setup}).

\begin{figure*}[h!]
    \centering
    \includegraphics[width=\textwidth]{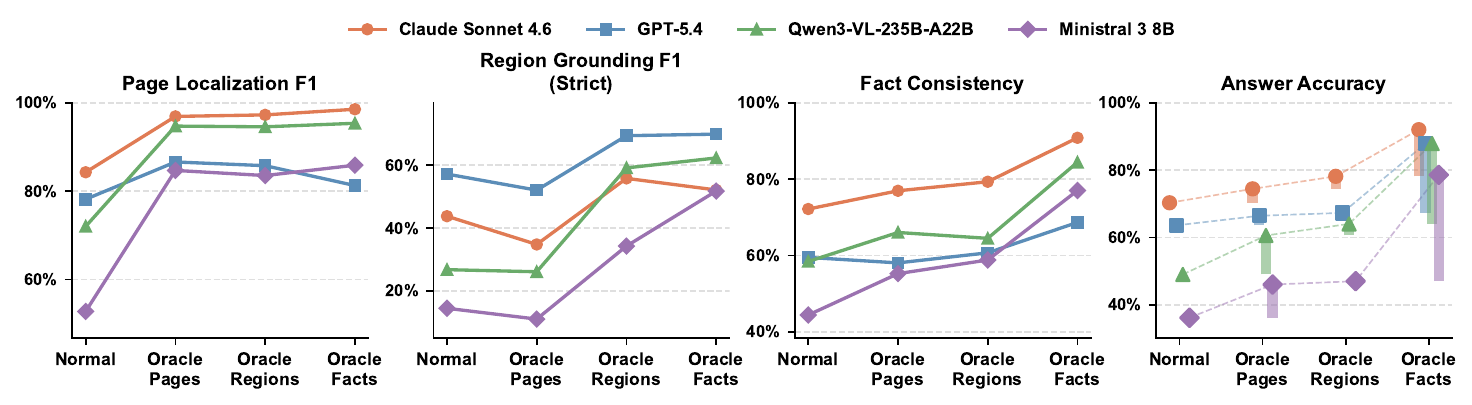}
    \caption{
        Oracle Evidence Access Study. Four trajectory metrics under the standard setting and three cumulative oracle settings for four models.
    }
    \label{fig:oracle_waterfall}
\end{figure*}

As shown in Fig.~\ref{fig:oracle_waterfall}, providing oracle facts yields the largest accuracy gain across all models, whereas oracle regions produce marginal improvements over oracle pages alone, identifying fact extraction---not region grounding---as the primary bottleneck. Even with full oracle fact access, substantial gaps persist (Claude Sonnet 4.6: 92.0\% vs.\ Ministral3-8B: 78.6\%), indicating that intrinsic reasoning capacity, rather than evidence access, bounds performance. Conversely, weaker open-weight models benefit disproportionately from oracle pages (e.g., Qwen3-VL-235B-A22B: $+$11.6\,pp vs.\ Claude Sonnet 4.6: $+$4.1\,pp), suggesting that long-context retrieval is a more binding constraint for smaller architectures. Extending the analysis to trajectory-level metrics reveals a counter-intuitive pattern: Strict Region F1 can decrease under oracle pages, as models shift from conservative large boxes to precise but mislocalized predictions---a \emph{conservative-to-aggressive strategy shift} documented with trajectory metric observations in Appendix~\ref{appendix:oracle_observations} and case studies in Appendix~\ref{appendix:oracle_grounding_case_study}.

\begin{figure}[t]
    \centering
    \includegraphics[width=\textwidth]{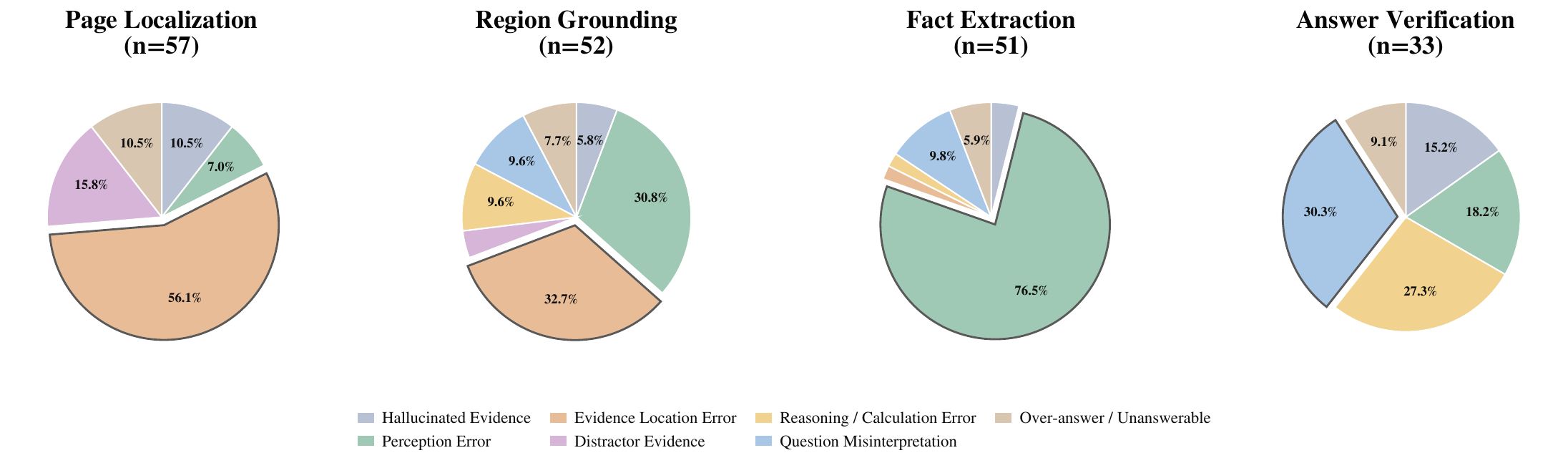}
    \caption{Error type distributions at different stages of DocScope.}
    \label{fig:error_analysis}
\end{figure}

\subsection{Error Analysis}
\label{sec:error_analysis}

We conducted a manual error analysis on nearly 200 erroneous samples from the four evaluated models, using the same set of models as in Section~\ref{sec:analysis:oracle}. The results reveal clear stage-specific failure patterns, as shown in Fig.~\ref{fig:error_analysis}. Errors in the Page Localization and Region Grounding stages are mainly caused by incomplete or incorrect evidence localization, showing that models still struggle to identify the precise supporting evidence in visually complex documents. When the evidence location is largely correct, Fact Extraction Stage errors are dominated by perception failures, such as misreading tables, charts, OCR text, or visual elements. In the Answer Verification Stage, errors mainly stem from misinterpretation and reasoning or calculation mistakes. Overall, these findings suggest that current MLLMs still face substantial challenges in producing correct and verifiable reasoning trajectories, requiring improvements in both fine-grained evidence grounding and factual reasoning. More details are provided in Appendix~\ref{app:error_taxonomy}.

\section{Related Work}
\label{sec:related_work}
\subsection{Benchmarks for Document Understanding}
\label{subsec:document_understanding_benchmarks}

Document-oriented benchmarks have evolved from single-page or format-specific settings, including document images, infographics, charts, and slides~\citep{mathew2021docvqa,mathew2022infographicvqa,masry2022chartqa,tanaka2023slidevqa}, to realistic multi-page document understanding scenarios such as MP-DocVQA~\citep{tito2023hierarchical}, DUDE~\citep{vanlandeghem2023dude}, MMLongBench-Doc, LongDocURL, and M-LongDoc~\citep{ma2024mmlongbenchdoc,deng2025longdocurl,chia2025mlongdoc}. Domain-specific efforts, including FinMMDocR~\citep{tang2026finmmdocr} and MMESGBench~\citep{zhang2025mmesgbench}, extend document understanding to vertical applications. However, these benchmarks evaluate answer correctness or coarse evidence use, leaving fine-grained, trajectory-level verifiability underexplored.

\subsection{Benchmarks for Multimodal Evidence Grounding}
\label{subsec:multimodal_evidence_grounding}

Multimodal evidence-grounding benchmarks such as MCiteBench~\citep{hu-etal-2025-mcitebench} and MAVIS~\citep{song2026mavis} introduce evidence attribution into multimodal tasks, but typically assume predefined or retrieved evidence pools rather than requiring models to navigate complete long documents. Recent benchmarks, including BBox-DocVQA, SIN-Bench, BRIDGE, and SciEGQA~\citep{yu2025bbox,xiang2026bridge,yu2026sciegqadatasetscientificevidencegrounded,ren2026sin}, move toward finer-grained grounding over bounding regions and atomic claims. Nevertheless, most of these efforts are confined to relatively short scientific papers or their evaluation of reasoning traces covers only limited aspects, leaving a gap in handling more general, longer documents with fine-grained verifiability and correctness analysis of reasoning trajectories. Our proposed \textbf{DocScope} is designed to bridge precisely this gap.

\subsection{Models for Document Understanding}
\label{subsec:document_understanding_models}

MLLM-based document understanding systems have made substantial progress in perceiving, reading, and reasoning. Existing approaches span OCR-dependent models such as LayTextLLM, DocVLM, and DocLayLLM~\citep{lu2025laytextllm,nacson2025docvlm,liao2025doclayllm}, OCR-free systems including TextMonkey, Mini-Monkey, mPLUG-DocOwl, DocPedia, Docopilot, URaG, TokenFD and DocSeeker~\citep{liu2026textmonkey,huang2024mini,hu2024mplug,hu2025mplug,feng2024docpedia,duan2025docopilot,shi2026urag,yan2026docseeker,guan2025tokenfd}, and agent-augmented RAG frameworks such as MDocAgent, SimpleDoc, and VidoRAG~\citep{han2025mdocagent,jain2025simpledoc,wang2025vidorag}. These advances have expanded the capability frontier of document QA.

\section{Conclusion}
\label{sec:conclusion}

In this work, we introduce DocScope, a benchmark for trustworthy long-document understanding through verifiable reasoning trajectories. Through multi-stage evaluation, DocScope offers a fine-grained diagnosis of how reliably a model's answers can be traced back to their source documents. Our analysis reveals several key challenges that merit future attention: achieving precise region grounding, aggregating evidence dispersed across lengthy documents, and faithfully perceiving and reasoning. The results further underscore the significant influence of model architecture. We hope DocScope serves as a diagnostic foundation for building document-understanding systems that are not only accurate but also verifiable and auditable.

\begin{ack}
This work was supported in part by the New Generation Artificial Intelligence-National Science and Technology Major Project (Grant No. 2025ZD0123602).
\end{ack}

\bibliographystyle{plainnat}
\bibliography{references}

\clearpage
\appendix

{\centering \Large \textbf{Appendix} \par}

\lstset{
    basicstyle=\small\ttfamily,
    breaklines=true,
    postbreak=\mbox{\textcolor{green}{$\hookrightarrow$}\space},
    frame=none,
    showstringspaces=false,
    tabsize=2,
    captionpos=b,
    breakatwhitespace=false,
    escapeinside={\%*}{*)},
    morekeywords={*,...}
}

\begingroup
\small
\startcontents[appendix]
\printcontents[appendix]{}{1}{\section*{Contents}}
\endgroup

\newpage

\section{Limitations}
\label{sec:limitation}

Although DocScope enables fine-grained evaluation of verifiable reasoning traces, annotation completeness remains challenging for long documents. Our sampling analysis (Appendix~\ref{appendix:gt_evidence_completeness}) suggests that the annotations cover sufficient evidence, but the complexity of long documents makes it difficult to guarantee a unique minimal evidence set. Given current MLLMs' hallucination and perceptual limitations, fine-grained human annotation remains a practical and reliable choice. Future work will scale the dataset and extend it to multilingual settings.

\section{Dataset Construction and Annotation Details}
\label{appendix:dataset}

\subsection{Dataset Construction Details}
\label{appendix:construction_details}

\paragraph{Data Collection.}
We source documents from the publicly available FinePDF corpus, applying metadata-based filters: 35--100 pages, English as the primary language, average text density above 80 tokens per page, and crawl dates later than 2025. This yields over 3{,}000 candidates. We then run PP-DocLayoutV3~\citep{paddle-ocr-15} for layout analysis, retaining documents with rich interleaved text, figures, and tables while excluding pages dominated by whitespace or sparse text, reducing the set to over 500. A final manual inspection removes low-resolution scans and overly specialized material (e.g., mathematics or chemistry papers), producing a pool of high-quality, visually rich documents.

\paragraph{Question Synthesis.}
For each document, we cluster page embeddings via jina-embeddings-v4~\citep{jinav4} and select the largest contiguous, information-dense segment as the context window. Claude-Opus-4.6~\citep{claude-opus46} is then prompted with the selected page images to produce diverse questions spanning visual recognition, structural extraction, numerical reasoning, entity comparison, semantic understanding, temporal reasoning, and unanswerable cases; the eight synthesis prompts are summarized in Appendix~\ref{appendix:question_synthesis_prompts}. Every answerable question is required to have a verifiable answer and to avoid explicit localization cues such as page numbers. After quality filtering and deduplication with Claude-Opus-4.6, we conduct multi-model blind testing with Claude-Opus-4.5, Claude-Opus-4.6, GPT-5.1, GPT-5.2, GPT-5.4, and Gemini-2.5-Pro~\citep{claude-opus-4-5,claude-opus46,gpt-51-1113-global,gpt-5.2-1211-global,gpt-5.4-0305-global,gemini-2.5-pro-06-17} to discard questions answerable without document access. Claude-Opus-4.6 is used in a final pass to refine question wording. In the end, we synthesized a total of 1,300 questions throughout the entire process, which were subsequently used for manual annotation and filtering.

\paragraph{Human Annotation and Quality Control.}
All annotation and verification are performed manually by 13 annotators recruited from two independent channels. Each annotator handles 100--120 questions. Annotations are then refined through a human-in-the-loop verification stage, in which annotators may consult outputs from Gemini-3.1-Flash-Lite~\citep{gemini31-flash-lite} to check and revise their work. Two senior members outside the annotation team serve as adjudicators and perform a final review of all questions and annotations; throughout the process, both annotators and adjudicators may revise question wording or flag low-quality samples for exclusion. In addition, during the adjudication process, we revised the answers to only 48 questions (3.7\% of the initially synthesized questions), indicating high annotation quality and strong consistency with human judgment. Finally, throughout the entire process, 135 questions were deemed overly specialized, incomprehensible, or unreasonable during the annotation and adjudication process and were excluded from the final Benchmark.

\paragraph{Personally Identifiable and Sensitive Information Control}
To reduce Personally Identifiable and Sensitive Information in the dataset, we directly removed inappropriate sensitive information appearing in questions or documents during data adjudication. After data annotation was completed, we adopted gemini-3.1-pro-preview for automated PII detection and filtering, which ultimately eliminated 41 questions(3\% of the initially synthesized questions) of the total data.

Overall, we initially synthesized 1300 questions. A total of 135 were excluded during annotation and adjudication, and another 41 were removed in the PII control process. The final benchmark contains 1124 questions.

\subsection{Question Synthesis Prompt Summaries}
\label{appendix:question_synthesis_prompts}

We design eight category-specific prompts to synthesize cross-page question-answer pairs from PDF page screenshots. Across all categories, the prompts require questions to be answerable from the provided document pages, avoid explicit page or region identifiers in the question text, and use semantic or visual descriptions for localization.

\begin{figure}[H]
    \centering
    \includegraphics[width=\linewidth]{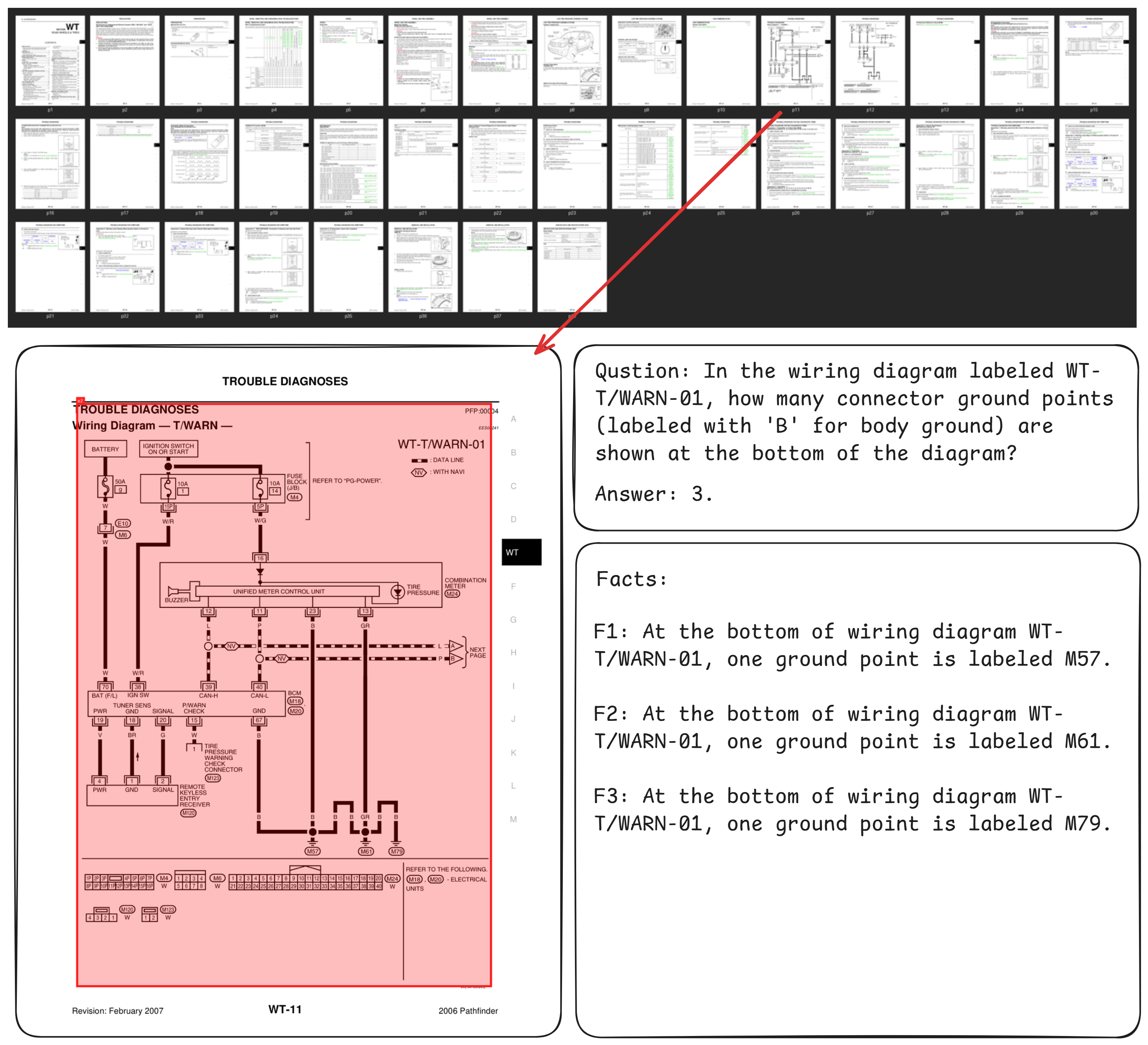}
    \caption{Example of Class 1: Visual Element Counting and Identification.}
    \label{fig:class_example_1}
\end{figure}

\paragraph{Class 1: Visual Element Counting and Identification.}
This prompt targets questions that require identifying, counting, comparing, or filtering visual elements across pages, such as photos, charts, icons, colors, shapes, people, or object categories. It emphasizes visually grounded localization and encourages diverse patterns, including cross-page counting, conditional filtering, visual-attribute association, visual comparison, and visual-text cross-reference. The detailed question-answer example is shown in Fig.~\ref{fig:class_example_1}.

\begin{figure}[H]
    \centering
    \includegraphics[width=\linewidth]{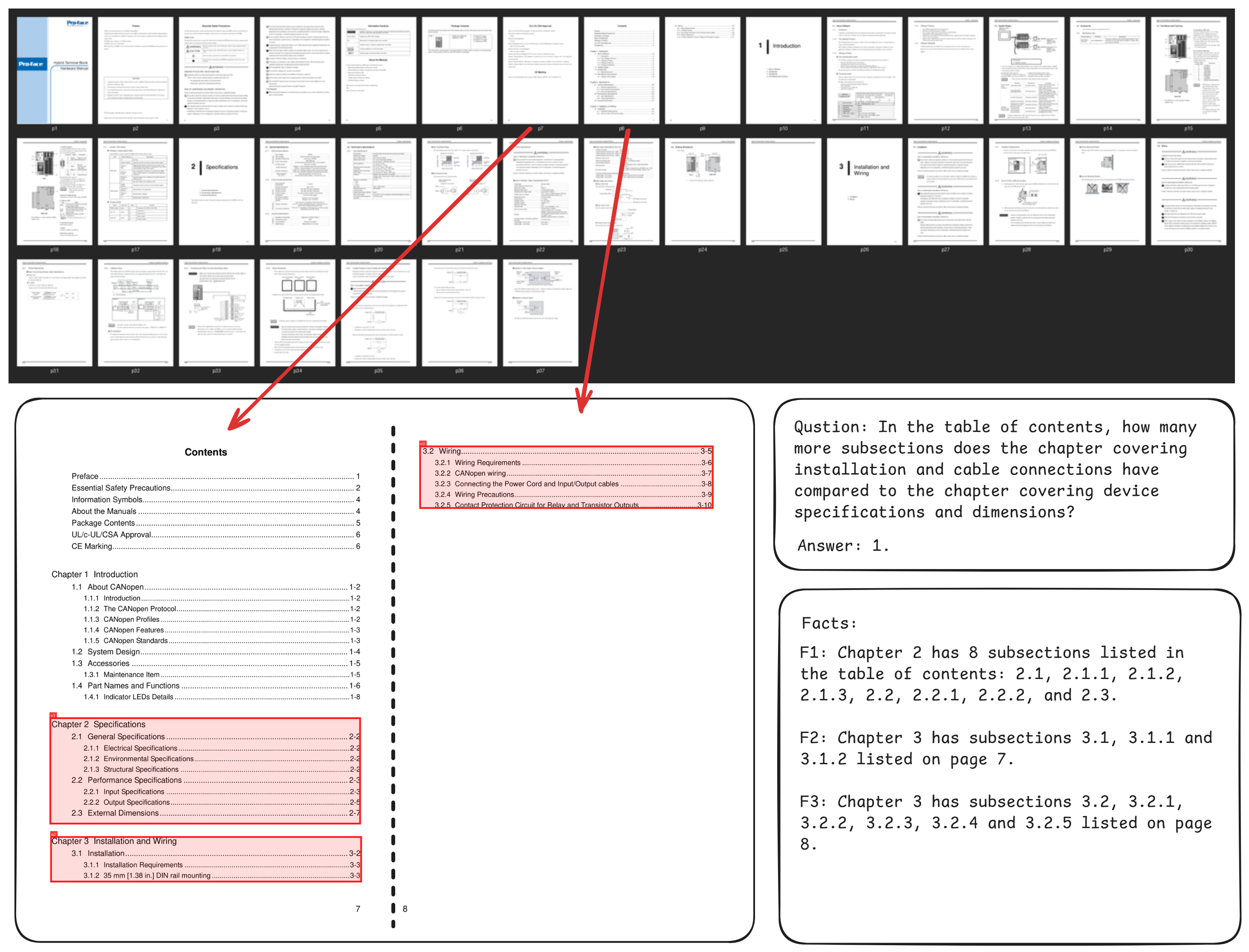}
    \caption{Example of Class 2: Document Structure and Metadata.}
    \label{fig:class_example_2}
\end{figure}

\paragraph{Class 2: Document Structure and Metadata.}
This prompt focuses on questions about document organization and structural metadata, such as sections, headings, tables, captions, lists, appendices, references, or recurring layout elements. The generated questions require models to connect structural cues across multiple pages rather than relying on a single local heading or table entry. The detailed question-answer example is shown in Fig.~\ref{fig:class_example_2}.

\begin{figure}[H]
    \centering
    \includegraphics[width=\linewidth]{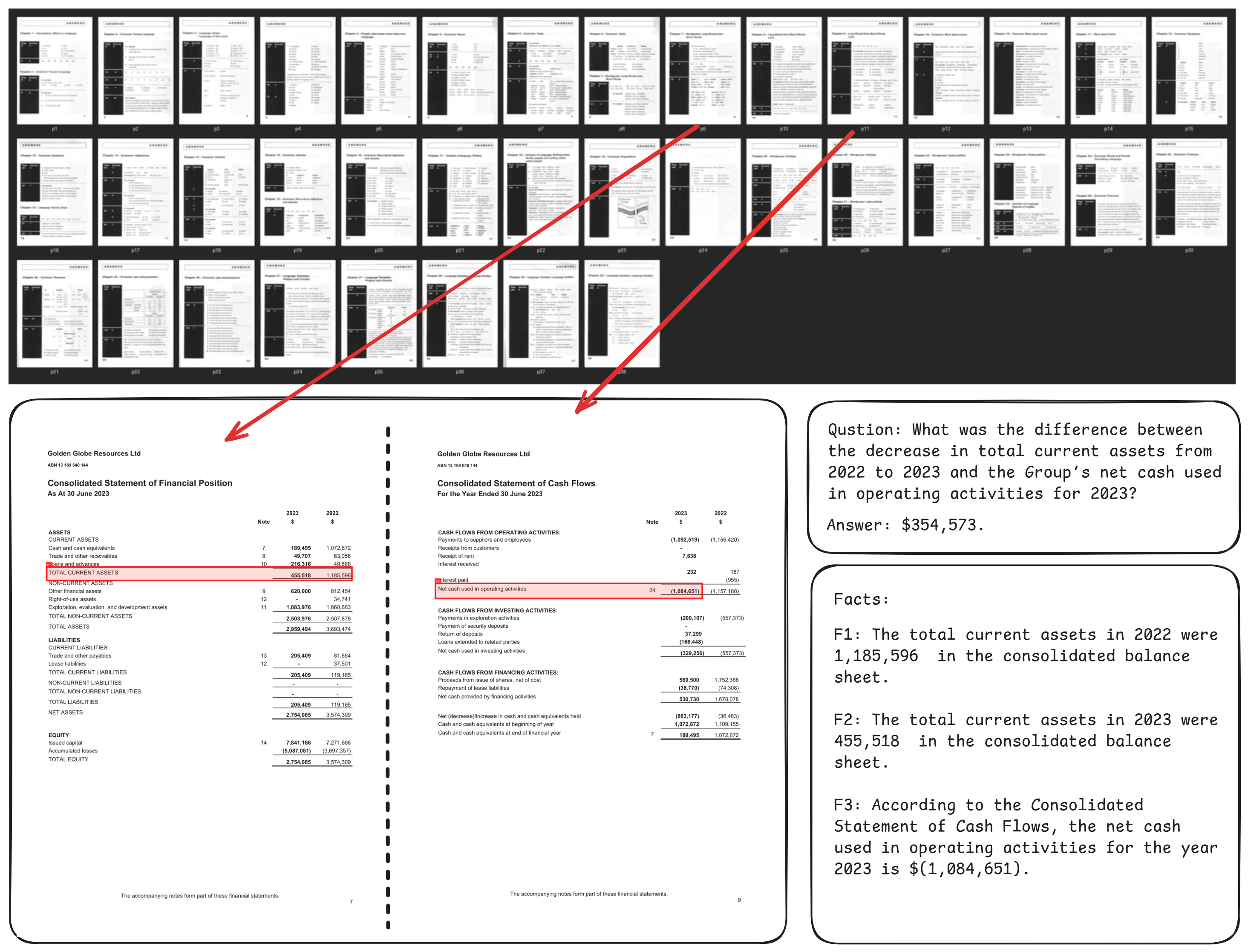}
    \caption{Example of Class 3: Numerical and Statistical Data.}
    \label{fig:class_example_3}
\end{figure}

\paragraph{Class 3: Numerical and Statistical Data.}
This prompt synthesizes questions involving numerical values, tables, charts, rankings, percentages, measurements, or statistical comparisons distributed across pages. It encourages operations such as locating relevant quantities, comparing values, aggregating evidence, matching numbers across visual and textual regions, and deriving concise numerical or categorical answers. The detailed question-answer example is shown in Fig.~\ref{fig:class_example_3}.

\begin{figure}[H]
    \centering
    \includegraphics[width=\linewidth]{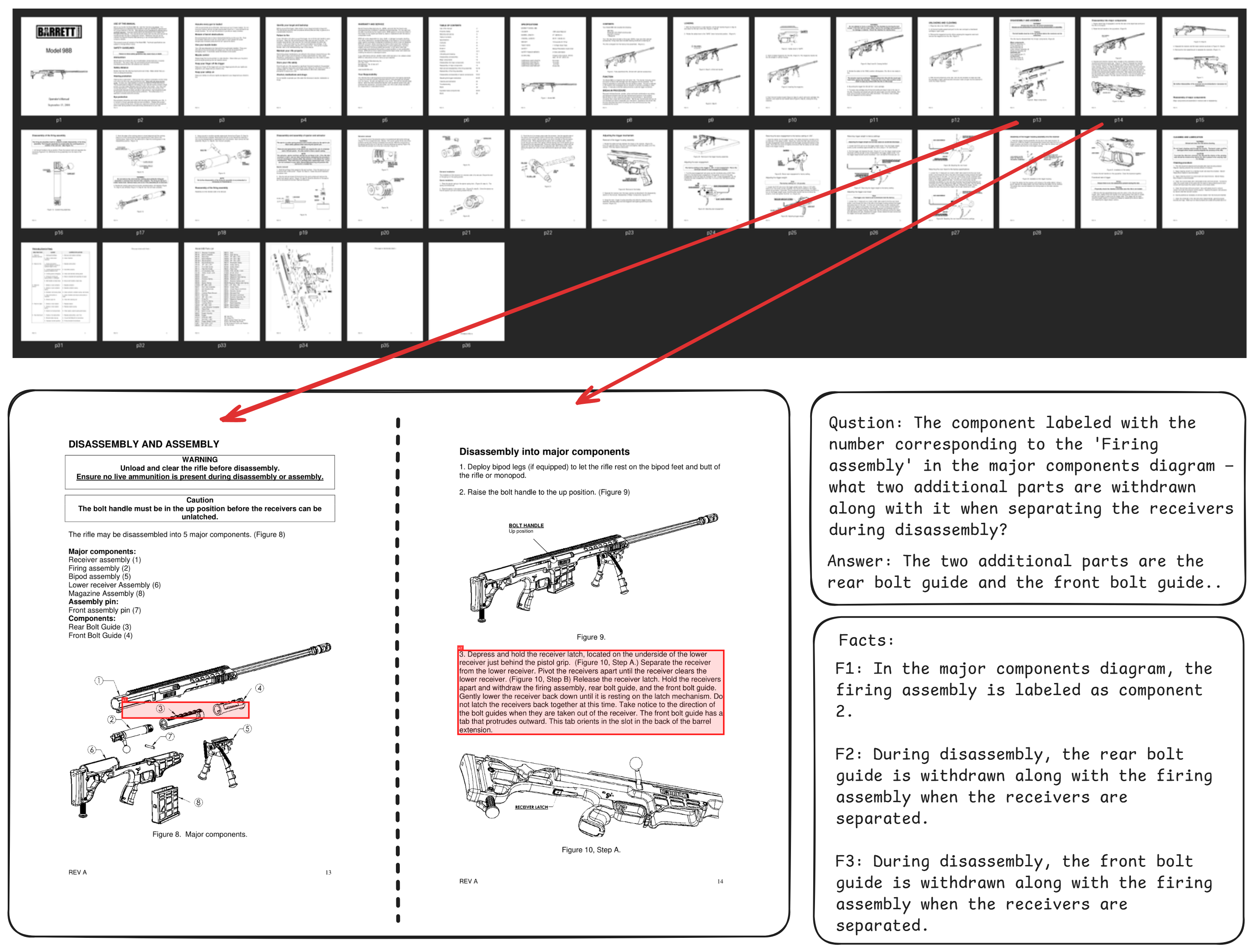}
    \caption{Example of Class 4: Technical Systems and Operational Procedures.}
    \label{fig:class_example_4}
\end{figure}

\paragraph{Class 4: Technical Systems and Operational Procedures.}
This prompt targets questions about technical mechanisms, system components, workflows, procedures, operating steps, or cause-effect relations described across pages. The resulting questions require integrating diagrams, process descriptions, instructions, and explanatory text to recover how a system works or how a procedure should be executed. The detailed question-answer example is shown in Fig.~\ref{fig:class_example_4}.

\begin{figure}[H]
    \centering
    \includegraphics[width=\linewidth]{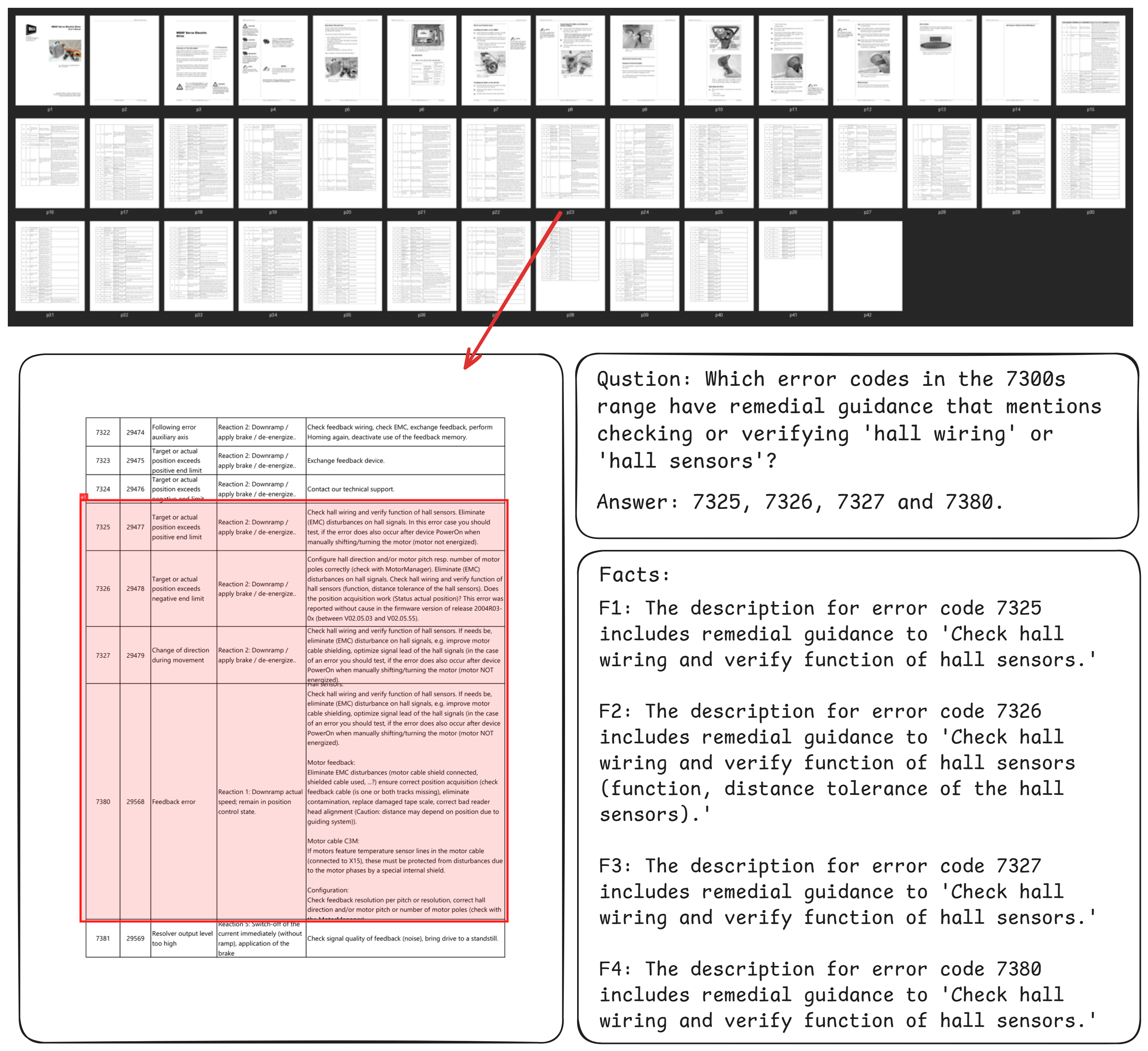}
    \caption{Example of Class 5: Entity Attributes and Comparative Relationships.}
    \label{fig:class_example_5}
\end{figure}

\paragraph{Class 5: Entity Attributes and Comparative Relationships.}
This prompt generates questions about entities and their attributes, including organizations, products, people, locations, methods, or other named objects. It emphasizes cross-page comparison, attribute matching, relation extraction, and distinguishing entities that share similar descriptions but differ in specific properties. The detailed question-answer example is shown in Fig.~\ref{fig:class_example_5}.

\begin{figure}[H]
    \centering
    \includegraphics[width=\linewidth]{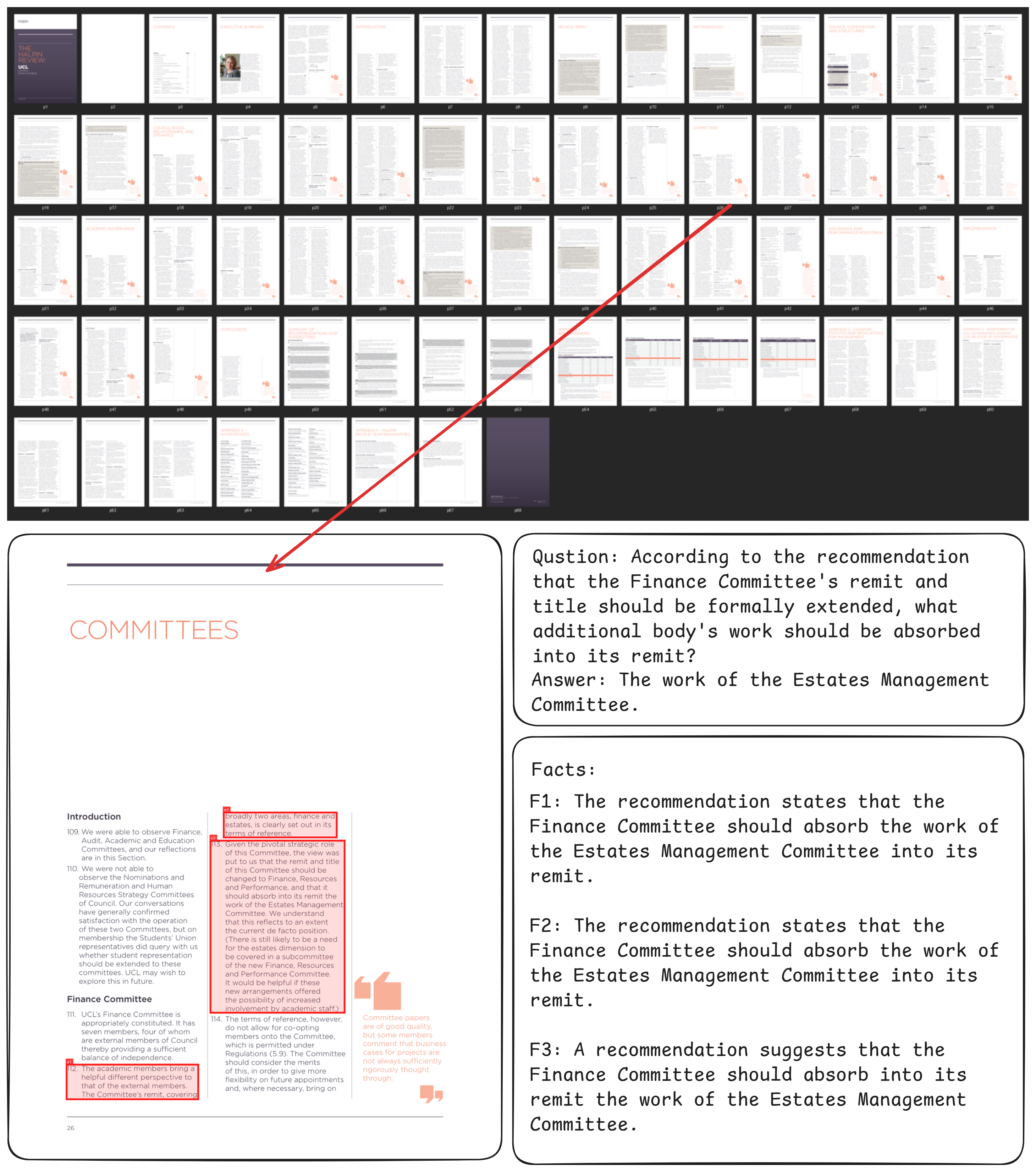}
    \caption{Example of Class 6: Semantic Content and Conceptual Meaning.}
    \label{fig:class_example_6}
\end{figure}

\paragraph{Class 6: Semantic Content and Conceptual Meaning.}
This prompt focuses on high-level semantic understanding, including definitions, conceptual relationships, claims, themes, explanations, and implications that are distributed across the document. Questions in this category require synthesizing textual and visual evidence to infer the intended meaning rather than extracting a single surface phrase. The detailed question-answer example is shown in Fig.~\ref{fig:class_example_6}.

\begin{figure}[H]
    \centering
    \includegraphics[width=\linewidth]{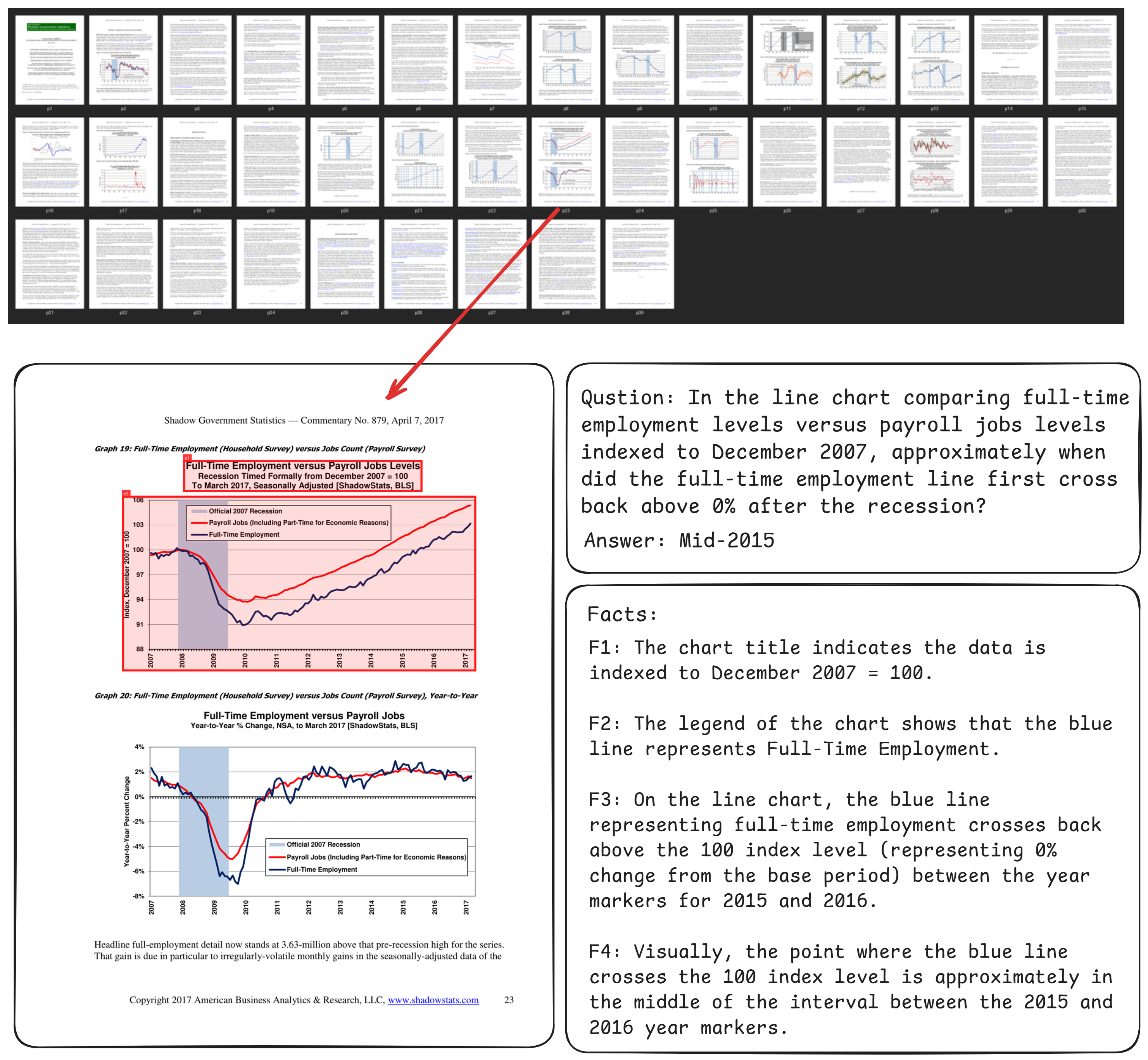}
    \caption{Example of Class 7: Time, Date, and Sequential Relationships.}
    \label{fig:class_example_7}
\end{figure}

\paragraph{Class 7: Time, Date, and Sequential Relationships.}
This prompt targets temporal reasoning and ordered relationships, including dates, timelines, stages, historical sequences, deadlines, versions, or event ordering. It requires models to locate time-related evidence across pages and reason about chronological order, duration, precedence, or temporal correspondence. The detailed question-answer example is shown in Fig.~\ref{fig:class_example_7}.

\begin{figure}[H]
    \centering
    \includegraphics[width=\linewidth]{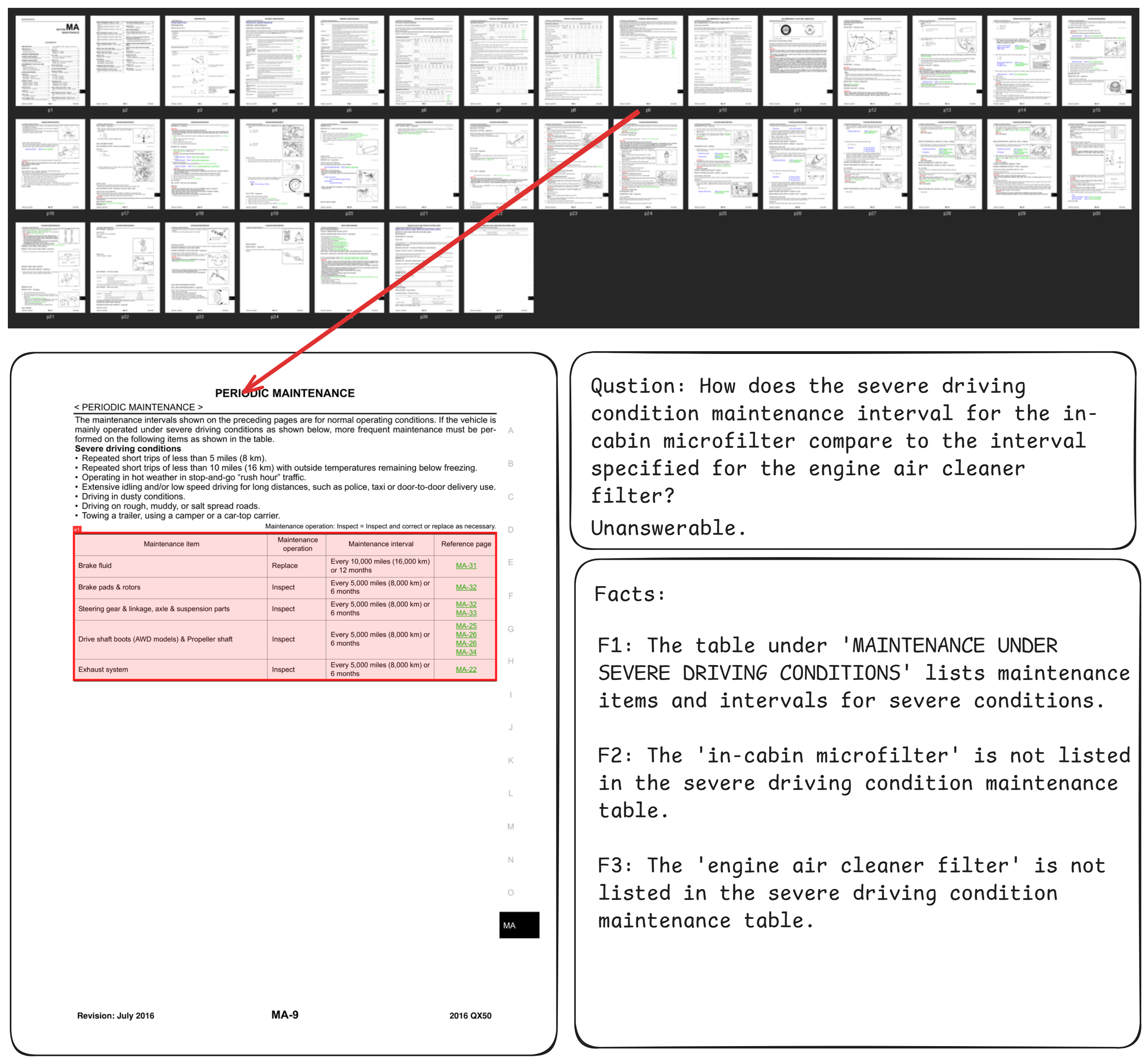}
    \caption{Example of Class 8: Unanswerable Questions.}
    \label{fig:class_example_8}
\end{figure}

\paragraph{Class 8: Unanswerable Questions.}
This prompt synthesizes questions that appear plausible given the document domain but cannot be answered from the provided pages. It requires the question to be closely related to the document while missing necessary evidence, so that a model must recognize insufficiency rather than hallucinate an unsupported answer. The detailed question-answer example is shown in Fig.~\ref{fig:class_example_8}.

\clearpage

\subsection{Distribution Analysis of Synthetic Questions}

Since the questions in our benchmark are synthetically generated, a natural concern is whether they exhibit the common pitfalls of LLM-synthesized data, such as limited diversity or distributional divergence from real-world data. To address this, we conduct a distributional similarity analysis between DocScope and MMLongBench-Doc, whose questions are crafted by human experts after reading the documents and can thus serve as a reasonable proxy for authentic questions posed in long-document scenarios.

\begin{figure}[H]
    \centering
    \includegraphics[width=\linewidth]{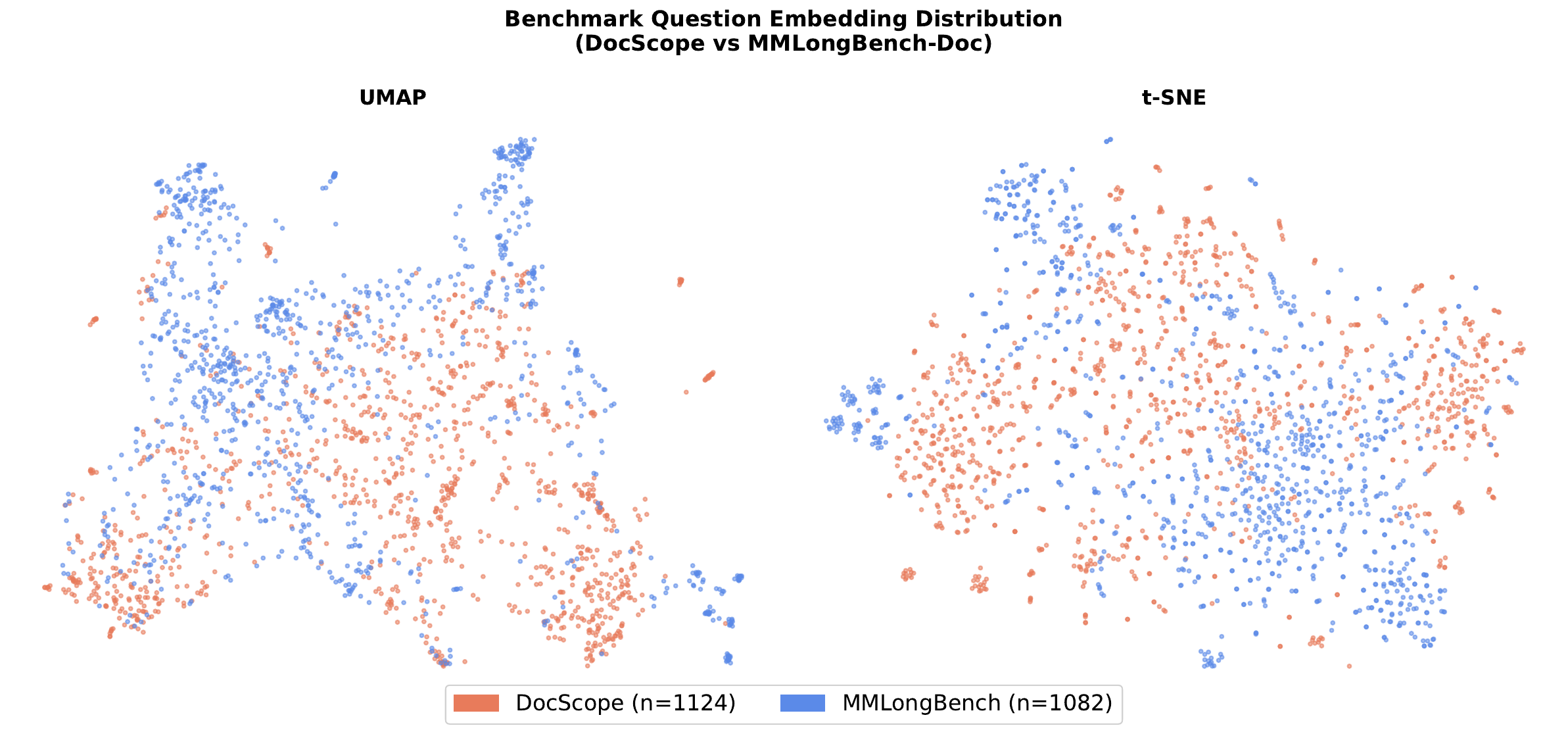}
    \caption{UMAP and t-SNE visualization of DocScope and MMLongBench-Doc embeddings.}
    \label{appendix:fig:embedding_viz}
\end{figure}

Specifically, we embed both sets of questions using the Qwen3 text-embedding-v4 model~\citep{zhang2025qwen3}. As shown in Fig.~\ref{appendix:fig:embedding_viz}, the two-dimensional projections obtained via UMAP and t-SNE reveal substantial overlap between the question embeddings of DocScope and MMLongBench-Doc, indicating that the two benchmarks cover similar regions in the semantic space.

To further quantify distributional similarity, we report three complementary metrics: centroid cosine similarity, Maximum Mean Discrepancy (MMD), and Fréchet Distance. As shown in Tab.~\ref{tab:cosine_similarity}, the centroid cosine similarity between the two benchmarks is 0.9351, indicating strong alignment in overall semantic orientation. Meanwhile, the intra-set mean pairwise cosine similarities of DocScope and MMLongBench-Doc are 0.2369 and 0.2391, respectively, suggesting comparable levels of semantic diversity. Furthermore, as reported in Tab.~\ref{tab:mmd_multiscale} and~\ref{tab:frechet_distance}, the low multi-scale MMD and PCA-based Fréchet Distance further confirm that the distributional discrepancy between the two datasets is small in the representation space.

\begin{table}[H]
\centering
\caption{Cosine similarity between DocScope and MMLongBench-Doc embeddings.}
\label{tab:cosine_similarity}
\begin{tabular}{lc}
\toprule
\textbf{Metric} & \textbf{Value} \\
\midrule
Centroid cosine similarity & 0.9351 \\
Average intra-dataset cosine similarity of DocScope & 0.2369 \\
Average intra-dataset cosine similarity of MMLongBench-Doc & 0.2391 \\
\bottomrule
\end{tabular}
\end{table}

\begin{table}[H]
\centering
\caption{Multi-scale RBF-kernel MMD between DocScope and MMLongBench-Doc embeddings.}
\label{tab:mmd_multiscale}
\begin{tabular}{ccc}
\toprule
\textbf{RBF kernel scale $\sigma$} & \textbf{MMD$^2$} & \textbf{MMD} \\
\midrule
0.1 & 0.000017 & 0.0042 \\
0.5 & 0.008191 & 0.0905 \\
1.0 & 0.015226 & 0.1234 \\
2.0 & 0.006433 & 0.0802 \\
5.0 & 0.001194 & 0.0346 \\
\midrule
Average & 0.006212 & 0.0788 \\
\bottomrule
\end{tabular}
\end{table}

\begin{table}[H]
\centering
\caption{Fréchet distance between DocScope and MMLongBench-Doc embeddings after PCA.}
\label{tab:frechet_distance}
\begin{tabular}{cc}
\toprule
\textbf{PCA dimensions} & \textbf{Fréchet Distance} \\
\midrule
64  & 0.1063 \\
128 & 0.1470 \\
256 & 0.2101 \\
\bottomrule
\end{tabular}
\end{table}

\subsection{Annotation Details}
\label{appendix:annotation_details}

\begin{figure}[h]
    \centering
    \includegraphics[width=0.9\linewidth]{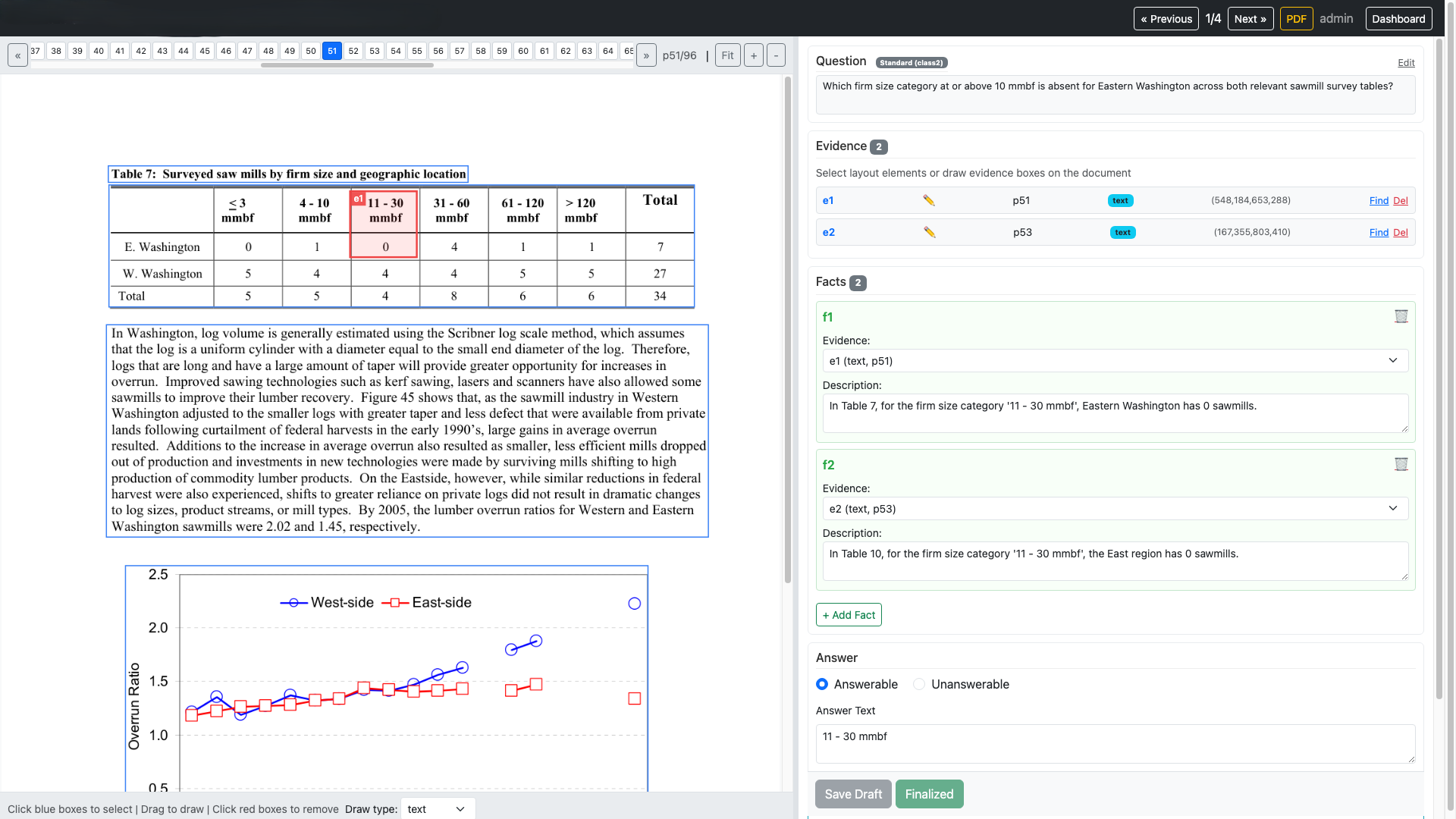}
    \caption{Annotation interface used in DocScope.}
    \label{appendix:fig:platform_part1}
\end{figure}

\begin{figure}[h]
    \centering
    \includegraphics[width=0.9\linewidth]{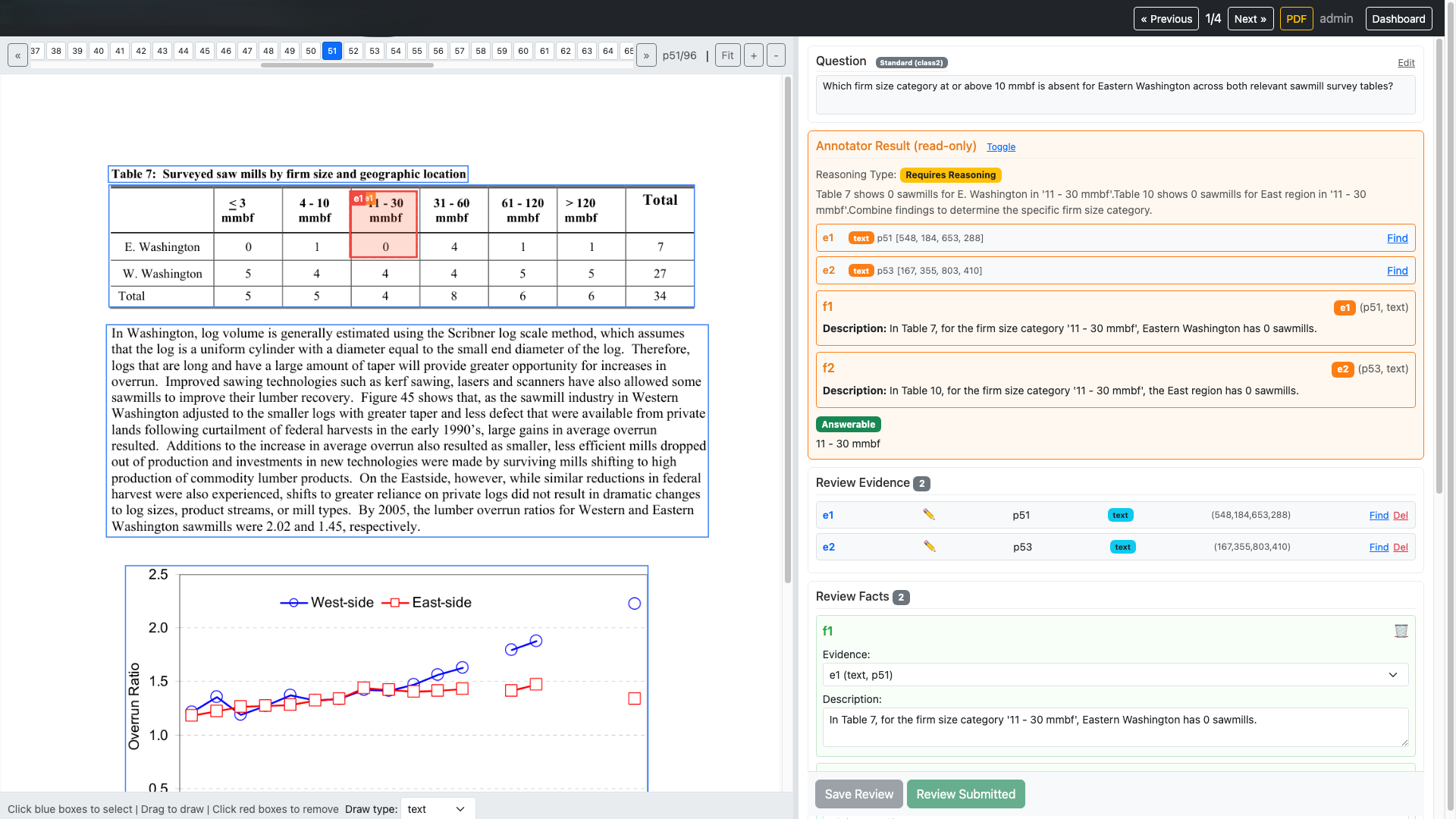}
    \caption{Adjudication interface used in DocScope.}
    \label{appendix:fig:platform_part2}
\end{figure}

The annotation was conducted by a team of 13 trained annotators from two independent channels. Annotators were compensated either through regular employment arrangements or at a rate no lower than the applicable local minimum wage. Regarding working hours and human effort, each annotator received detailed annotation training before being allowed to begin the annotation task. Subsequently, annotators worked on a half-time basis. Specifically, each annotator worked an average of 4 hours per day for 5 days. The annotation stage alone therefore required a total of 260 person-hours. The annotation process was carried out on a dedicated web-based platform. Using this platform, annotators systematically selected pages relevant to each question, highlighted supporting evidence spans, annotated fact-bearing statements, and provided the final answers. Fig.~\ref{appendix:fig:platform_part1} and Fig.~\ref{appendix:fig:platform_part2} show the annotation interface and adjudication review interface used in the DocScope system. These interfaces support document browsing, page-level selection, evidence localization, fact-bearing statement annotation, answer entry, and related functions.

\subsection{Validating Ground-Truth Evidence Completeness}
\label{appendix:gt_evidence_completeness}

\begin{figure}[H]
    \centering
    \includegraphics[width=0.9\linewidth]{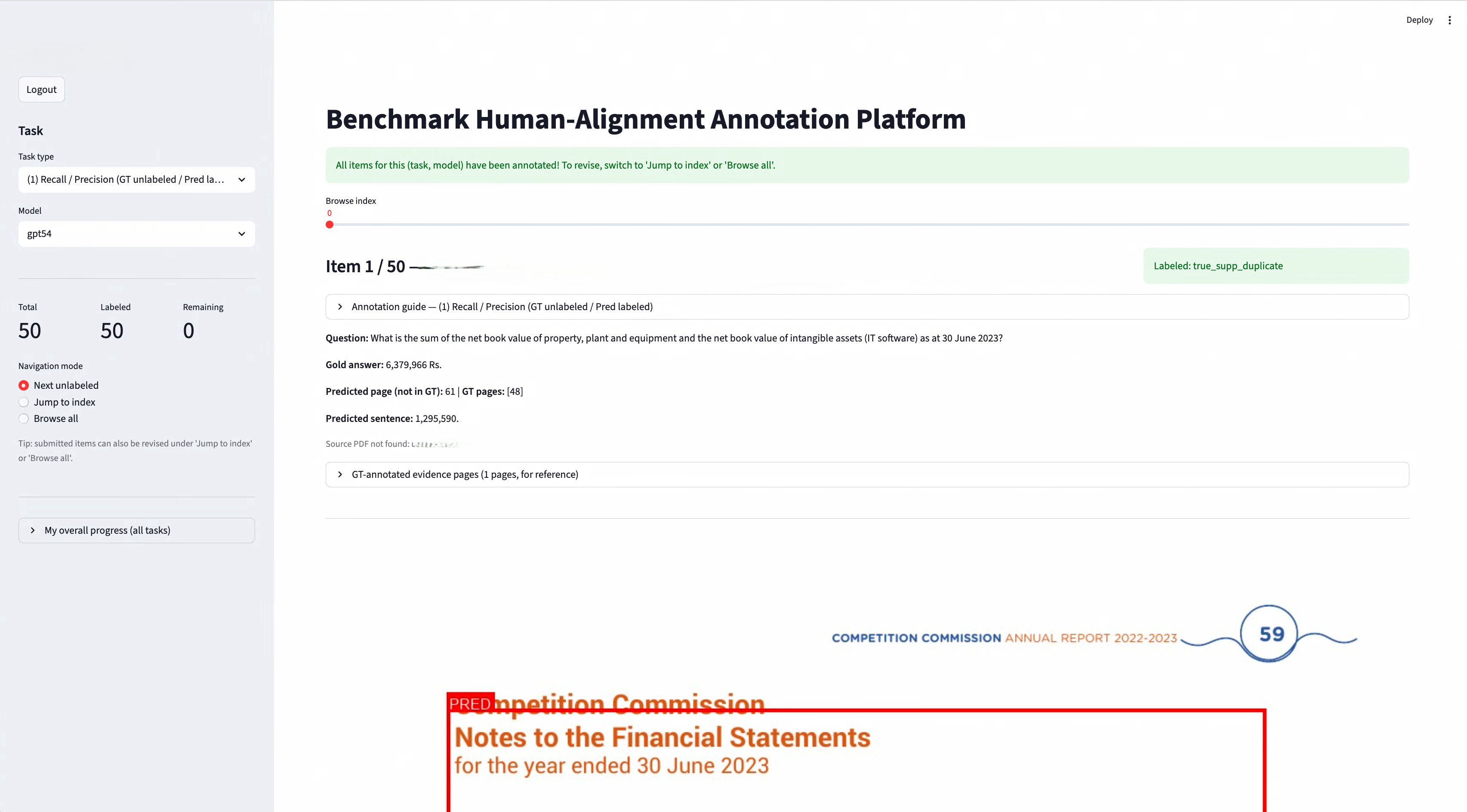}
    \caption{Annotation platform for the human alignment of ground-truth evidence completeness.}
    \label{fig:gt_evidence_completeness_platform}
\end{figure}

To verify the completeness of evidence annotations in our benchmark, we randomly inspect benchmark samples and conduct human review to determine whether the evidence annotations of each instance miss any information that could affect the final answer. If a large number of samples contain missing evidence, the validity of benchmark evaluation may be compromised. We recruit four professional annotators and evaluate 150 instances in total, sampled from the inference results of three models, with 50 distinct instances per model. Each instance is classified into one of four categories: \textit{Required Missing}, where the page contains indispensable evidence omitted from the benchmark annotations; \textit{Duplicate Evidence}, where the page can also support the answer but the existing evidence annotations are already sufficient; \textit{Tangential}, where the page is related to the question but does not directly support the answer; and \textit{Wrong Fact}, where the model incorrectly treats unsupported or contradictory page content as evidence. Fig.~\ref{fig:gt_evidence_completeness_platform} shows the annotation platform used for this human review.

As shown in Tab.~\ref{tab:gt_evidence_completeness}, only 2.67\% of the inspected samples are labeled as required missing evidence, while the majority are tangential mentions or model hallucinations. This indicates that missing key evidence pages are rare in our benchmark, and their likelihood of affecting evaluation validity is low.

\begin{table}[H]
\centering
\caption{
Human verification results for ground-truth evidence completeness. Only \textit{Required Missing} indicates a true omission in the ground-truth evidence annotations.
}
\label{tab:gt_evidence_completeness}
\vspace{0.5em}
\small
\setlength{\tabcolsep}{7pt}
\renewcommand{\arraystretch}{1.15}
\begin{tabular}{lcccc}
\toprule
\textbf{Model} &
\textbf{Wrong Fact} &
\textbf{Tangential} &
\textbf{Required Missing} &
\textbf{Duplicate Evidence} \\
\midrule
Gemma4-26B    & 9  & 38 & 0 & 3 \\
GPT-5.4       & 10 & 29 & 4 & 7 \\
Qwen3.5-397B  & 2  & 42 & 0 & 6 \\
\midrule
\textbf{Overall} &
\textbf{14.00\%} &
\textbf{72.67\%} &
\textbf{2.67\%} &
\textbf{10.67\%} \\
\bottomrule
\end{tabular}
\end{table}

\subsection{Additional Statistics of DocScope}
\label{appendix:additional_stats}

\begin{figure}[H]
\centering
\begin{minipage}[t]{0.48\linewidth}
\centering
\includegraphics[width=\linewidth]{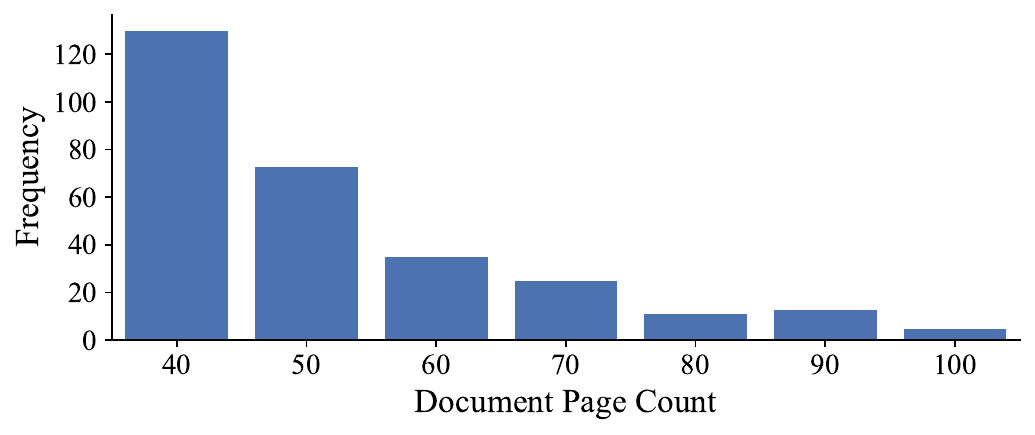}\\[-0.3em]
\scriptsize\textbf{(a)} Document page-count distribution
\end{minipage}\hfill
\begin{minipage}[t]{0.48\linewidth}
\centering
\includegraphics[width=\linewidth]{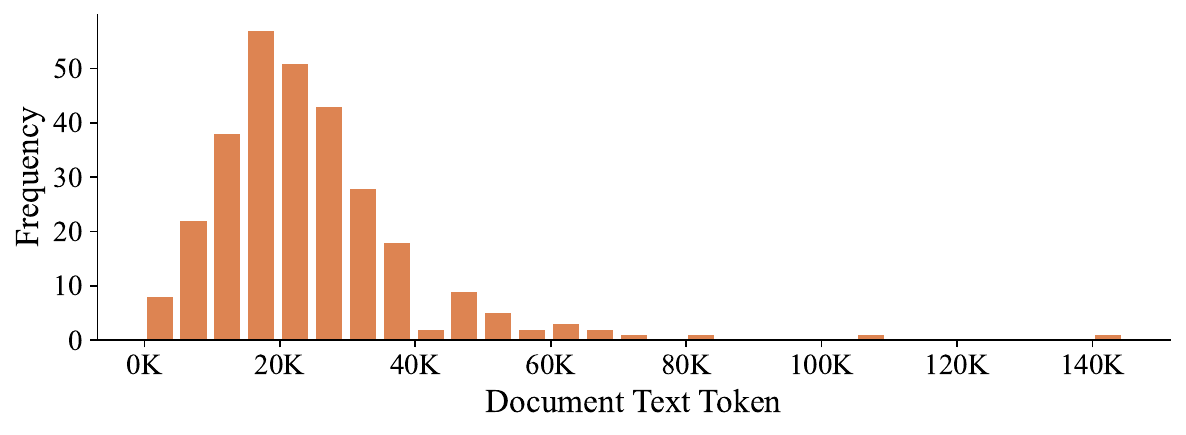}\\[-0.3em]
\scriptsize\textbf{(b)} Document text-token distribution
\end{minipage}
\caption{Additional document-level distributions in DocScope. (a) Distribution of document page counts. (b) Distribution of document text-token counts.}
\label{fig:appendix_document_distributions}
\end{figure}

In addition to the statistical information already provided in the main text, we further analyze the distribution of document pages and the number of text tokens contained therein. As shown in Fig.~\ref{fig:appendix_document_distributions}, most documents have between 40 and 50 pages, while a considerable number of documents exceed 70 pages. Moreover, the text token count of the majority of documents centers around 20K tokens, and the distribution exhibits a distinct long-tail feature, indicating that some documents have relatively high text density.

\section{Task and Evaluation Protocol Details}
\label{appendix:task_evaluation_protoca_detail}

\subsection{Inference Prompt}
\label{appendix:inference_prompt}
For completeness and reproducibility, we present below the inference prompt used in the DocScope, which defines the model's document-question-answering task, citation requirements, page-numbering rules, and output protocol.
\begin{tcolorbox}[breakable,enhanced jigsaw,colback=green!5!white,colframe=green!75!black,title=Document QA Citation Prompt]
\textbf{USER PROMPT}
\lstset{basicstyle=\ttfamily\scriptsize}
\begin{lstlisting}[breaklines=true]
You are an expert document QA system. You are given page images of a PDF document and a question about the document. Each image is wrapped by text markers indicating its global page number.

<hard_constraints>
BEFORE YOU ANSWER, INTERNALIZE THIS CONTRACT. VIOLATING ANY RULE MEANS COMPLETE FAILURE:

1. ZERO TOLERANCE FOR MISSING CITATIONS: EVERY SINGLE sentence stating a fact from the document MUST end with EXACTLY ONE formal citation. No exceptions. No excuses.
2. CITATION FORMAT: The citation MUST strictly match this format: `[page=N, doc_page="...", bbox=[x1, y1, x2, y2]]`
3. SYNTAX ALERT: Pay close attention to the closing brackets. The citation MUST end with TWO right brackets `]]` and then the period. (Correct: `0.512]].` / Incorrect: `0.512].`)
4. FORBIDDEN: NEVER use natural language to cite pages (e.g., DO NOT write "on page 5", "in Table 7 on global page 51", or "image 58"). You MUST use the bracket format.
5. MANDATORY FINAL ANSWER TAG: You MUST conclude your response with a concise final answer wrapped STRICTLY and EXACTLY as:
   `<answer> your final answer </answer>`
6. NO MARKDOWN: Do not use headings, bold, italics, lists, or tables in your reasoning or answer. Plain prose only (fenced code blocks and LaTeX math are allowed when strictly necessary).
</hard_constraints>

## Page Numbering Rule
- Each page image is preceded and followed by a text marker (e.g., `=== GLOBAL PAGE X (start) ===`). `X` is the **global page number** (integer, starting from 1).
- The document itself may print its own page number inside the page (e.g., "12", "iv"). This is the **document page number**.
- The `page` field in your citation MUST use the **global page number**.
- The `doc_page` field MUST use the **document page number** as a string. If not visible, set `doc_page="none"`.

## CRITICAL Citation Requirement
- **Fact vs. Reasoning**: A sentence is "fact-bearing" if it contains a number, date, name, quote, or claim from the document. A sentence is "pure reasoning" ONLY if it is arithmetic/logical inference based on already-cited facts. EVERY fact-bearing sentence requires a citation.
- **Writing Rule (One Citation Per Sentence)**: After stating ONE fact from the document, immediately close the sentence with the citation and a period. Then start a NEW sentence for the next fact. Prefer short sentences.
- **Coordinates**: The `bbox` uses **normalized coordinates** `[x1, y1, x2, y2]` in the range `[0, 1]`.
- **Uncertainty Fallback**: If you cannot precisely localize the region, provide a conservative bounding box that safely contains the relevant content. NEVER omit a citation due to bbox uncertainty.

## If Unanswerable (CRITICAL)
If the question cannot be answered from the document, briefly explain why, still citing any relevant regions you checked, such as the table of contents. Then, you MUST output exactly:
`<answer> Unanswerable </answer>`
Do NOT output variations like `<answer> I cannot answer </answer>`.

## Good & Bad Examples

BAD EXAMPLE 1 (Natural language citation - FORBIDDEN):
The revenue grew 13% in 2023 based on Table 1 on page 5.

BAD EXAMPLE 2 (Missing the second closing bracket `]` - FORBIDDEN):
Revenue was $5.2B in 2023 [page=5, doc_page="3", bbox=[0.412, 0.215, 0.687, 0.248].

BAD EXAMPLE 3 (Multiple facts, citation in the middle - FORBIDDEN):
Revenue was $5.2B in 2023 [page=5, doc_page="3", bbox=[0.41, 0.21, 0.68, 0.24]] and $4.6B in 2022.

GOOD EXAMPLE:
Let me analyze the financial data. The document reports a total revenue of $5.2 billion for fiscal year 2023 [page=5, doc_page="3", bbox=[0.412, 0.215, 0.687, 0.248]]. The total revenue for fiscal year 2022 was $4.6 billion [page=5, doc_page="3", bbox=[0.412, 0.252, 0.687, 0.285]]. The difference is $0.6B, which corresponds to approximately a 13% increase. This growth is confirmed in the management commentary [page=4, doc_page="none", bbox=[0.085, 0.612, 0.915, 0.648]].
<answer> The company's total revenue in 2023 was $5.2 billion, up approximately 13% from $4.6 billion in 2022. </answer>

## Output Protocol
1. Output your detailed reasoning process step by step.
2. Pre-output Self-Check (Perform silently):
   - Did I use the formal `[page=...]` format instead of saying "on page X"?
   - Does EVERY bbox array end with `]]`?
   - Does EVERY fact-bearing sentence have exactly ONE citation at the end?
   - Is my final answer explicitly wrapped in `<answer>` tags?
3. Output your final concise answer wrapped STRICTLY as:
`<answer> your final answer </answer>`
\end{lstlisting}
\end{tcolorbox}

\subsection{Evaluation Metric Definitions}
\label{appendix:metric_definitions}

This section provides the formal definitions of the metrics summarized in Section~\ref{subsec:evaluation_protocol}. Throughout, superscript $^{*}$ denotes gold annotations and $\hat{\mathcal{P}}_q=\mathcal{P}_q\cap\mathcal{P}_q^{*}$ denotes correctly retrieved pages for question $q$.

\paragraph{Region Grounding.}
For each page $p\in\hat{\mathcal{P}}_q$, predicted bounding boxes and gold evidence regions are rendered on the page image, and the multimodal judge labels each gold region as \texttt{covered}, \texttt{imprecise}, or \texttt{not\_covered}. Let $c_{q,p}$, $m_{q,p}$, and $|G_{q,p}|$ denote the number of covered, imprecise, and total gold regions for question $q$ on page $p$, and let $n_{q,p}^{\mathrm{pred}}$ be the number of predicted boxes. Writing $h = c$ for strict and $h = c + m$ for lenient, recall and precision are

\begin{equation}
\mathrm{R}
=
\frac{\sum_{q}\sum_{p} h_{q,p}}
     {\sum_{q}\sum_{p} |G_{q,p}|},
\qquad
\mathrm{P}
=
\frac{\sum_{q}\sum_{p} \min(h_{q,p},\; n_{q,p}^{\mathrm{pred}})}
     {\sum_{q}\sum_{p} n_{q,p}^{\mathrm{pred}}},
\end{equation}

where all inner sums range over $p\in\hat{\mathcal{P}}_q$. Precision is capped by $n_{q,p}^{\mathrm{pred}}$ because the judge labels gold regions rather than individual predicted boxes. F1 is computed in the standard way.

\paragraph{Fact Extraction.}
The judge compares each extracted fact $f\in\mathcal{F}_q$ against the gold evidence on $\hat{\mathcal{P}}_q$ and labels it as \texttt{consistent} or \texttt{not\_consistent}; the latter covers both hallucinated facts and cases where the model fails to mention relevant factual statements. We report the micro-averaged consistency rate:

\begin{equation}
\mathrm{Consistency}
=
\frac{
\sum_{q}
\bigl|\{f\in\mathcal{F}_q \mid \ell(f)=\texttt{consistent}\}\bigr|
}{
\sum_{q}|\mathcal{F}_q|
}.
\end{equation}

\subsection{Judge Prompt}
\label{appendix:bbox_v4_prompt}

\paragraph{Region Grounding Judge Prompt}
The bbox grounding judge consumes a single page image with all GT (green) and predicted (red) bounding boxes overlaid, plus the page-level list of GT boxes to label. Each batched call labels every GT box on that page in one JSON array, sharing the image cost across GT boxes. The exact template (with placeholders rendered at call time) is reproduced verbatim below.

\begin{tcolorbox}[breakable,enhanced jigsaw,colback=green!5!white,colframe=green!75!black,title=Region Grounding Judge Prompt]
\textbf{USER PROMPT}
\lstset{basicstyle=\ttfamily\scriptsize}
\begin{lstlisting}[breaklines=true]
You are evaluating whether the model-predicted evidence boxes successfully *recall* each
ground-truth (GT) evidence region on a single document page. You will judge multiple GT
boxes at once for the same page.

The page image contains:
- GREEN boxes: ground-truth evidence regions, each labelled GOLD[i] (1-based).
- RED boxes:   model-predicted evidence regions.

# Special Structural Rules (apply first, before general criteria)

  Rule 1 - Table cell/row/column -> whole table Pred:
      If GT marks a single row/column/cell and Pred covers the whole table -> covered.
  Rule 2 - Bullet/numbered list, one item GT -> whole list Pred:
      If GT marks a single bullet item and Pred covers the entire list -> imprecise.
  Rule 3 - Pred smaller than GT but mostly overlapping:
      If Pred covers the majority of the GT region's content -> covered.
  Rule 4 - Chart/figure/image as a whole:
      If GT marks a chart/figure as a whole and Pred clearly targets it -> covered
      (no need to be tight or complete).

# Input
- Question: {question}
- Gold answer (reference): {gold_answer}
- Page: {gt_page}
- Total GT boxes on this page: {gold_total_on_page}
- GT boxes to judge (one per row): {gold_facts_block}
- Predicted boxes (normalized, shared): count={n_pred}, boxes={pred_bboxes_norm}

Each row is `id=g<i> :: index_on_page=<i>, element_type=<...>, gold_bbox_px=<...>`.
Echo the `id` verbatim in your output.

# Core Criterion
This is a recall-oriented judgment. Credit is given if the GT evidence content is
effectively included in the union of all red predicted boxes, regardless of how
large/loose/imprecise the prediction is. Always evaluate the union of all red boxes;
no single red box needs to cover the GT alone. Judge every GT box independently.

# Labels
- "covered":     GT evidence is fully or effectively recalled by the red union, even if
                 the red prediction is much larger than the GT, includes unrelated content,
                 has minor clipping artifacts, or matches Rule 1 / Rule 4.
- "imprecise":   the red union meaningfully overlaps the GT but fails to recall a
                 substantial portion (Rule 2 case, partial table/chart/paragraph coverage).
- "not_covered": red union does not meaningfully overlap the GT (disjoint, boundary-only
                 contact, or refers to a clearly different region).

# Decision Procedure
1. Check if any Special Structural Rule applies; if so, apply it directly.
2. Otherwise: locate GOLD[i], form the red union, then:
     - GT content effectively recalled?            -> covered
     - Meaningful overlap but substantial part missed? -> imprecise
     - Otherwise                                    -> not_covered
Tie-breaking: prefer "covered" when >=90% of GT content / all semantically important
content is inside the red union.

# Output Format
Return exactly one JSON object on one line:
{"items":[{"id":"<gt_id>","label":"covered"|"imprecise"|"not_covered",
           "reason":"<one concise sentence>"}, ...]}
List must contain exactly one entry per input id, in order, no duplicates.
\end{lstlisting}
\end{tcolorbox}

\paragraph{Fact Extraction Judge Prompt} The fact extraction judge is a text-only model (Qwen3.6-Plus) that takes the question, the participant's free-form trajectory, and all GT facts anchored to the same page, and labels every GT fact independently in a single batched call. Admissibility is gated on structured page citations in the model output, and each fact is judged on its own evidence unit so that batch siblings cannot influence one another. The exact template (with placeholders rendered at call time) is reproduced verbatim below.

\begin{tcolorbox}[breakable,enhanced jigsaw,colback=green!5!white,colframe=green!75!black,title=Fact Extraction Judge Prompt]
\textbf{USER PROMPT}
\lstset{basicstyle=\ttfamily\scriptsize}
\begin{lstlisting}[breaklines=true]
You are judging whether a participant model's response supports each ground-truth (GT) fact for a single page. You will be given **multiple GT facts at once** (all from the same page) and must label **every one** of them independently.
# Input
- **Question:** {question}
- **GT source page (shared by all GT facts below):** {gt_page}
- **GT facts to judge (one per row, each with a stable `id`):**
{gold_facts_block}
- **Sibling GT facts (context only, do NOT judge):** {sibling_facts}
- **Participant model output:** {model_raw}
Each row in `GT facts to judge` looks like `id=<fact_id> :: <fact text>`. The `id` is a short tag such as `f1`, `f2`, etc. - use it verbatim as the key in your output.
---
# Core Rule: Only structured page citations count
For **every** GT fact, you may use **only** the evidence unit immediately before a structured citation of this exact form:
```
[page=<number>, doc_page="<string>", bbox=[...]]
```
A citation is **admissible only when its `page` field equals {gt_page}**.
The `doc_page` field is context only and must **never** be used to decide admissibility.
**These do NOT count as admissible citations:**
- "page 5", "p. 5", "(page 5)", "[page 5]", "on page 5"
- Any citation where `page` != {gt_page}, even if `doc_page` matches
- Any citation with a missing `page` field
**Final answers are always ignored**, regardless of position.
---
# Decision Process (apply independently to EACH GT fact)
## Step 1 - Find admissible citations
Scan for all structured citations beginning with `[page={gt_page},`.
If none exist ->label every GT fact as `"not_consistent"`.
## Step 2 - Extract the immediately preceding evidence unit
For each admissible citation, use **only** the single unit immediately before it:
- one sentence, OR
- one bullet/list item, OR
- one table row, OR
- one compact key-value line
If the citation appears inside a sentence, use only the part of that sentence before the citation.
**Consecutive citation sequences:** When multiple citations appear back-to-back with no intervening text, the evidence unit immediately before the **first** citation in that sequence is treated as shared evidence for **all** citations in the sequence.
Do **not** use any other text.
## Step 3 - Judge GT fact support (independently per fact)
For each GT fact `id`, ask: does any extracted evidence unit clearly support **that specific** GT fact?
To label `"consistent"`, the evidence must include:
- the **same object/target** described by that GT fact, AND
- the **same value, count, relation, or claim** in that GT fact.
Exact wording is not required. Clear paraphrase is allowed.
The same evidence unit may support multiple GT facts simultaneously - judge each fact on its own merits, never trade one fact's verdict against another's.
---
# Label Definitions (per GT fact)
## `"consistent"`
At least one admissible evidence unit clearly supports this specific GT fact. Equivalent descriptions and clear paraphrases are accepted.
## `"not_consistent"`
Use this for every other case for this specific GT fact, including:
- No admissible citation exists
- Relevant text is not immediately before an admissible citation
- Evidence mentions the right object but wrong value, or right value but wrong object
- Evidence is vague, incomplete, or off-scope
- Evidence directly conflicts with this GT fact
---
# Additional Rules (apply per-fact)
**R1. Qualifiers are required - with exceptions.**
If the GT fact includes a qualifier (year, unit, category, role, status, etc.), the admissible evidence must support that qualifier.
- **Exception - section numbers:** If the GT fact references a section by number (e.g., "section 4.2"), the evidence need only match the section name or content; the section number does not need to be explicitly stated.
- **Exception - page references in evidence text:** Page numbers mentioned within evidence text (e.g., "on page 8") do not affect admissibility and should not be used to disqualify otherwise supporting evidence.
**R2. Enumeration and counting - inclusion rule.**
- If the GT fact states that a specific item X exists (typically as part of a counting question), and the admissible evidence lists multiple items **including X**, this counts as `"consistent"` for that GT fact.
- If the GT fact states there is **exactly 1** item of type X, and the admissible evidence includes X among multiple items of that type, this still counts as `"consistent"` (the GT claim about X's existence is supported).
- If the GT fact states a **minimum count N** (e.g., "N tables with property X"), and the total items of type X found across all admissible evidence units is **less than N**, label `"not_consistent"`.
**R3. Ordered lists - position inference.**
If the GT fact specifies that an item is at a particular ordinal position (e.g., "the fourth product type"), the position may be inferred from the order in which items appear in an admissible list. Explicit ordinal labeling (e.g., "fourth") is not required.
**R4. Long lists and broad tables.**
A match inside a long list or table is admissible only if the specific item immediately before the citation directly supports the GT fact. Incidental matches do not count.
**R5. Secondary errors.**
If admissible evidence supports the GT fact but contains an unrelated secondary error ->still `"consistent"`. If the evidence gives a conflicting value for the same object and same relation ->`"not_consistent"`.
**R6. Prefer not_consistent when unclear.**
Use `"consistent"` only when support is clear and specific. Ambiguous or incomplete evidence ->`"not_consistent"`.
**R7. Multiple citations.**
If multiple admissible citations exist, evaluate each evidence unit separately. Label `"consistent"` for a given GT fact if at least one qualifies for *that* fact.
**R8. Independence across GT facts (BATCH-SPECIFIC).**
Judge every GT fact on its own. Never:
- promote one fact to `"consistent"` because a related fact in the batch is `"consistent"`,
- demote one fact because the model is generally unreliable on the page,
- collapse all facts to a single label.
You must emit **one label per `id`** in the input.
---
# Examples
**Example 1 - One evidence unit, multiple admissible facts**
GT page: 5
Input:
```
id=f1 :: The table has 3 columns.
id=f2 :: The table is on page 5.
```
> "The table has 3 columns. `[page=5, doc_page="3", bbox=[...]]`"
 `f1: consistent` (object + value match), `f2: not_consistent` (the page-5 location is the citation itself, not a claim in the evidence sentence; the evidence sentence does not state where the table is).
**Example 2 - Mixed admissible / inadmissible**
GT page: 5
Input:
```
id=f1 :: Ward A has an overall score of 65%.
id=f2 :: Ward A has a water taste score of 44%.
```
> "Ward A has water taste of 44% `[page=6, doc_page="5", bbox=[...]]`. Ward A has an overall score of 65% `[page=5, doc_page="3", bbox=[...]]`."
 `f1: consistent` (admissible evidence supports it), `f2: not_consistent` (the water-taste sentence's citation has `page=6`, not 5, and `doc_page="5"` does not count).
**Example 3 - Shared evidence in consecutive citations**
GT page: 5
Input:
```
id=f1 :: Table 16 shows a mode value of 6.0.
id=f2 :: Table 12 shows a mode value of 6.0.
```
> "The tables with mode 6.0 are Table 12, Table 16, Table 18. `[page=5, doc_page="4", bbox=[...]]`. `[page=7, doc_page="6", bbox=[...]]`."
 `f1: consistent`, `f2: consistent` (shared evidence sentence supports both).
**Example 4 - Some facts get no evidence at all**
GT page: 36
Input:
```
id=f1 :: The fourth product type listed is "Loans secured by guarantees".
id=f2 :: The seventh product type listed is "Bonds".
```
> "Those product types are mortgage loans, collateral loans, unsecured loans, loans secured by guarantees, credit facilities, and financial guarantees `[page=36, doc_page="36", bbox=[...]]`."
 `f1: consistent` (fourth item matches), `f2: not_consistent` (the list has only six items; "Bonds" is not present and there is no seventh).
**Example 5  No admissible citation anywhere**
GT page: 14
Input:
```
id=f1 :: The title of section 4.2 is "Infrastructure EMEA".
id=f2 :: Section 4.2 contains one table for EBITDA.
```
> "The 'Infrastructure EMEA' section contains one table for EBITDA on page 14."
 `f1: not_consistent`, `f2: not_consistent` (the only page reference is a free-text "on page 14", which is NOT a structured citation; nothing is admissible).
---
# Output Format
Return exactly one JSON object on one line. No markdown, no commentary, no extra text.
The object must have a single key `"items"` whose value is a list of per-fact judgments. Each judgment is an object with exactly three fields: `id`, `label`, `reason`.
- `id` MUST equal one of the input `id`s, verbatim.
- `label` MUST be exactly `"consistent"` or `"not_consistent"`.
- `reason` MUST be one concise sentence specific to that GT fact.
- The list MUST contain exactly one entry per input `id`, in the same order, with no duplicates and no extras.
Schema:
{{"items": [{{"id": "<fact_id>", "label": "consistent" | "not_consistent", "reason": "<one concise sentence>"}}, ...]}}
\end{lstlisting}
\end{tcolorbox}

\paragraph{Answer Verification Judge Prompt}
The answer verification judge is a text-only model that takes the question, the gold answer, and the model answer, then determines whether the model answer is factually consistent with the gold answer. It compares semantic correctness rather than surface wording, ignores minor formatting differences, and strictly checks numbers, entities, dates, labels, calculations, and required list completeness. The judge returns a structured JSON object with a boolean consistency label and a brief one-sentence rationale.

\begin{tcolorbox}[breakable,enhanced jigsaw,colback=green!5!white,colframe=green!75!black,title=Answer Verification Judge Prompt]
\textbf{USER PROMPT}
\lstset{basicstyle=\ttfamily\scriptsize}
\begin{lstlisting}[breaklines=true]
You are an answer consistency judge. Determine whether model_answer is factually consistent with gold_answer, using the question as context.
Judging rules:
1. Compare factual meaning, not exact wording.
2. Ignore minor formatting differences such as commas, currency symbols, capitalization, punctuation, spacing, or unit spacing. For example, "16Mbytes" and "16 Mbytes" are consistent.
3. model_answer may include extra explanation, units, or supporting values if they do not contradict gold_answer.
4. If the core number, entity, category, date, percentage, ratio, or calculation result differs from gold_answer, mark it inconsistent.
5. If the question or gold_answer requires multiple components, items, or a complete list, model_answer must include all required content. Missing required items or adding incorrect extra items is inconsistent.
6. If model_answer contains the correct answer but also adds a false or contradictory statement, mark it inconsistent.
7. Be strict with proper names, organization names, product names, and labels. If the name clearly differs, mark it inconsistent.
8. Do not use external knowledge. Judge only from question, gold_answer, and model_answer.
9. If gold_answer gives separate component values but the question asks for a combined total, a correct combined total in model_answer is consistent, as long as it directly answers the question and does not contradict gold_answer.
Return only valid JSON:
{{
  "consistent": true or false,
  "reason": "Briefly explain the judgment in one sentence."
}}
question: {question}
gold_answer: {gold_answer}
model_answer: {model_answer}
\end{lstlisting}
\end{tcolorbox}

\section{Judge Validation and Scoring Robustness}
\label{appendix:judge_validation}

\subsection{Judge--Human Alignment on Grounding Consistency}
\label{sec:analysis:bbox_alignment}

\begin{figure}[H]
    \centering
    \includegraphics[width=0.9\linewidth]{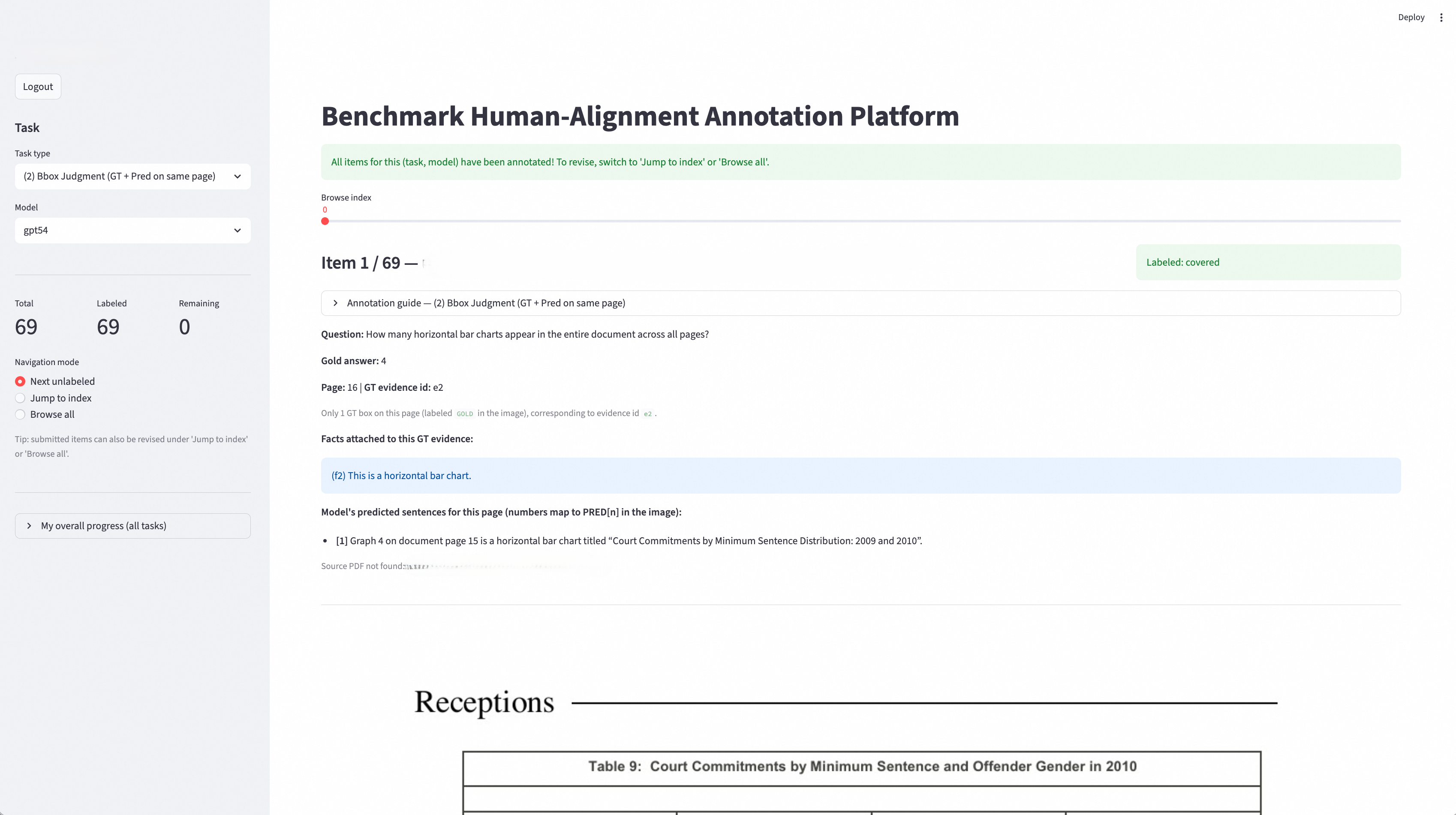}
    \caption{Annotation platform for judge-human alignment on grounding consistency.}
    \label{fig:human_platform_2}
\end{figure}

We validate the LLM-as-a-judge protocol for bounding-box grounding by measuring agreement between judge predictions and a human gold standard. Each prediction is assigned one of three ordinal labels: \textit{covered}, \textit{imprecise}, or \textit{not\_covered}. Because the judge sees the full set of GT and predicted boxes for a page in a single image, we use a page-batched protocol (one vision call per page-question pair) to amortise the image cost; the prompt is in Appendix~\ref{appendix:bbox_v4_prompt}. We additionally apply a validity gate that labels a page \textit{not\_covered} without an LLM call when no predicted bbox lies in the normalized $[0,1]^2$ range. We compare four judge models on the same $n=190$ scored items. 
We also verify the reliability of the human gold standard. For grounding-consistency annotations, annotators achieve a Krippendorff's $\alpha$ of 0.5297 and a majority agreement rate of 0.8930 over the same 190 scored items, indicating acceptable inter-annotator consistency for region grounding.
Fig.~\ref{fig:human_platform_2} shows the annotation platform used for judge-human alignment on grounding consistency.

\begin{table}[H]
\centering
\caption{Bbox judge--human alignment. We report accuracy, Cohen's $\kappa$, macro-F1, and MCC between the judge label and the human-majority label. Cost is the per-question vision-judging cost in USD, estimated from OpenRouter prices.}
\label{tab:bbox_judge_alignment_v4}
\small
\setlength{\tabcolsep}{6pt}
\begin{tabular}{lccccc}
\toprule
Judge model & Accuracy & $\kappa$ & Macro-F1 & MCC & Cost / question (USD) \\
\midrule
GPT-5.4              & 0.774 & 0.631 & 0.697 & 0.670 & $6.65\times10^{-3}$ \\
Claude Opus 4.7      & 0.842 & 0.731 & 0.759 & 0.745 & $1.32\times10^{-2}$ \\
\textbf{Qwen3.6-Plus} & 0.879 & 0.791 & 0.795 & 0.807 & $\mathbf{8.9\times10^{-4}}$ \\
GPT-5.5              & \textbf{0.895} & \textbf{0.821} & \textbf{0.839} & \textbf{0.829} & $1.32\times10^{-2}$ \\
\bottomrule
\end{tabular}
\end{table}

GPT-5.5 leads on $\kappa$, macro-F1 and MCC, while Qwen3.6-Plus is the second-best judge at \textbf{15$\times$ lower cost} ($\kappa$ gap of only $0.030$). GPT-5.4 trails the other three judges by a wide margin ($\kappa$ gap of $0.16$--$0.19$), indicating that the bbox grounding task is not yet saturated for mid-tier judges. Considering alignment quality on this spatially demanding task, we adopt \textbf{GPT-5.5} as the default judge for region grounding in our main results. To rule out the possibility that purely geometric reference metrics (IoU, IoM, GT-recall) could replace the LLM judge, we further run a paired-bootstrap comparison under Kendall $\tau_b$; the analysis and significance tests are deferred to Appendix~\ref{appendix:bbox_judge_vs_geom}.

\subsection{Bbox LLM Judge vs.\ Rule-Based Geometric Baselines}
\label{appendix:bbox_judge_vs_geom}

A natural concern about the bbox alignment results in \S\ref{sec:analysis:bbox_alignment} is whether expensive vision-LLM judging is needed at all, given that bbox tasks have well-defined geometric reference metrics: pairwise IoU, IoM (intersection-over-minimum, $|G \cap U| / \min(|G|, |U|)$), and GT-recall ($|G \cap U| / |G|$), where $U$ is the union of valid normalized predicted boxes on the page. We compute all three for every scored item and use them as continuous-valued predictors of the human ordinal label, then compare them against the discrete LLM judge under the same Kendall $\tau_b$ via paired bootstrap (2000 resamples). Results are in Tab.~\ref{tab:bbox_judge_vs_geom_v4}.

\begin{table}[H]
\centering
\caption{Per-judge paired-bootstrap difference $\tau_b(\text{judge}) - \tau_b(\text{geometric})$. Positive values indicate the LLM judge aligns with humans better than the geometric baseline; one-sided $p$ tests $H_0: \tau_b(\text{judge}) \le \tau_b(\text{geom})$. Bold marks $p < 0.05$.}
\label{tab:bbox_judge_vs_geom_v4}
\small
\setlength{\tabcolsep}{4.5pt}
\begin{tabular}{lcccccc}
\toprule
Judge & $\Delta$ vs.\ IoU & $p$ & $\Delta$ vs.\ IoM & $p$ & $\Delta$ vs.\ GT-recall & $p$ \\
\midrule
GPT-5.4              & $-0.105$ & $0.99$  & $-0.086$ & $0.98$  & $-0.114$ & $1.00$  \\
Claude Opus 4.7      & $-0.002$ & $0.51$  & $+0.016$ & $0.31$  & $-0.011$ & $0.63$  \\
GPT-5.5              & $\mathbf{+0.057}$ & $\mathbf{0.013}$ & $\mathbf{+0.075}$ & $\mathbf{0.004}$ & $\mathbf{+0.047}$ & $\mathbf{0.020}$ \\
\textbf{Qwen3.6-Plus} & $\mathbf{+0.064}$ & $\mathbf{0.003}$ & $\mathbf{+0.083}$ & $\mathbf{<0.001}$ & $\mathbf{+0.055}$ & $\mathbf{0.003}$ \\
\bottomrule
\end{tabular}
\end{table}

The geometric baselines themselves are not weak: across all four runs, $\tau_b$(IoM) and $\tau_b$(GT-recall) are around $0.83$--$0.86$, comparable to a moderately strong judge. However, only GPT-5.5 and Qwen3.6-Plus produce strictly better alignment than every geometric baseline at $p < 0.05$. GPT-5.4 is in fact \emph{worse} than the geometric baselines under $\tau_b$, and Claude Opus 4.7 is statistically indistinguishable from them. This split mirrors the absolute $\kappa$ ranking in Tab.~\ref{tab:bbox_judge_alignment_v4} and reinforces our judge selection: \textbf{rule-based geometric metrics are a credible weak baseline but cannot replace a frontier-quality LLM judge for the bbox grounding task}, and the gap is realised only when the judge itself crosses a quality threshold.

We use Kendall $\tau_b$ because the human label is ordinal with three levels and the geometric baselines are continuous, a setting where Pearson and Spearman are biased while $\tau_b$ handles the ordinal ties correctly. For each (judge, geometric metric) pair we pair the two $\tau_b$ scores on the same $B=2000$ resampled rows and report the one-sided $p$-value for $H_0: \tau_b(\text{judge}) \le \tau_b(\text{geom})$. The geometric baselines and the LLM judge are evaluated on identical rows, so the comparison is strictly within-subject.

\subsection{Judge--Human Alignment on Factual Consistency}
\label{sec:analysis:judge_alignment}

\begin{figure}[H]
    \centering
    \includegraphics[width=0.9\linewidth]{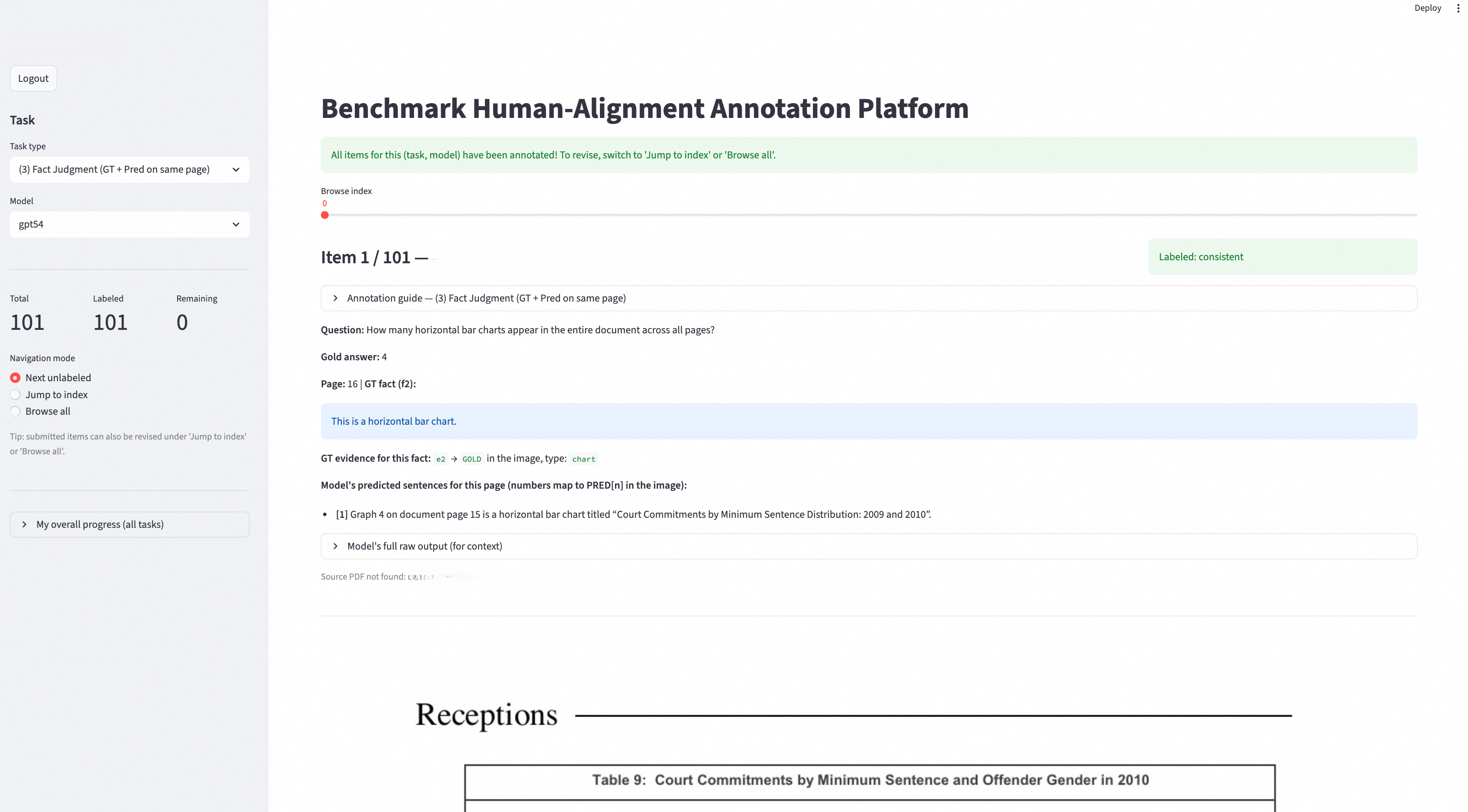}
    \caption{Annotation platform for judge-human alignment on factual consistency.}
    \label{fig:human_platform_3}
\end{figure}

We validate the LLM-as-a-judge protocol by measuring agreement between judge predictions and a human gold standard, and report accuracy, Cohen's $\kappa$, macro-F1, and MCC. For Qwen3.6-Plus we additionally include a batch-inference variant. Results are summarized in Tab.~\ref{tab:judge_alignment_v2}. We also verify the reliability of the human gold standard. For factual-consistency annotations, annotators achieve a Krippendorff's $\alpha$ of 0.6115 and a majority agreement rate of 0.9181 over the annotated items, indicating acceptable inter-annotator consistency for fact-level judgments. Fig.~\ref{fig:human_platform_3} shows the annotation platform used for judge-human alignment on factual consistency.

\begin{table}[H]
\centering
\caption{Judge--human alignment across different judge models. Cost is the per-question judging cost in USD, estimated from OpenRouter unit prices.}
\label{tab:judge_alignment_v2}
\small
\begin{tabular}{lccccc}
\toprule
Judge model & Accuracy & $\kappa$ & Macro-F1 & MCC & Cost / question (USD) \\
\midrule
GPT-5.4              & 0.908 & 0.768 & 0.884 & 0.773 & $1.61\times 10^{-2}$ \\
GPT-5.5              & 0.907 & 0.791 & 0.895 & 0.801 & $3.21\times 10^{-2}$ \\
Claude Opus 4.7      & 0.913 & 0.779 & 0.889 & 0.783 & $3.15\times 10^{-2}$ \\
DeepSeek-V4-Flash    & 0.915 & 0.775 & 0.887 & 0.780 & $8.3\times 10^{-4}$ \\
Qwen3.6-Plus (batch) & 0.931 & 0.827 & 0.913 & 0.834 & $\mathbf{1.50\times 10^{-3}}$ \\
\textbf{Qwen3.6-Plus} & \textbf{0.954} & \textbf{0.880} & \textbf{0.940} & \textbf{0.881} & $2.09\times 10^{-3}$ \\
\bottomrule
\end{tabular}
\end{table}

Qwen3.6-Plus clearly leads on every metric ($\kappa = 0.880$), exceeding the next-best judge GPT-5.5 by $0.089$ and the same-prompt Claude Opus 4.7 by $0.101$. The other four judges cluster tightly in $\kappa \in [0.768, 0.791]$, indicating that the binary factual-consistency task is largely saturated for frontier judges except that Qwen3.6-Plus opens an additional gap. Batch inference yields a $\Delta\kappa = -0.053$ drop for Qwen3.6-Plus, which we hypothesize stems from inter-sample interference within batched prompts. Considering both alignment quality and cost, we adopt \textbf{Qwen3.6-Plus}(batch) as the default fact-extraction judge in our main results.

\subsection{Judge--Human Alignment on Answer Verification}
\label{app:analysis:judge_alignment_answer}
\begin{figure}[H]
    \centering
    \includegraphics[width=0.9\linewidth]{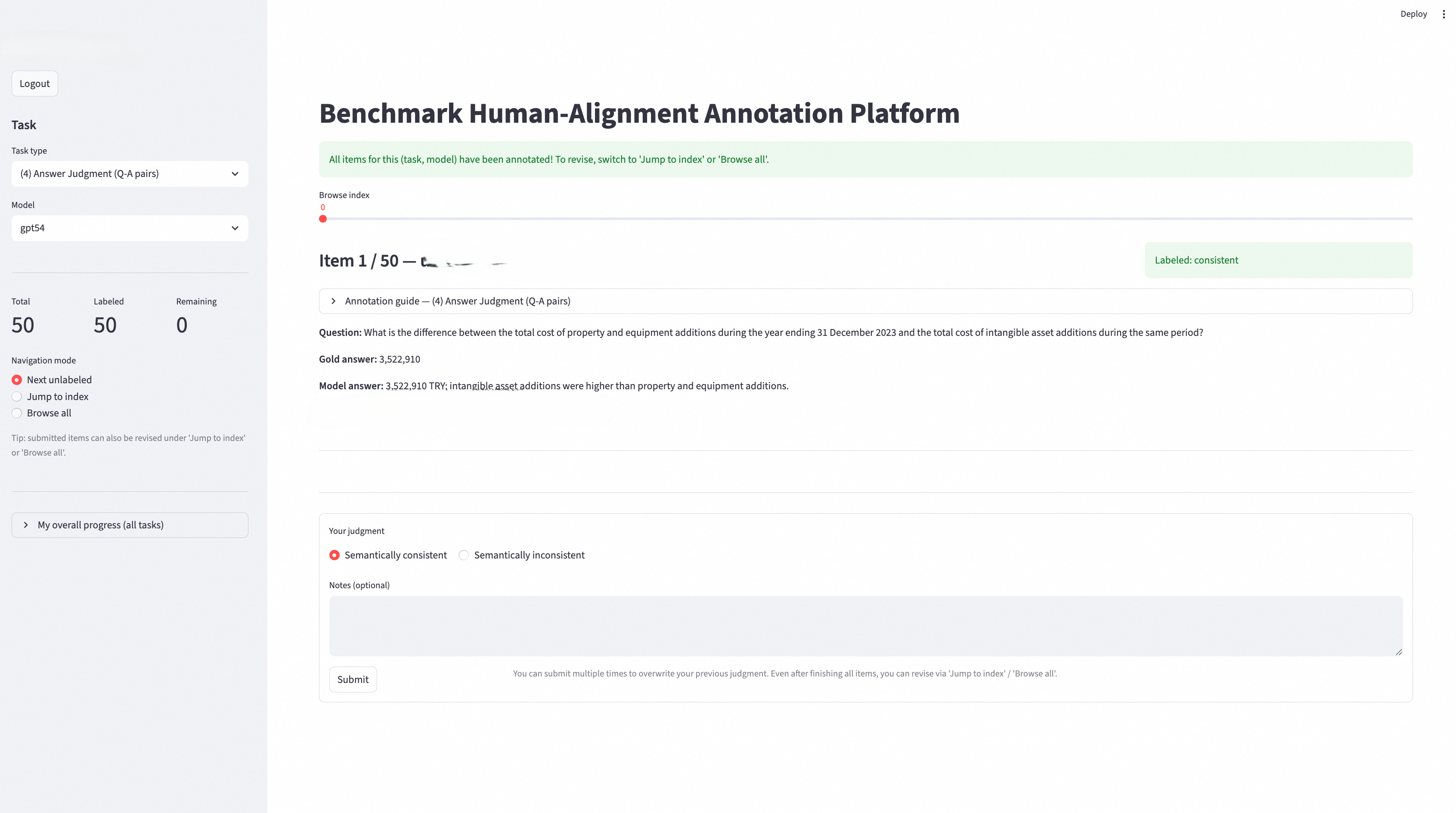}
    \caption{Annotation platform for judge--human alignment on answer verification.}
    \label{fig:human_platform_4}
\end{figure}

To validate the reliability of LLM-as-Judge for answer correctness evaluation in our benchmark, we conduct a human alignment study by sampling 150 benchmark instances and comparing the judgments from different judge models against human annotations. We also verify the reliability of the human gold standard. For answer-verification annotations, annotators achieve a Krippendorff's $\alpha$ of 0.8906 and a majority agreement rate of 0.9733 over 150 benchmark instances, indicating strong inter-annotator consistency for answer-level judgments. Fig.~\ref{fig:human_platform_4} shows the annotation platform used for judge--human alignment on answer verification. 

As shown in Tab.~\ref{tab:judge_human_alignment}, all judge models achieve strong alignment with human judgments, with accuracy above 0.96 and Cohen's Kappa above 0.91. Among them, Qwen3.6-Plus achieves the best overall performance, obtaining the highest Accuracy, Kappa, Macro-F1, and MCC scores of 0.973, 0.946, 0.973, and 0.946, respectively. Meanwhile, its evaluation cost is substantially lower than stronger proprietary models such as GPT-5.5 and Claude Opus 4.7. Considering its superior alignment, robustness, and cost-effectiveness, we adopt \textbf{Qwen3.6-Plus} as the answer-verification judge in our main results.

\begin{table}[H]
\centering
\caption{
Human alignment results of different LLM-as-Judge models for answer correctness evaluation.
}
\label{tab:judge_human_alignment}
\vspace{0.5em}
\small
\setlength{\tabcolsep}{6pt}
\renewcommand{\arraystretch}{1.12}
\begin{tabular}{lccccc}
\toprule
Judge model & Accuracy & $\kappa$ & Macro-F1 & MCC & Cost / question (USD) \\
\midrule
Claude Opus 4.7      & 0.966 & 0.932 & 0.966 & 0.932 & $4.11\times 10^{-3}$ \\
DeepSeek-V4-Flash    & 0.960 & 0.920 & 0.960 & 0.920 & $\mathbf{5.43\times 10^{-5}}$ \\
GPT-5.4              & 0.966 & 0.933 & 0.966 & 0.934 & $1.49\times 10^{-3}$ \\
GPT-5.5              & 0.967 & 0.933 & 0.967 & 0.933 & $3.80\times 10^{-3}$ \\
\textbf{Qwen3.6-Plus} & \textbf{0.973} & \textbf{0.946} & \textbf{0.973} & \textbf{0.946} & $9.06\times 10^{-4}$ \\
\bottomrule
\end{tabular}
\end{table}

\subsection{Comparison with Prior Answer Verification Method}
\label{app:compare_with_mmlongbench}
In Appendix~\ref{app:analysis:judge_alignment_answer}, we validate the reliability of our LLM-as-Judge protocol for answer correctness evaluation through a human alignment study. Moreover, we compare our protocol with the answer verification method used in MMLongBench-Doc, a prior long-context multimodal benchmark. MMLongBench-Doc~\citep{ma2024mmlongbenchdoc} adopts a two-stage verification pipeline: it first uses GPT-4o to extract the final answer from the model response, and then applies type-aware verification according to the answer format. For Integer, Float, and String answers, an LLM is prompted to judge whether the extracted prediction matches the reference answer. For List-type answers, MMLongBench-Doc uses a specialized prompt and F1-style matching to account for partially correct predictions.

Although this pipeline provides a practical way to standardize evaluation across heterogeneous answer formats, it relies on an explicit answer extraction step and type-specific matching rules. This design can be less robust for reasoning-model outputs, where responses often contain long explanations, implicit final answers, paraphrased expressions, or semantically correct answers that differ from the reference in surface form.

Following the human alignment study in Appendix~\ref{app:analysis:judge_alignment_answer}, we reproduce the MMLongBench-Doc answer verification method on the same human-annotated subset and compare it with our LLM-as-Judge protocol. Human alignment is measured using generalized accuracy and AUROC, while DeLong tests are used to assess whether the AUROC differences between our protocol and the MMLongBench-Doc method are statistically significant.

\begin{table}[H]
\centering
\caption{
Comparison between the MMLongBench-Doc answer verification method and our verification protocol.
AUROC Gain denotes the AUROC improvement over the MMLongBench-Doc method on the corresponding paired samples.
$p$-values are computed using DeLong tests.
Bold values indicate statistically significant improvements over the MMLongBench-Doc method with $p<0.05$.
}
\label{tab:mmlongbench_judge_comparison}
\small
\setlength{\tabcolsep}{5pt}
\renewcommand{\arraystretch}{1.12}
\begin{tabular}{lcccc}
\toprule
Method & Gen. Acc. & AUROC & AUROC Gain & $p$-value \\
\midrule
MMLongBench-Doc 
& 0.800 & 0.826 & -- & -- \\
Claude Opus 4.7 
& 0.966 & 0.966 & $\mathbf{+0.141}$ & $\mathbf{1.28{\times}10^{-5}}$ \\
DeepSeek-V4-Flash 
& 0.960 & 0.961 & $\mathbf{+0.135}$ & $\mathbf{2.18{\times}10^{-5}}$ \\
GPT-5.4 
& 0.966 & 0.967 & $\mathbf{+0.144}$ & $\mathbf{1.07{\times}10^{-5}}$ \\
GPT-5.5 
& 0.967 & 0.967 & $\mathbf{+0.141}$ & $\mathbf{1.08{\times}10^{-5}}$ \\
\textbf{Qwen3.6-Plus}
& \textbf{0.973} & \textbf{0.973} & $\mathbf{+0.150}$ & $\mathbf{1.79{\times}10^{-6}}$ \\
\bottomrule
\end{tabular}
\end{table}

As shown in Tab.~\ref{tab:mmlongbench_judge_comparison}, the reproduced MMLongBench-Doc method obtains 0.800 generalized accuracy and 0.826 AUROC. In contrast, all variants of our judge-based protocol achieve over 0.960 on both metrics, indicating substantially stronger alignment with human annotations. In terms of AUROC, our protocol consistently improves over the MMLongBench-Doc method by 0.135--0.150 across different judge models. These gains are statistically significant under DeLong tests, with all $p$-values below $2.2{\times}10^{-5}$. Among all judges, Qwen3.6-Plus performs best, achieving 0.973 generalized accuracy and 0.973 AUROC, with the largest AUROC gain of 0.150 and the strongest significance level of $p=1.79{\times}10^{-6}$. These results demonstrate that our judge-based protocol is more closely aligned with human annotations than the MMLongBench-Doc method, supporting its use as the default answer verification scheme in our benchmark.

\section{Experiment Environment for Evaluation}

\label{appendix:experiment_environment}

The inference experiments were primarily conducted on a server equipped with four 80GB NVIDIA L20Z GPUs. Specifically, for proprietary and open-weight models, we performed inference through their corresponding API endpoints. The maximum number of output tokens was set to 16,384 for all models.

For SimpleDoc~\citep{jain2025simpledoc}, we used ColQwen2.5~\citep{faysse2024colpali} as the embedding model and vector retriever for document pages, Qwen2.5-VL-32B~\citep{bai2025qwen25vl} as the page summarization model, Qwen3-30B-A3B~\citep{yang2025qwen3} as the reranker based on the summarized text, and Qwen2.5-VL-32B as the final QA model. For the VidoRAG framework, we used ColQwen2 as the embedding model and retriever, and Qwen2.5-VL-7B-Instruct as the reasoning model for the multi-agent system. For URaG~\citep{shi2026urag}, we adopted the default settings reported in the original paper, using Top-5 retained pages for retrieval and the hidden states from the 6th early layer as retrieval features.

For Docopilot~\citep{duan2025docopilot}, due to the context-length constraints of their backbone models, we followed a protocol similar to that used in MMLongBench-Doc~\citep{ma2024mmlongbenchdoc}: images were merged into a fixed number of five composite images, with at most three columns per group. In addition, although Intern-S1-Pro~\citep{zou2026intern} has publicly available weights, its large parameter scale required us to use the official API. Owing to the API limit on the number of input images, we adopted a similar merging strategy and combined all images into 20 composite images.

\section{Additional Analyses}
\label{appendix:additional_analysis}

\subsection{Further Analysis of Evidence Distribution Factors and Question Difficulty}
\label{appendix:evidence_distribution_difficulty}

In Section~\ref{sec:analysis:evidence_distribution}, we discuss three representative evidence-layout factors: the number of ground-truth evidence pages, the maximum adjacent gap between evidence pages, and the number of separated evidence clusters.
In this appendix, we extend the analysis to six factors by additionally considering the overall page span covered by the evidence, the mean pairwise distance among evidence pages, and the normalized mean position of evidence pages within the document.
These complementary factors provide a more detailed view of whether model performance is mainly affected by evidence quantity, long-range dispersion, structural fragmentation, or evidence location.

\begin{figure*}[t]
    \centering
    \includegraphics[width=\textwidth]{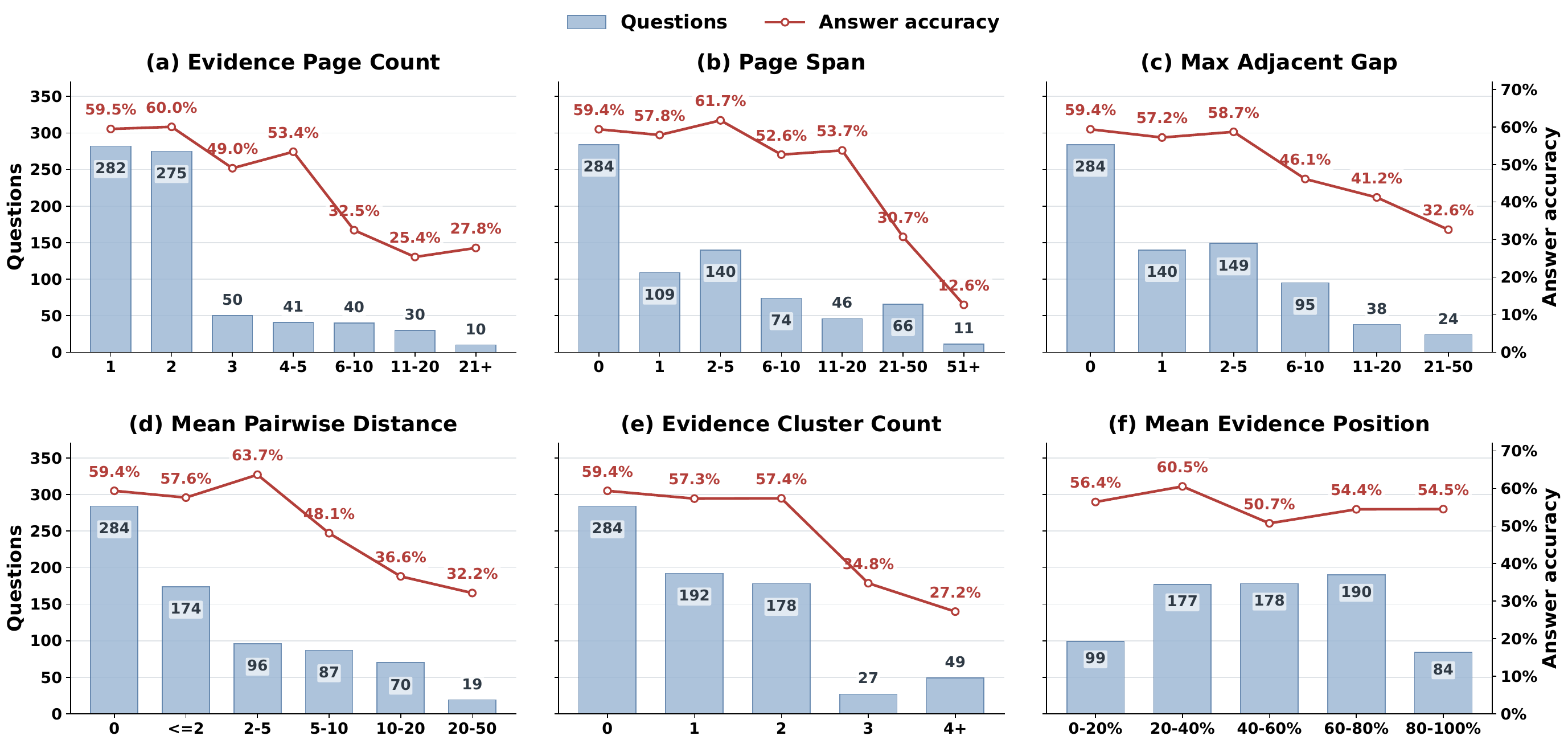}
    \caption{
    Detailed relationship between evidence distribution factors and answer accuracy.
    Bars denote the number of questions in each bin, while red lines denote answer accuracy.
    We analyze six evidence-layout factors:
    (a) the number of ground-truth evidence pages,
    (b) the page span covered by the evidence,
    (c) the maximum adjacent gap between evidence pages,
    (d) the mean pairwise distance among evidence pages,
    (e) the number of separated evidence clusters,
    and (f) the normalized mean position of evidence pages within the document.
    }
    \label{fig:evidence_factors_accuracy_six_panels}
\end{figure*}

The page span in Fig.~\ref{fig:evidence_factors_accuracy_six_panels}(b) measures the overall range between the earliest and latest evidence pages, while the mean pairwise distance in Fig.~\ref{fig:evidence_factors_accuracy_six_panels}(d) captures the global dispersion among all evidence pages.
Both factors show that answer accuracy tends to decrease when evidence is distributed over longer distances, indicating that difficulty arises not only from a single large discontinuity between adjacent evidence pages, as discussed in the main text, but also from the broader spread of evidence across the document.
In particular, even when the number of required evidence pages is limited, a large page span or a large mean pairwise distance can still make the question challenging, since the model must retrieve, retain, and integrate information from distant document regions.
By contrast, the normalized mean position of evidence pages in Fig.~\ref{fig:evidence_factors_accuracy_six_panels}(f) does not exhibit a clear monotonic relationship with answer accuracy.
This suggests that whether the evidence appears earlier or later in the document is less important than how dispersed the evidence is.
Overall, these additional factors reinforce the conclusion that the key bottleneck in multimodal long-document question answering lies in integrating evidence across long-range and disconnected contexts, rather than merely locating evidence at a particular document position.

\subsection{Detailed Results of Evidence-Chain Completeness among Correct Answers}
\label{app:evidence_chain_completeness}
Detailed results of Section~\ref{sec:evidence_chain_completeness} are shown in Tab.~\ref{tab:evidence_chain_completeness}.

\begin{table*}[t!]
\centering
\fontsize{4.9pt}{5.6pt}\selectfont
\setlength{\tabcolsep}{1.35pt}
\renewcommand{\arraystretch}{0.90}
\caption{
Detailed breakdown of evidence-chain completeness among answer-correct samples.
P, B, and F denote page recall, bounding-box localization, and fact support, respectively.
\cmark{} indicates completeness and \xmark{} indicates incompleteness.
The largest mutually exclusive condition in each row is highlighted in \textbf{bold}.
}
\label{tab:evidence_chain_completeness}
\resizebox{\linewidth}{!}{%
\begin{tabular}{lccccccr}
\toprule
\textbf{Model} &
\makecell{\textbf{P\cmark}\\\textbf{B\cmark F\cmark}} &
\makecell{\textbf{P\xmark}\\\textbf{B\xmark F\xmark}} &
\makecell{\textbf{P\xmark}\\\textbf{B/F\cmark}} &
\makecell{\textbf{P\cmark}\\\textbf{B\xmark F\xmark}} &
\makecell{\textbf{P\cmark}\\\textbf{B\xmark F\cmark}} &
\makecell{\textbf{P\cmark}\\\textbf{B\cmark F\xmark}} &
\makecell{\textbf{Answer}\\\textbf{Correct}} \\
\midrule
\rowcolor{gray!10}
\multicolumn{8}{c}{\textit{Proprietary Models}} \\
Gemini 3.1 Pro &
83 (14.04\%) & 86 (14.55\%) & 1 (0.17\%) & 141 (23.86\%) & \textbf{260 (43.99\%)} & 20 (3.38\%) & 591 \\
Gemini 3.1 Flash Lite &
98 (19.22\%) & 98 (19.22\%) & 0 (0.00\%) & 90 (17.65\%) & \textbf{204 (40.00\%)} & 20 (3.92\%) & 510 \\
Claude Opus 4.7 &
99 (17.37\%) & 85 (14.91\%) & 1 (0.18\%) & 113 (19.82\%) & \textbf{246 (43.16\%)} & 26 (4.56\%) & 570 \\
Claude Sonnet 4.6 &
120 (22.64\%) & 75 (14.15\%) & 1 (0.19\%) & 76 (14.34\%) & \textbf{215 (40.57\%)} & 43 (8.11\%) & 530 \\
GPT-5.4 &
\textbf{136 (29.06\%)} & 89 (19.02\%) & 0 (0.00\%) & 65 (13.89\%) & 133 (28.42\%) & 45 (9.62\%) & 468 \\
Qwen3.6 Plus &
45 (8.98\%) & 112 (22.36\%) & 0 (0.00\%) & 95 (18.96\%) & \textbf{207 (41.32\%)} & 42 (8.38\%) & 501 \\
\midrule
\rowcolor{gray!10}
\multicolumn{8}{c}{\textit{Open-weight Models}} \\
Intern-S1-Pro &
3 (1.15\%) & \textbf{232 (88.89\%)} & 0 (0.00\%) & 9 (3.45\%) & 15 (5.75\%) & 2 (0.77\%) & 261 \\
Qwen3.5-397B-A17B &
23 (4.81\%) & 134 (28.03\%) & 0 (0.00\%) & 126 (26.36\%) & \textbf{174 (36.40\%)} & 21 (4.39\%) & 478 \\
Qwen3.5-122B-A10B &
8 (1.68\%) & \textbf{284 (59.79\%)} & 1 (0.21\%) & 70 (14.74\%) & 97 (20.42\%) & 15 (3.16\%) & 475 \\
Qwen3.5-27B &
52 (10.53\%) & 69 (13.97\%) & 1 (0.20\%) & 132 (26.72\%) & \textbf{206 (41.70\%)} & 34 (6.88\%) & 494 \\
Qwen3-VL-235B-A22B &
19 (4.92\%) & 108 (27.98\%) & 1 (0.26\%) & 86 (22.28\%) & \textbf{150 (38.86\%)} & 22 (5.70\%) & 386 \\
Qwen3-VL-32B &
13 (3.34\%) & \textbf{147 (37.79\%)} & 1 (0.26\%) & 103 (26.48\%) & 107 (27.51\%) & 18 (4.63\%) & 389 \\
Qwen3-VL-30B-A3B &
4 (1.57\%) & \textbf{122 (47.84\%)} & 0 (0.00\%) & 79 (30.98\%) & 43 (16.86\%) & 7 (2.75\%) & 255 \\
Qwen3-VL-8B &
3 (1.18\%) & \textbf{212 (83.46\%)} & 0 (0.00\%) & 10 (3.94\%) & 26 (10.24\%) & 3 (1.18\%) & 254 \\
Gemma-4-31B &
52 (11.69\%) & 68 (15.28\%) & 2 (0.45\%) & 79 (17.75\%) & \textbf{226 (50.79\%)} & 18 (4.04\%) & 445 \\
Gemma-4-26B-A4B &
3 (1.15\%) & \textbf{110 (42.15\%)} & 0 (0.00\%) & 55 (21.07\%) & 93 (35.63\%) & 0 (0.00\%) & 261 \\
Ministral3-14B &
1 (0.32\%) & \textbf{135 (43.69\%)} & 1 (0.32\%) & 76 (24.60\%) & 92 (29.77\%) & 4 (1.29\%) & 309 \\
Ministral3-8B &
2 (0.74\%) & \textbf{126 (46.67\%)} & 1 (0.37\%) & 59 (21.85\%) & 79 (29.26\%) & 3 (1.11\%) & 270 \\
\bottomrule
\end{tabular}%
}
\vspace{-2mm}
\end{table*}

\subsection{Oracle Evidence Access Study: Detailed Setup}
\label{appendix:oracle_setup}

To identify which stage of the evidence-grounding pipeline constitutes the dominant bottleneck, we progressively supply models with gold annotations, incrementally removing the demands of each trajectory stage. Four models spanning a broad capability spectrum are selected: Claude Sonnet 4.6, GPT-5.4, Qwen3-VL-235B-A22B, and Ministral3-8B. The experiment comprises three cumulative oracle settings:

\begin{enumerate}
    \item \textbf{Oracle Pages.} The input context is restricted to the gold evidence pages while retaining the standard reasoning prompt, removing the page-localization burden.
    \item \textbf{Oracle Regions.} Building on (1), textual bounding-box descriptions of key evidence regions are injected into the prompt, additionally removing the region-grounding burden.
    \item \textbf{Oracle Facts.} Building on (2), the atomic facts contained in each annotated region are additionally provided, further removing the perceptual and fact-extraction burden.
\end{enumerate}

\subsection{Oracle Evidence Access Study: Trajectory Metric Observations}
\label{appendix:oracle_observations}

Beyond the answer-accuracy trends discussed in Section~\ref{sec:analysis:oracle}, we observe two consistent patterns across trajectory metrics. First, Page Localization F1 rises sharply once oracle pages are provided and then plateaus, confirming that the oracle effectively removes the page-retrieval burden. Second, Fact Consistency increases steadily across all oracle settings for every model, mirroring the answer-accuracy finding that fact extraction is the primary bottleneck.

\subsection{Oracle Grounding Behavior Case Study}
\label{appendix:oracle_grounding_case_study}

As discussed in Section~\ref{sec:analysis:oracle}, providing oracle pages counter-intuitively \emph{decreases} Region Grounding F1 for stronger models. We present two representative cases illustrating the conservative-to-aggressive strategy shift.

\paragraph{Case 1: Claude Sonnet 4.6 on a table page.}
Under the standard setting (left), the model outputs two large bounding boxes (Pred~1, Pred~2) that broadly cover the table area, incidentally encompassing the gold evidence region (a specific table row). Under the oracle-pages setting (right), the model shifts to a single smaller box targeting a specific row---but selects the wrong row, missing the gold region entirely. The large conservative boxes in the standard setting achieved \texttt{covered} by accident; the precise but mislocalized box in the oracle setting is judged \texttt{not\_covered}.

\begin{figure}[h]
    \centering
    \begin{minipage}[t]{0.48\linewidth}
        \centering
        \includegraphics[width=\linewidth]{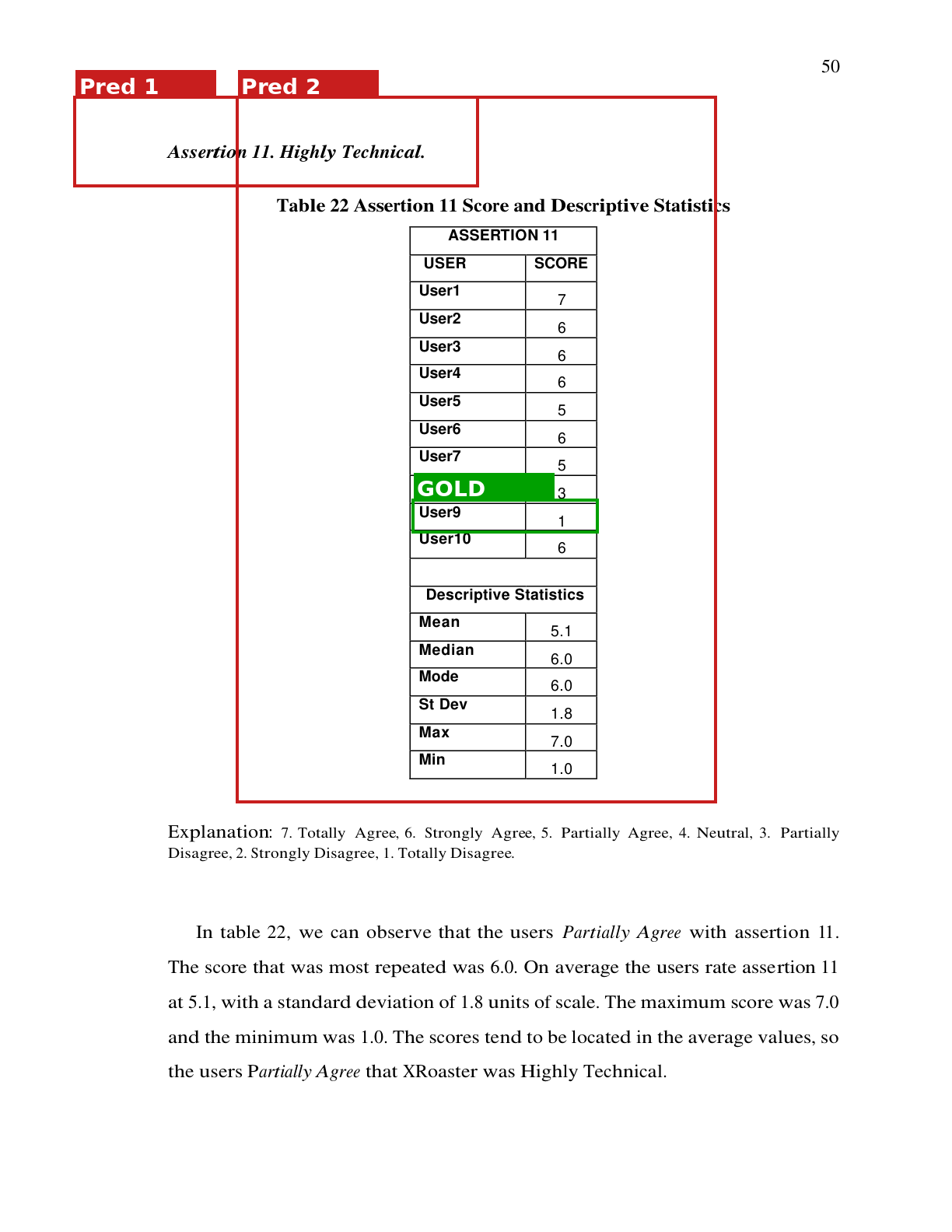}
        \small Standard setting
    \end{minipage}\hfill
    \begin{minipage}[t]{0.48\linewidth}
        \centering
        \includegraphics[width=\linewidth]{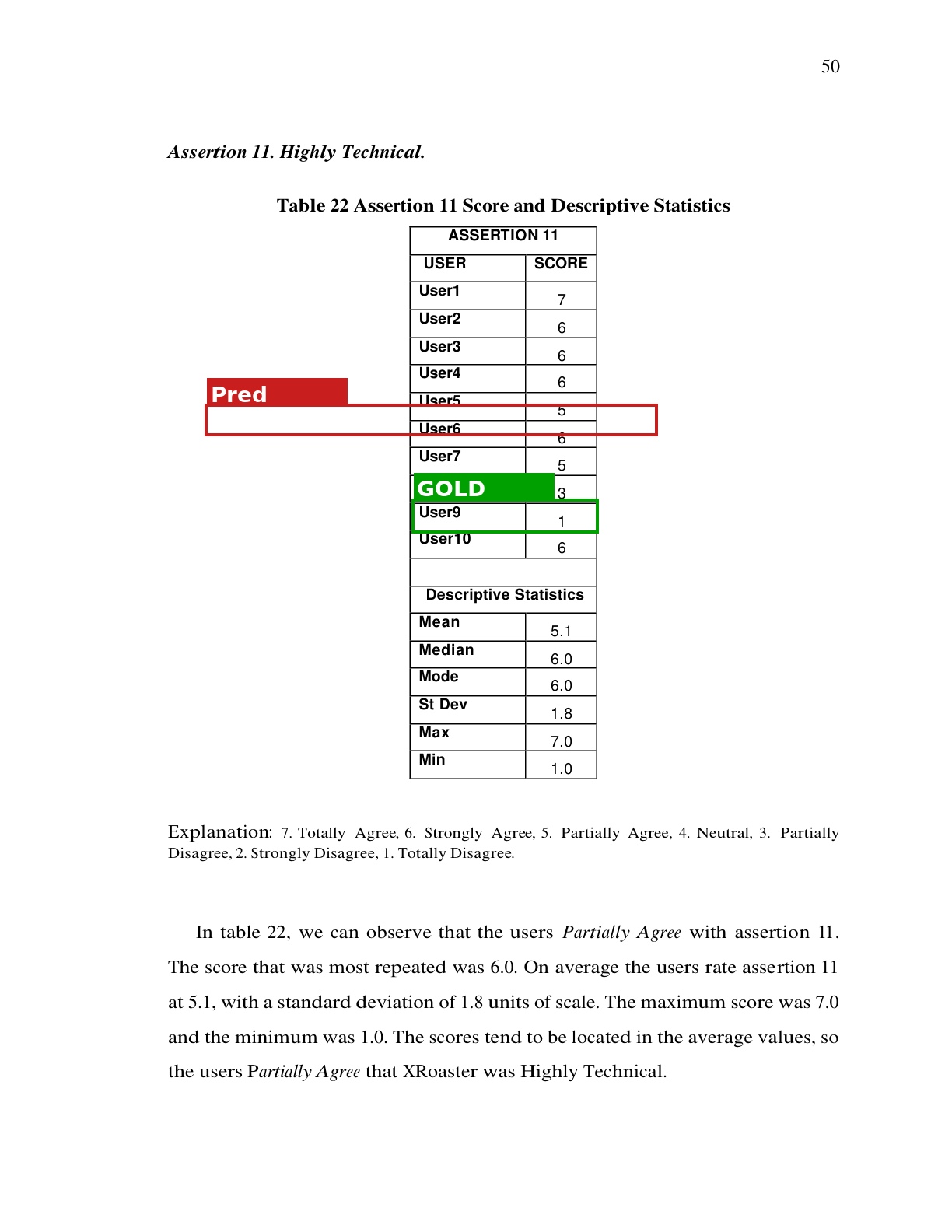}
        \small Oracle pages setting
    \end{minipage}
    \caption{Case 1: Claude Sonnet 4.6 grounding behavior on a table page. Green boxes denote gold evidence regions; red boxes denote model predictions. Under the standard setting, large conservative boxes cover the gold region incidentally. Under oracle pages, the model produces a smaller, more targeted box that misses the correct row.}
    \label{fig:case_oracle_claude}
\end{figure}

\paragraph{Case 2: GPT-5.4 on a financial table.}
A similar pattern is observed with GPT-5.4. In the standard setting (left), two large predicted boxes span broad table sections, covering the gold evidence row within their extent. In the oracle-pages setting (right), the model produces a single compact box aimed at a specific row of the financial table---but targets the wrong row, landing several entries away from the gold region. This confirms that the strategy shift is not model-specific but reflects a general tendency: when the search space is narrowed by oracle pages, models attempt finer-grained localization that exceeds their actual spatial precision.

\begin{figure}[h]
    \centering
    \begin{minipage}[t]{0.48\linewidth}
        \centering
        \includegraphics[width=\linewidth]{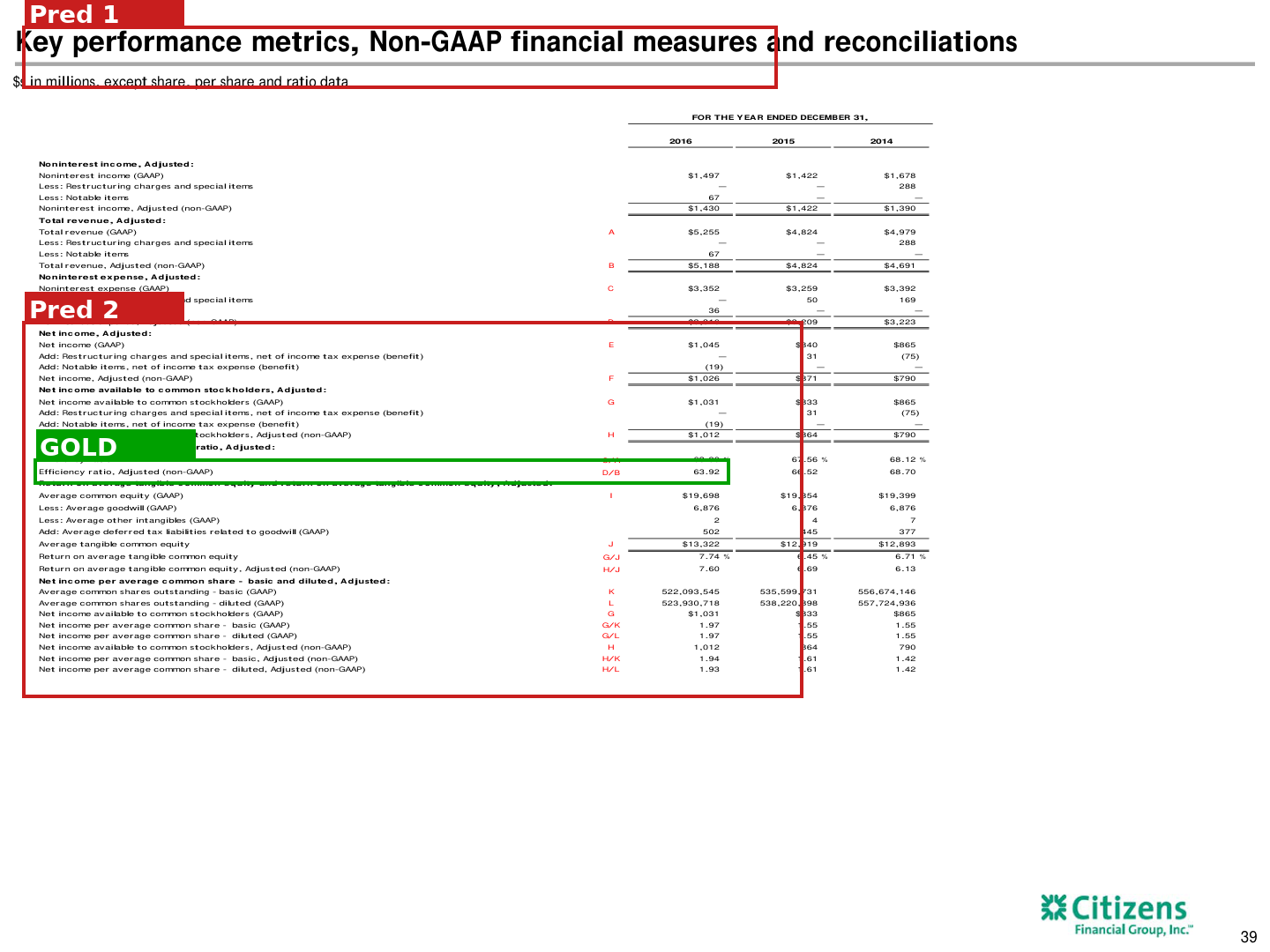}
        \small Standard setting
    \end{minipage}\hfill
    \begin{minipage}[t]{0.48\linewidth}
        \centering
        \includegraphics[width=\linewidth]{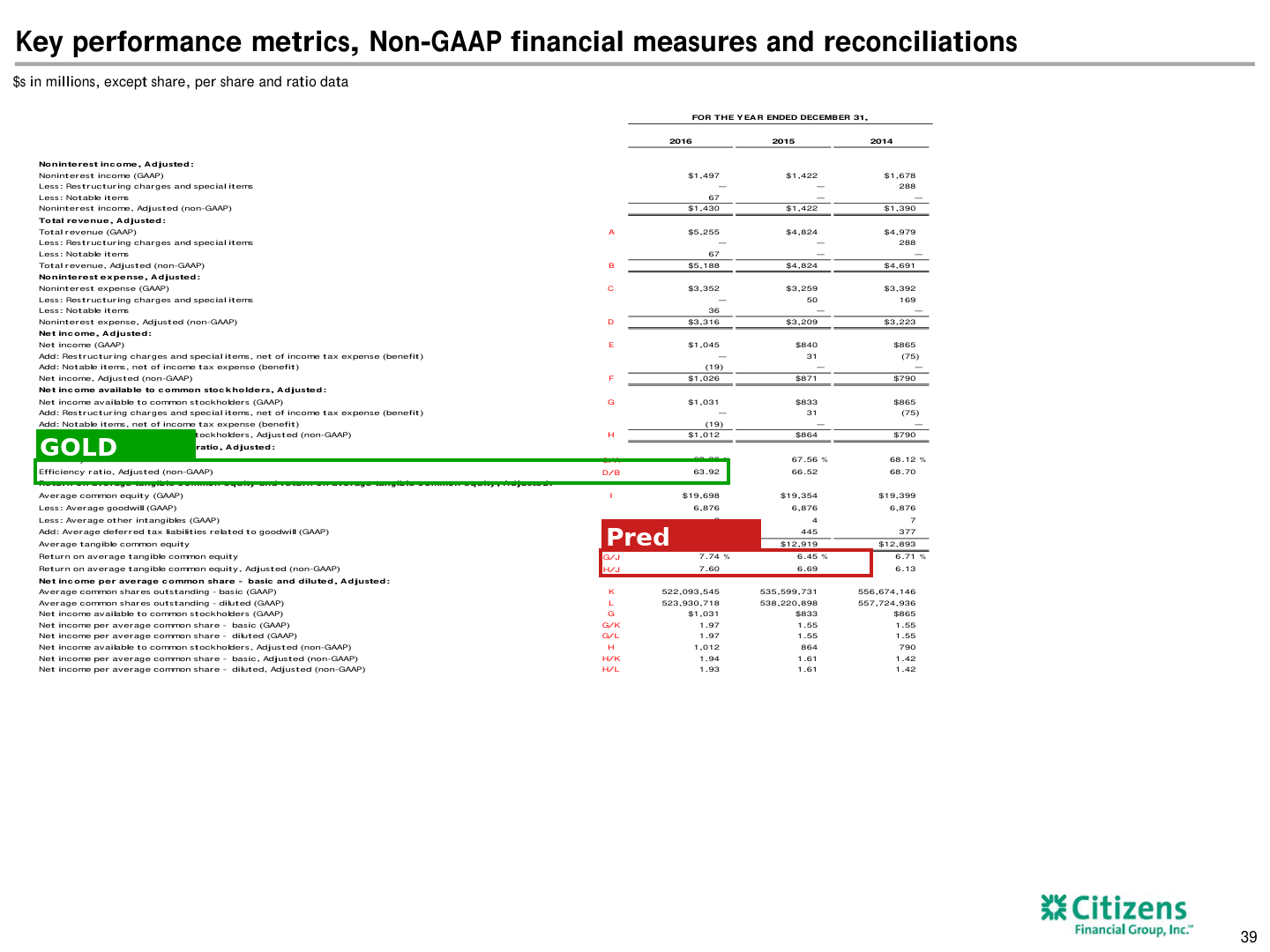}
        \small Oracle pages setting
    \end{minipage}
    \caption{Case 2: GPT-5.4 grounding behavior on a financial table page. The same conservative-to-aggressive shift is observed: large boxes in the standard setting cover the gold region, while a smaller targeted box in the oracle setting misses it.}
    \label{fig:case_oracle_gpt}
\end{figure}

\subsection{Error Analysis}
\label{app:error_taxonomy}

To conduct a more fine-grained analysis of model failures in multimodal long-document question answering, we manually analyzed nearly 200 erroneous samples from the four evaluated models. We divide the error data into four stages, and further annotate each case with one of seven error types. This stage-wise annotation allows us to identify not only whether the final answer is wrong, but also where the evidence chain breaks down.

\paragraph{Stages}
We define four error stages:

\textbf{Page Stage} refers to errors that occur during evidence page retrieval, where the model fails to retrieve the complete or correct ground-truth evidence pages, or introduces incorrect pages.

\textbf{BBox Stage} refers to cases where the model retrieves the correct page but fails to accurately localize the evidence region within the page. Examples include incomplete bounding boxes, localization to the wrong table, row, column, or subfigure, or omission of essential contextual information.

\textbf{Fact Stage} refers to cases where the retrieved page and localized evidence region are largely correct, but the model reads, extracts, or states a fact that is semantically inconsistent with the ground truth.

\textbf{Final Answer Stage} refers to cases where the preceding evidence pages, localized regions, and factual information are largely usable, but the model still makes an error during final answer generation.

\paragraph{Error Types}
We define seven fine-grained error types:

\textbf{Hallucinated Evidence} occurs when the page or bounding box cannot support the fact claimed by the model. This includes cases where the model output lacks a citation, or where the citation only supports a positional description while the model subsequently generates unsupported factual content.

\textbf{Perception Error} refers to failures in reading visual or textual content, including errors in interpreting figures, tables, OCR text, numerical values, legends, labels, points, lines, or bars.

\textbf{Evidence Location Error} refers to incomplete or incorrect evidence localization, such as missing evidence pages, incomplete bounding boxes, localization to adjacent rows, wrong tables, wrong subfigures, or omission of necessary titles, legends, and contextual information.

\textbf{Distractor Evidence} refers to cases where the model introduces irrelevant or distracting evidence, such as unrelated pages, incorrect tables, or irrelevant regions on the same page, and incorporates them into the reasoning process.

\textbf{Reasoning/Calculation Error} occurs when the extracted facts are largely correct, but the model makes errors in subsequent reasoning, calculation, or aggregation, such as difference computation, counting, deduplication, time interval calculation, unit conversion, or format conversion.

\textbf{Question Misinterpretation} refers to errors caused by misunderstanding the intent of the question, including the target entity, scope, definition, comparison criterion, or constraints.

\textbf{Over-answer/Unanswerable Error} refers to incorrect answerability judgments, such as providing a concrete answer when the question should be considered unanswerable, or incorrectly refusing to answer when sufficient evidence is available.

As shown in Fig.~\ref{fig:error_analysis}, the error distribution exhibits clear stage-specific patterns. In the Page Stage, Evidence Location Error accounts for the largest proportion of errors, reaching 56.1\%, indicating that coarse-grained page retrieval is still a major bottleneck. In the BBox Stage, Evidence Location Error remains the most frequent error type at 32.7\%, while Perception Error also becomes prominent at 30.8\%. This suggests that even after retrieving the correct page, models often fail to identify the precise supporting region or correctly interpret the localized visual evidence. In the Fact Stage, Perception Error dominates the distribution, accounting for 76.5\% of errors, which shows that models still struggle to accurately read tables, charts, OCR text, and other visual elements when the evidence location is largely correct. In the Final Answer Stage, errors mainly come from Question Misinterpretation and Reasoning/Calculation Error, accounting for 30.3\% and 27.3\%, respectively. This indicates that even when the evidence chain is mostly usable, models may still fail to understand the question intent or perform the required reasoning, calculation, or aggregation.

Overall, these findings suggest that current MLLMs still face substantial challenges in producing correct and verifiable reasoning trajectories. The main failure source shifts from evidence localization in the early stages, to visual perception in the fact extraction stage, and finally to question understanding and reasoning in the answer generation stage. Therefore, improving multimodal long-document QA requires progress not only in page-level retrieval, but also in fine-grained evidence grounding, factual perception, and robust reasoning over extracted evidence.

\section{Broader Impacts}
\label{appendix:Broader_impacts}

This research delivers positive social value by advancing the development of more credible and auditable long-document question-and-answer systems. By requiring models to provide supporting pages, locate evidence regions, present fundamental factual content, and deliver final answers, DocScope enables researchers to identify when models produce unsubstantiated or hallucinated responses. This is extremely valuable for high-risk and document-intensive fields such as education, law, finance, healthcare, and public administration. The benchmark also reveals the gap between answer accuracy and the reliability of evidence chains, prompting future related systems to prioritize process transparency rather than merely pursuing the correctness of final answers. Meanwhile, this research may also exert negative impacts if the performance improvements in benchmark tests are mistakenly regarded as proof that current multimodal large models can be fully and reliably applied to real-world decision-making.

\clearpage
\section{Datasheet for DocScope}

\subsection{Motivation}

\noindent \textbf{1. For what purpose was the dataset created? Was there a specific task in mind? Was there a specific gap that needed to be filled? Please provide a description.}

\textbf{A1:} DocScope is a benchmark designed to evaluate the multimodal long-document understanding capabilities of models, primarily targeting question answering over long PDF documents. Unlike existing benchmarks that mainly focus on end-to-end answer accuracy, DocScope aims to provide a more fine-grained diagnostic evaluation by assessing models’ abilities in evidence localization, information extraction, cross-page reasoning, and hallucination control. The dataset fills the gap in existing long-document evaluations where the source of model failures is difficult to distinguish, enabling researchers to more clearly analyze whether errors arise from localization, extraction, reasoning, or generation.

\noindent \textbf{2. Who created this dataset (e.g., which team, research group) and on behalf of which entity (e.g., company, institution, organization)?}

\textbf{A2:} This dataset is created by the authors of this paper.

\noindent \textbf{3. Who funded the creation of the dataset? If there is an associated grant, please provide the name of the grantor and the grant name and number.}
  
\textbf{A3:} N/A.

\subsection{Composition}

\noindent \textbf{1. What do the instances that comprise the dataset represent? Please provide a description.}

\textbf{A1:} DocScope currently contains 1,124 QA instances, each consisting of a question and its corresponding answer grounded in a long document. The questions cover both single-page understanding and multi-page reasoning scenarios, aiming to evaluate model performance under different evidence scopes. The dataset includes unanswerable questions as well as seven answerable question types: Visual Element Counting \& Identification, Document Structure \& Metadata, Numerical \& Statistical Data, Technical Systems \& Operating Procedures, Entity Attributes \& Comparative Relations, Semantic Content \& Conceptual Meaning, and Time, Date \& Sequential Relations. Overall, these instances represent common multimodal, multi-page, and multi-type information needs in real-world long-document understanding scenarios.

\noindent \textbf{2. How many instances are there in total (of each type, if appropriate)?}

\textbf{A2:} There are 1,124 QA instances in total, including 1,046 answerable instances across seven question types---Visual Element Counting \& Identification, Document Structure \& Metadata, Numerical \& Statistical Data, Technical Systems \& Operational Procedures, Entity Attributes \& Comparative Relationships, Semantic Content \& Conceptual Meaning, and Time, Date \& Sequential Relationships---plus 78 unanswerable instances.

\noindent \textbf{3. Does the dataset contain all possible instances or is it a sample? If the dataset is a sample, then what is the larger set?}

\textbf{A3:} DocScope is a curated sample rather than an exhaustive collection of all possible instances. The larger set consists of potential QA pairs constructed from real-world long documents. In DocScope, QA pairs are synthesized from real documents using a strong multimodal model,  Claude-Opus-4.6, followed by strict review to ensure quality and reliability.

\noindent \textbf{4. What data does each instance consist of? “Raw” data or features?}

\textbf{A4:} Each instance in DocScope consists of a question, an answer, the supporting evidence required for reasoning, evidence bounding-box coordinates in the document, and the specific facts used in the reasoning process. The data are provided as raw document-based QA annotations rather than pre-extracted feature representations.

\noindent \textbf{5. Is there a label or target associated with each instance? If so, please provide a description.}

\textbf{A5:} Yes. Each instance is associated with a target answer. For answerable questions, the target is the ground-truth answer derived from the supporting evidence in the document, along with evidence annotations such as evidences and bounding-box coordinates. For unanswerable questions, the target indicates that the question cannot be answered based on the document content.

\noindent \textbf{6. Is any information missing from individual instances?}

\textbf{A6:} No. 

\noindent \textbf{7. Are relationships between individual instances made explicit?}

\textbf{A7:} Yes. Instances are explicitly categorized by question type and answerability. Answerable instances are grouped into seven question types: Visual Element Counting \& Identification, Document Structure \& Metadata, Numerical \& Statistical Data, Technical Systems \& Operational Procedures, Entity Attributes \& Comparative Relationships, Semantic Content \& Conceptual Meaning, and Time, Date \& Sequential Relationships. Unanswerable instances are treated as a separate category.

\noindent \textbf{8. Are there recommended data splits (e.g., training, development/validation, testing)?}

\textbf{A8:} Yes. DocScope is primarily intended as an evaluation benchmark and is split into a validation set and a test set.

\noindent \textbf{9. Are there any errors, sources of noise, or redundancies in the dataset?}

\textbf{A9:} DocScope is constructed through model-assisted synthesis followed by strict review to reduce errors, noise, and unsupported annotations. However, as the QA pairs are synthesized from complex real-world long documents, residual annotation errors or ambiguous cases may still exist.

\noindent \textbf{10. Is the dataset self-contained, or does it link to or otherwise rely on external resources?}

\textbf{A10:} DocScope is self-contained. The released dataset will include the source PDF documents, questions, answers, supporting evidence, evidence bounding-box coordinates, and factual reasoning annotations.

\noindent \textbf{11. Does the dataset contain data that might be considered confidential?}

\textbf{A11:} No. The source documents are publicly available documents, and the dataset does not intentionally contain confidential information.

\noindent \textbf{12. Does the dataset contain data that might be offensive?}

\textbf{A12:} DocScope is not intended to contain offensive content. Manual review and filtering were conducted to remove or mitigate offensive, toxic, or sensitive content.

\subsection{Collection Process}

\noindent \textbf{1. How was the data associated with each instance acquired?}

\textbf{A1:} DocScope instances were acquired” from publicly available real-world long documents. QA pairs were synthesized using  Claude-Opus-4.6 and then strictly reviewed.

\noindent \textbf{2. What mechanisms or procedures were used to collect the data?}

\textbf{A2:} DocScope was built through a model-assisted and human-reviewed pipeline, including document collection, QA synthesis, evidence annotation, and quality review.

\noindent \textbf{3. If the dataset is a sample from a larger set, what was the sampling strategy?}

\textbf{A3:} Purposeful sampling was used to cover representative long-document QA scenarios, including different evidence scopes, answerability settings, and question types.

\noindent \textbf{4. Who was involved in the data collection process?}

\textbf{A4:} DocScope was created by the authors, with 13 additional dedicated annotators involved in annotation and review. The annotators worked for approximately five days and were compensated.

\noindent \textbf{5. Over what timeframe was the data collected?}

\textbf{A5:} The data collection, synthesis, annotation, and review process took approximately two weeks.

\subsection{Preprocessing/cleaning/labeling}

\noindent \textbf{1. Was any preprocessing/cleaning/labeling of the data done?}

\textbf{A1:} Yes. The dataset construction involved QA synthesis, evidence annotation, bounding-box annotation, factual reasoning annotation, and strict review. The review process was used to verify that each answer was supported by the corresponding document evidence and that unanswerable questions were correctly labeled.

\noindent \textbf{2. Was a data decontamination strategy employed?}

\textbf{A2:} Yes. Data decontamination was conducted through manual review and experimental checks to remove contaminated and duplicate instances.

\noindent \textbf{3. Is the software used to preprocess/clean/label the instances available?}

\textbf{A3:} Yes. The tools and scripts used for data generation will be released.

\subsection{Uses}

\noindent \textbf{1. Has the dataset been used for any tasks already? If so, please provide a description.}

\textbf{A1:} Yes. DocScope is used to evaluate multimodal long-document question answering. It supports fine-grained diagnosis of model abilities in evidence localization, information extraction, cross-page reasoning, answer generation, and hallucination control.

\noindent \textbf{2. Is there a repository that links to any or all papers or systems that use the dataset? If so, please provide a link or other access point.}

\textbf{A2:} N/A.

\noindent \textbf{3. What (other) tasks could the dataset be used for?}

\textbf{A3:} In addition to long-document question answering, DocScope can be used for evaluating evidence localization, multimodal information extraction, cross-page reasoning, visual grounding in documents, hallucination detection, and the robustness of models on unanswerable document-based questions.

\noindent \textbf{4. Is there anything about the composition of the dataset or the way it was collected that might impact future uses? Is there anything a future user could do to mitigate these undesirable harms?}

\textbf{A4:} Since the QA pairs are synthesized using a strong multimodal model and then reviewed, the dataset may reflect the coverage and biases of the source documents and the synthesis process. Future users should consider DocScope as an evaluation benchmark rather than a fully exhaustive representation of all long-document understanding scenarios. Potential risks can be mitigated by reporting performance by question type, evidence scope, and answerability, rather than relying only on aggregate accuracy.

\noindent \textbf{5. Are there tasks for which the dataset should not be used? If so, please provide a description.}

\textbf{A5:} No.

\subsection{Distribution}

\noindent \textbf{1. Will the dataset be distributed to third parties outside of the entity? If so, please provide a description.}

\textbf{A1:} Yes. DocScope will be publicly released to the research community after the paper is accepted.

\noindent \textbf{2. How will the dataset will be distributed? Does the dataset have a digital object identifier (DOI)?}

\textbf{A2:} DocScope will be distributed through GitHub and Hugging Face. No DOI is currently available.

\noindent \textbf{3. When will the dataset be distributed?}

\textbf{A3:} The dataset will be distributed after the paper is accepted.

\noindent \textbf{4. Will the dataset be distributed under a copyright or other license?}

\textbf{A4:} Yes. The dataset will be distributed under the Creative Commons Attribution-NonCommercial-ShareAlike 4.0 International License (CC BY-NC-SA 4.0).

\noindent \textbf{5. Have any third parties imposed IP-based or other restrictions on the data associated with the instances?}

\textbf{A5:} No. The source documents are from publicly available resources, and no third-party IP-based or other restrictions have been imposed on the dataset.

\noindent \textbf{6. Do any export controls or other regulatory restrictions apply to the dataset or to individual instances?}

\textbf{A6:} No.

\subsection{Maintenance}

\noindent \textbf{1. Who will be supporting/hosting/maintaining the dataset?}

\textbf{A1:} The authors.

\noindent \textbf{2. How can the owner/curator/manager of the dataset be contacted (e.g., email address)?}

\textbf{A2:} Email addresses will be provided on the project homepage post-publication.

\noindent \textbf{3. Is there an erratum? If so, please provide a link or other access point.}

\textbf{A3:} Any errata will be posted on the project GitHub repository.

\noindent \textbf{4. Will the dataset be updated? If so, please describe how often, by whom, and how updates will be communicated to users?}

\textbf{A4:} Yes. The authors plan to update the dataset, and updates will be communicated through the official GitHub and Hugging Face repositories.

\noindent \textbf{5. Will older versions of the dataset continue to be supported/hosted/maintained? If so, please describe how.}

\textbf{A5:} Yes. Older versions will be retained to support reproducibility.

\noindent \textbf{6. If others want to extend/augment/build on/contribute to the dataset, is there a mechanism for them to do so?}

\textbf{A6:} N/A.

\end{document}